\documentclass[sigconf,nonacm]{acmart}
\acmSubmissionID{papers\_523s1}

\author{Taesoo Kwon}
\affiliation{%
  \institution{Hanyang University}
  \city{Seoul}
  \country{South Korea}}
\email{taesoo@hanyang.ac.kr}

\author{Taehong Gu}
\affiliation{%
  \institution{Hanyang University}
  \city{Seoul}
  \country{South Korea}}
\email{gestoru@gmail.com}

\author{Jaewon Ahn}
\affiliation{%
  \institution{Hanyang University}
  \city{Seoul}
  \country{South Korea}}
\email{vpersie@hanyang.ac.kr}

\author{Yoonsang Lee}
\affiliation{%
  \institution{Hanyang University}
  \city{Seoul}
  \country{South Korea}}
\email{yoonsanglee@hanyang.ac.kr}

\usepackage{booktabs} %

\usepackage[ruled]{algorithm2e} %

\SetAlFnt{\small}
\SetAlCapFnt{\small}
\SetAlCapNameFnt{\small}
\SetAlCapHSkip{0pt}

\acmJournal{TOG}

\usepackage{amsmath}
\usepackage{amsfonts}
\usepackage{kotex}
\usepackage{multirow}
\usepackage{caption}
\usepackage{siunitx}
\usepackage{graphicx,subfigure}
\newcommand{\mb}{\mathbf}
\usepackage{array}
\newcolumntype{P}[1]{>{\centering\arraybackslash}p{#1}}

\usepackage{bigstrut}
\usepackage{multirow}

\usepackage{xcolor}
\definecolor{ys}{rgb}{0,0,0}    %
\definecolor{gu}{rgb}{0,0,0}    %

\definecolor{ysr}{rgb}{0,0,0}    %
\definecolor{ysq}{rgb}{0,0,0}    %
\definecolor{taesoo}{rgb}{0,0,0}    %
\definecolor{yst}{rgb}{.0,.0,.0}    %
\definecolor{ys2}{rgb}{0,0,0}    %
\definecolor{yq2}{rgb}{0,0,0}    %
\definecolor{taesoo2}{rgb}{0,0,0}    %
\definecolor{ys3}{rgb}{0,0,1}    %
\definecolor{rev}{rgb}{0,0,0}    %

\usepackage{wrapfig}

\begin{document}

\title{Adaptive Tracking of a Single-Rigid-Body Character in Various Environments}

  \begin{teaserfigure}
	\centering
  \includegraphics[trim=50 0 150 0, clip, width=.24\linewidth]{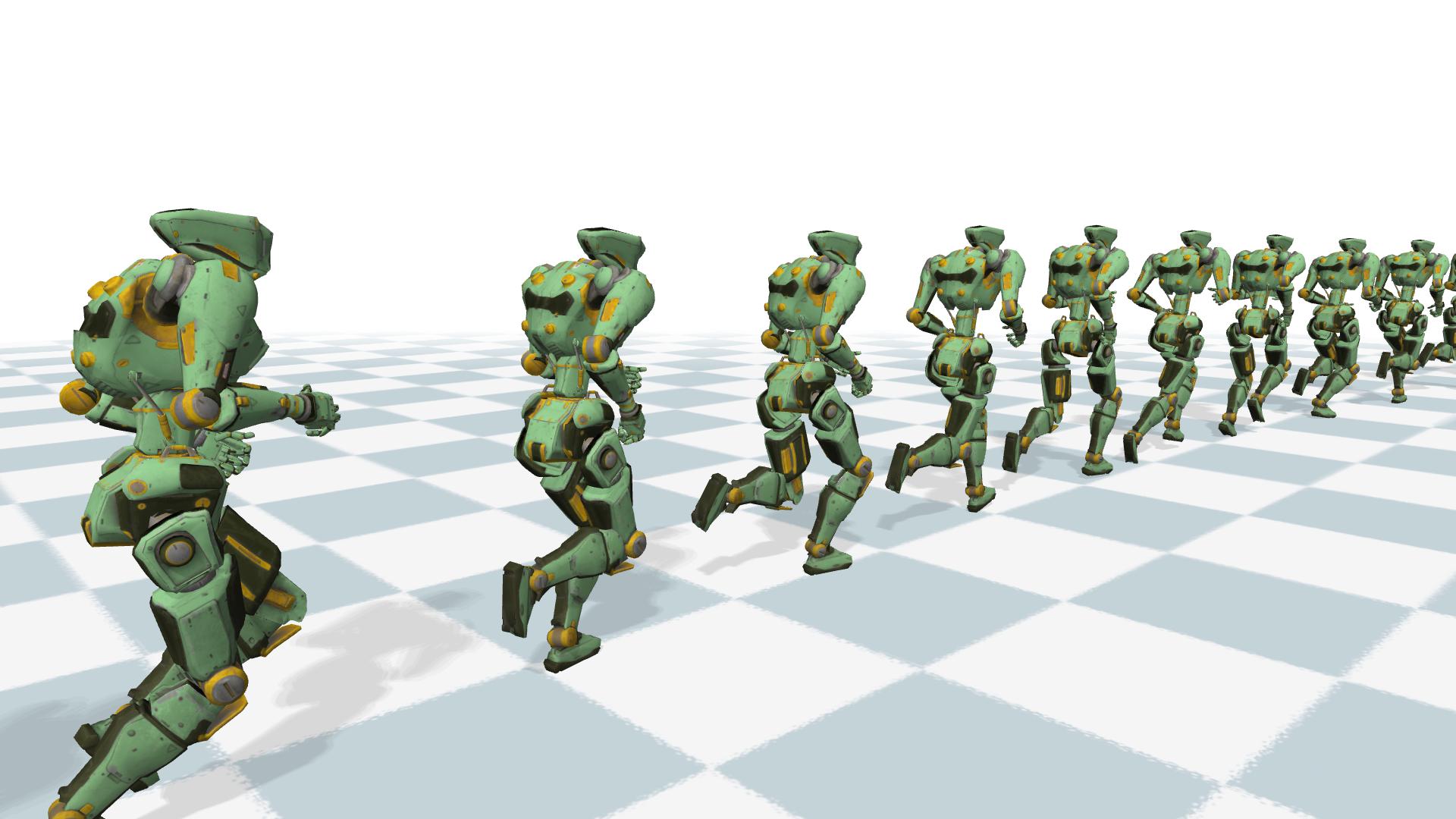}
 \includegraphics[trim=50 0 150 0, clip, width=.24\linewidth]{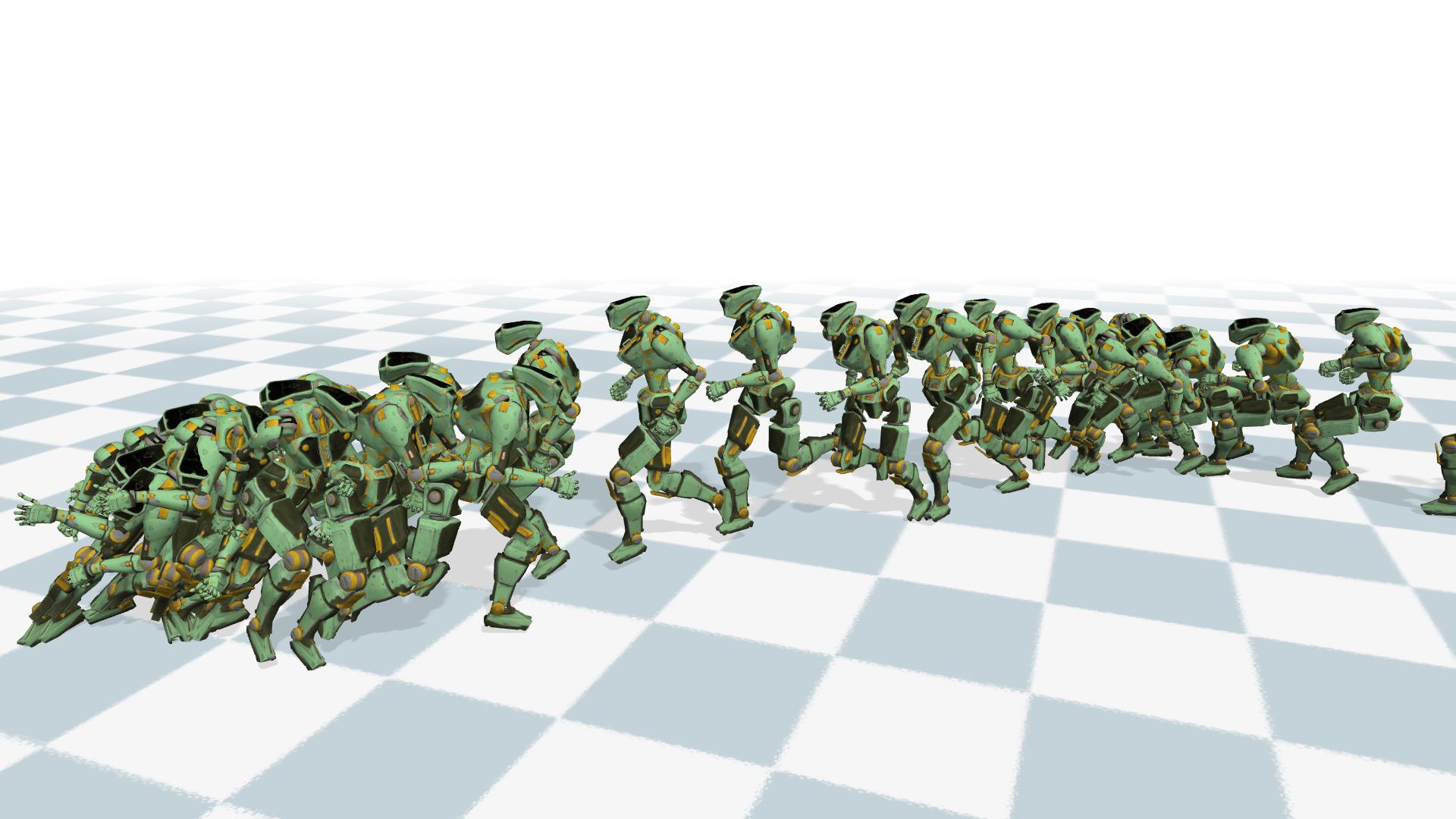}
  \includegraphics[trim=50 0 150 0, clip, width=.24\linewidth]{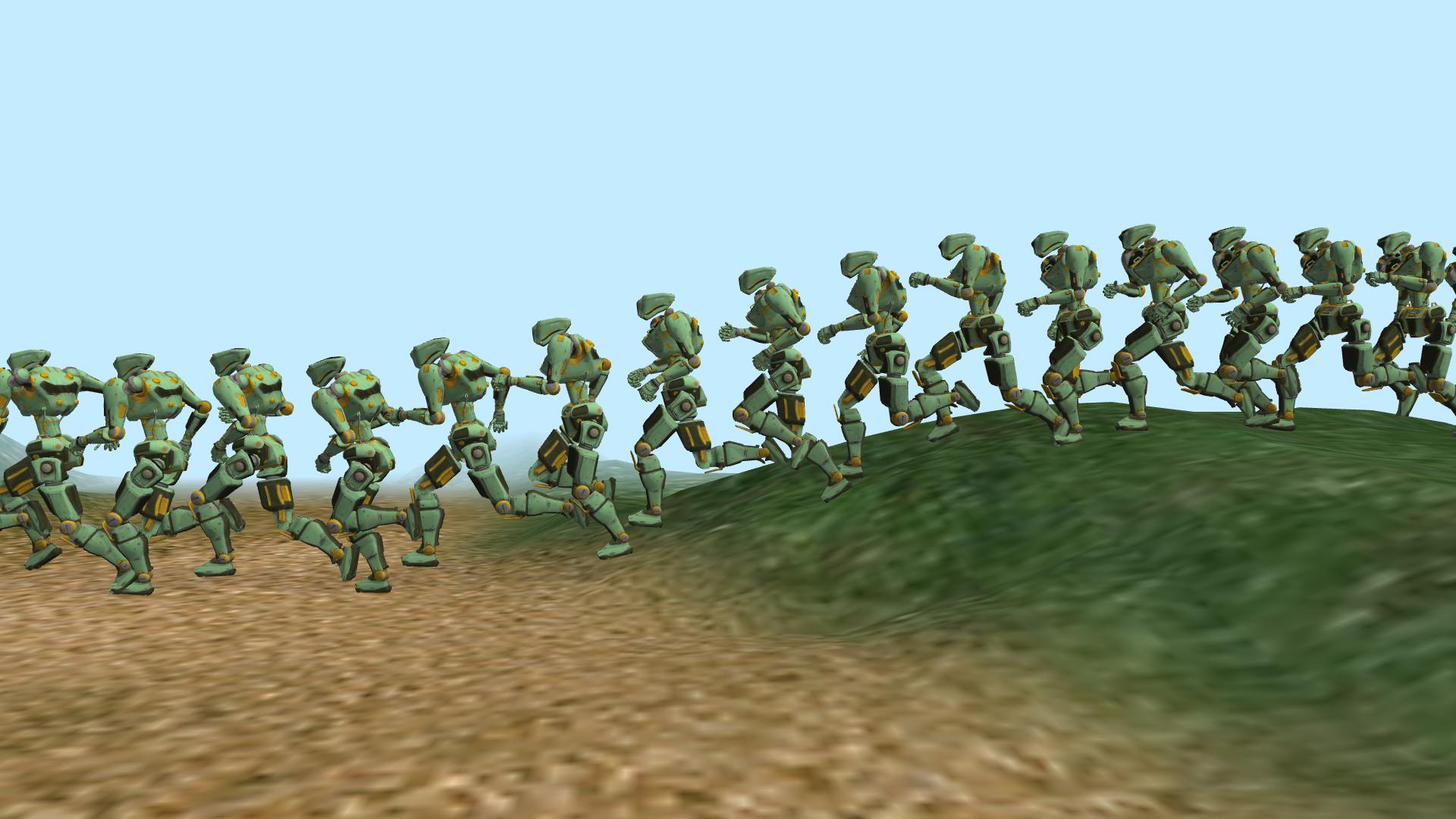}
  \includegraphics[trim=250 40 250 150, clip, width=.24\linewidth]{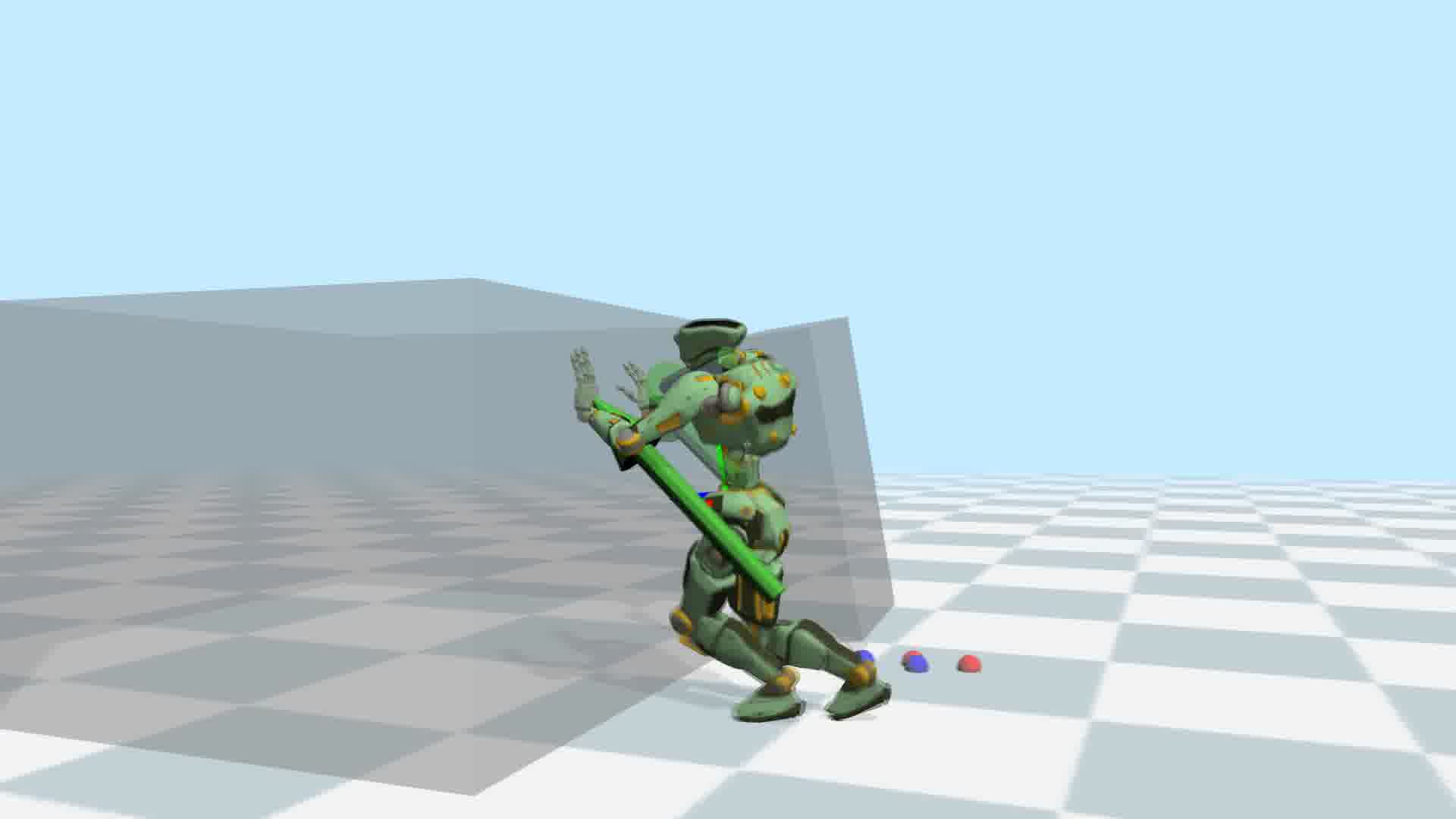}
    \caption{
        Our character can perform tasks in various environments using a policy learned from a short reference motion clip on flat ground. The results suggest that our controller can perform a range of activities, such as transitioning, climbing on rough terrain, and pushing objects, without additional training.
	  }
  \end{teaserfigure}

\begin{abstract}

Since the introduction of DeepMimic~\cite{peng2018deepmimic}, subsequent research has focused on expanding the repertoire of simulated motions across various scenarios.
In this study, we propose an alternative approach for this goal, a deep reinforcement learning method based on the simulation of a single-rigid-body character.
    Using the centroidal dynamics model (CDM) to express the full-body character as a single rigid body (SRB) and training a policy to track a reference motion, we can obtain a policy that is capable of adapting to various unobserved environmental changes and controller transitions without requiring any additional learning.
Due to the reduced dimension of state and action space, the learning process is sample-efficient.
The final full-body motion is kinematically generated in a physically plausible way, based on the state of the simulated SRB character.
    The SRB simulation is formulated as a quadratic programming (QP) problem, and the policy outputs an action that allows the SRB character to follow the reference motion.
    We demonstrate that our policy, efficiently trained within 30 minutes on an ultraportable laptop, has the ability to cope with environments that have not been experienced during learning, such as running on uneven terrain or pushing a box, and transitions  between learned policies, without any additional learning.

\end{abstract}


\maketitle

\section{Introduction}

Approaches based on deep reinforcement learning (DRL) are producing significant results in the control of physically simulated characters in recent years. 
DeepMimic, regarded as one of the pioneering studies, demonstrated exceptional motion quality for complex motions through motion capture data imitation~\cite{peng2018deepmimic}.
Expanding the repertoire of motions in diverse scenarios beyond mere imitation has been a focus of subsequent research.
Numerous efforts have been made in this direction, such as employing large motion datasets~\cite{bergamin_drecon_2019} and generative models~\cite{controlvae22}, or training controllers to fulfill a broad range of requirements without relying on specific reference motions~\cite{Lee:2021:Parameterized}.
However, these approaches often required longer learning time due to the need to experience and adapt to various changes in the environment or tasks.

In this study, we propose an alternative and orthogonal approach to the goal of going beyond examples. 
We express the character as a single rigid body (SRB) by using a simplified physical model called the centroidal dynamics, and train a policy for the SRB character to track a reference motion.
Then the learned policy is capable of adapting to various unobserved changes in the environment and controller transitions without any additional learning.
Additionally, due to the greatly reduced volume of state and action space, the learning process is sample-efficient.
The physical simulation and policy learning are conducted on the SRB character, while final full-body motion is kinematically generated from the simulated SRB states.
The policy obtains inputs on the state of the SRB character and produces actions to follow the given reference motion. 
In the simulation stage, a quadratic programming (QP) solver calculates the contact forces that best achieves the desired acceleration computed from the action and the reference motion. This is then used to update the state of the SRB character.

Our approach defines reinforcement learning (RL) tasks in a less specific (excluding full-body details) and more general form
(based on the first principle that character movement is caused by contact forces),
thereby allowing the agent to be less dependent on a particular situation of the full-body character. 
As a result, our method robustly performs transitions by switching or blending of learned policies, %
and shows the ability to cope with environments that have not been experienced during learning, such as running on uneven terrain, pushing a box, or balancing against external forces,
without any additional learning or parameterization.
Furthermore, our method is sample-efficient enough to obtain such a adaptive tracking policy in 30 minutes on an ultraportable laptop.

\section{Related work}

Earlier physics-based motion generation research manually created a locomotion control algorithm based on error feedback~\cite{yin2007simbicon,coros2010generalized,lee2010data}. 
Parameter optimization and simple balancing rules were used to generate a wide range of robust motion repertoire~\cite{agrawal2013diverse,ha2014iterative,wang2012optimizing}.
Controllers based on QP leveraging the equations of motion were developed to enhance robustness~\cite{abe2007multiobjective, da2008simulation,kwon2017momentum}.
Non-linear trajectory optimization was used to synthesize physically probable motions for various tasks by simultaneously considering multiple frames, as opposed to the single frame optimization used in QP~\cite{ye2010optimal,mordatch2012discovery,wampler2014generalizing}. 
Model-predictive control (MPC) methodologies performed online optimization, allowing the construction of complex motions in unknown environments~\cite{Macchietto09,tassa2012synthesis,hamalainen2015online}.
However, these MPC techniques that utilized the full-body physical model state space had limitations in terms of motion quality and long-term planning.
MPC techniques with simplified models allow interactive controls in more complicated environments, focusing on locomotion tasks~\cite{winkler2018gait,kwon2020fast}.
Some robotics studies have leveraged simplified models to enhance their locomotion control strategies~\cite{viereck2021learning,tsounis2020deepgait}.

RL has been widely used to develop controllers for simulated characters.
Earlier studies showed the advantage of creating controllers with minimal manual effort by designing simple rewards~\cite{coros2009robust,peng2015dynamic}. 
DRL expanded the capabilities to tackle various tasks~\cite{brockman2016openai,duan2016benchmarking,liu2017learning,peng2016terrain,rajeswaran2017learning}.
Some studies employed simplified models or multi-level learning for efficient long-term planning~\cite{2017-TOG-deepLoco,Brachiation2022}.
Notably, DeepMimic has achieved impressive motion quality for numerous reference motions~\cite{peng2018deepmimic},
but faced limitations in controller performance beyond reference motions.
Recent research has employed large motion datasets and encoding techniques to diversify the motions generated by RL-based controllers and to improve their generalization capability~\cite{Chentanez2018,bergamin_drecon_2019,Park2019,controlvae22,won_physics-based_2022,peng_ase_2022,peng18}.
Some studies trained controllers to satisfy a wide range of requirements without using corresponding reference motions, for example, transitions between different actions or changes in jump height~\cite{jump2021,chimeras22,Lee:2021:Parameterized,allsteps20,peng_amp_2021}.
Since these full-body simulation-based RL approaches require observation of various changes in environments and intent during the learning process, they often require a longer time for learning. 
In our method, a policy trained to track only a single reference motion within a relatively short time can respond to various environmental changes without experiencing such cases at training time, due to the flexibility of SRB-based modeling.
However, not based on full-body dynamics, our approach does not directly simulate each part of the body. As a result, it has the drawback of not being able to represent contact-rich physical interactions occurring at various parts of the actual full body. Contact forces can be applied only to the manually-classified contact points of the SRB character.

There have been numerous studies that use a large amount of motion data to train models and generate realistic real-time full-body motions purely kinematically~\cite{holden_phase_2017,zhang_2018,starke_neural_2019,Ling_2020,cho_motion_2021}.
Our approach can also be viewed as a kinematic controller, as it generates the final full-body motions kinematically.
However, unlike these methods that fail to exhibit physical interactions in response to environmental changes, our approach performs accurate physics simulation at the level of the SRB.
This allows us to create physically plausible full-body motions by reflecting the changes in the state of the SRB due to variations in the physical environment.

Among various previous studies, the following two studies are the closest to our study.
The research by Xie et al.~\cite{xie2022glide} has many commonalities with our work in its use of the SRB model, QP, and RL.
However, unlike \cite{xie2022glide} which adopted to generate four-legged locomotion by using manually set gait parameters such as foot phase offsets, we focus on creating dynamic and natural motions of a two-legged character, walking, and various other motions, by tracking motion capture data. 
Whereas a simple Raibert-style heuristic was used to determine foot placement in \cite{xie2022glide}, our policies are trained to output desired foot landing positions to ensure balancing. 

\cite{kwon2020fast} is closed to our study in terms of using the SRB model for a two-legged character, but has the following key differences:
i) \cite{kwon2020fast} generates motions through per-segment trajectory optimization, which takes too much time for smooth real-time performance.
Our RL-based system can generate motion at a much faster speed than real-time.
ii) To mitigate the runtime performance issue, \cite{kwon2020fast} proposed a supervised learning network that takes pendulum and footstep plans and generates full-body motion.
However, it may fail to produce adequate motion in scenarios deviating significantly from the training data, such as unexpected external forces. %
Our method can train an adaptive controller without additional learning, even in significantly different scenarios from the reference motion used in training.
iii) In \cite{kwon2020fast}, motion generation occurs per-segment, with planners generating trajectories at half-cycle intervals for a short future horizon.
Consequently, if an unexpected external force is applied during runtime, the character may respond with a half-cycle delay.
Our policy produces the desired contact position every frame, enabling an immediate response to external forces.

\begin{figure*}[tbp]
  \centering
  \includegraphics[trim=25 50 25 20, clip, width=.95\linewidth]{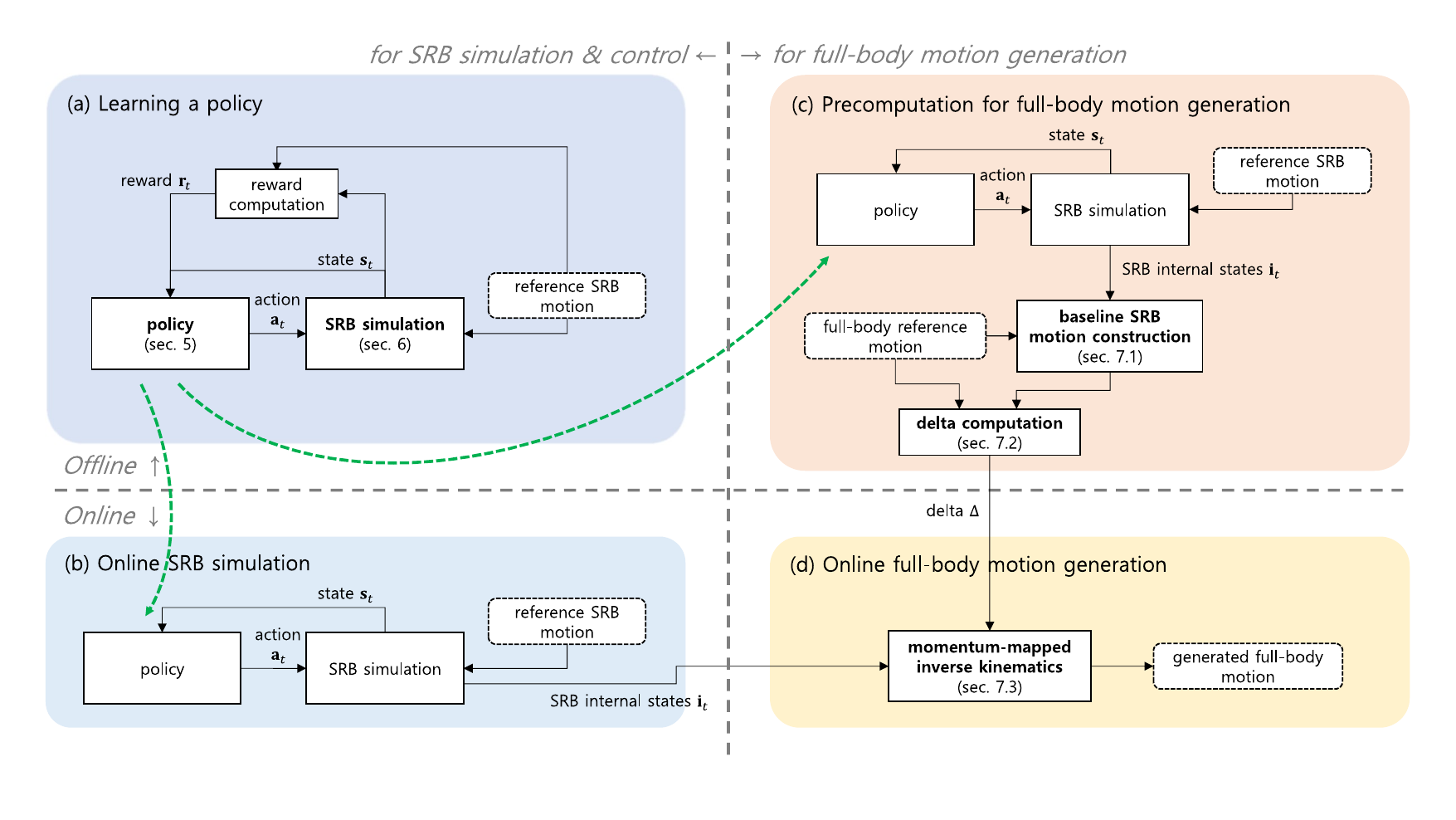}
  \caption{
      System overview.
           }
\label{fig:overview}
\end{figure*}

\section{SRB character and frames}

\begin{wrapfigure}{r}{0.17\columnwidth}
  \begin{center}
      \includegraphics[trim=50 50 70 20, clip, width=0.17\columnwidth]{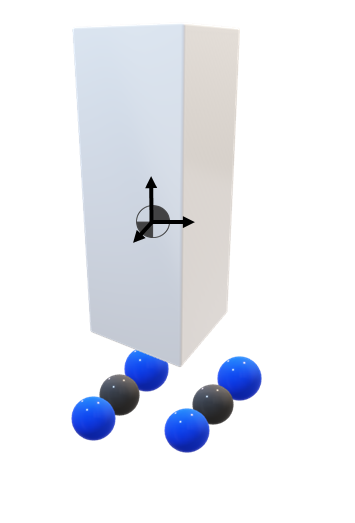}
  \end{center}
  \caption{\label{fig:singlerigidbody}
    SRB character
    }
\end{wrapfigure}

\textbf{SRB character} is a simplified representation of the key physical characteristics of an articulated character, %
consisting of a box-shaped rigid body (the gray box in Figure~\ref{fig:singlerigidbody}) that approximates the inertia of the full-body character, as well as four contact points (the blue spheres) attached on either side of the two feet (the black spheres indicating their centers).
\textcolor{rev}{The orientation of SRB center of mass is defined as the orientation of the box}, which implies the overall orientation of the character. 
The mass and inertia of the box are set as the mass (\SI{60}{kg}) of the reference character and its composite rigid body inertia calculated from the default posture (at attention posture).

\textbf{Reference SRB motion} is a reference motion expressed by the SRB character, %
whose center of mass position and orientation are set to those of the pelvis in the full-body reference motion.
The contact timing and position of the feet of the reference SRB motion are set identically to those of the full-body reference motion.
The simulated SRB character is controlled to contact the ground at the same time as the reference contact timing.

\textbf{SRB frames}. This paper uses the following three types of reference frames to express various states values:
\begin{itemize}
    \item \textbf{SRB frame} has the SRB character’s center of mass as its origin and is attached to the SRB character.
    \item \textbf{Forward-facing SRB frame} has the same origin as the SRB character frame, with its z-axis representing the horizontally projected z-axis of the SRB character frame (forward-facing direction), and its y-axis aligns with the global vertical axis.
    \item \textbf{Projected SRB frame} is obtained by vertically projecting the origin of the forward-facing SRB frame onto the ground or terrain.
\end{itemize}

Frames corresponding to the reference SRB motion shall be referred to by prefixing "reference" (e.g., "reference SRB frame").

\section{Overview}

Figure~\ref{fig:overview} outlines the overview of the system presented in this paper.

First, a policy is learned for the simulated SRB character to track a reference SRB motion (Figure~\ref{fig:overview} (a)).
In the SRB simulation, the desired acceleration of the SRB character is determined by both the action from the policy and the reference SRB motion. 
Then, a quadratic programming (QP) problem is solved to find the actual implementable acceleration that can closely match the desired acceleration based on the given internal state of the SRB character. 
The internal state of the SRB character is updated by integrating the obtained acceleration.
The reward is calculated by comparing the updated SRB internal state and the reference SRB motion.

The learned policy can be used to simulate the SRB character at runtime (Figure~\ref{fig:overview} (b)).
It allows not only tracking a single reference motion but also transitions between different motions through policy switching or blending, without requiring additional learning.

\textcolor{rev}{The simplified SRB model alone has limitations in representing the details of full-body motion. Therefore, in an offline stage, we calculate the difference between simulated SRB motion and full-body reference motion to capture the reference full-body details (Figure~\ref{fig:overview}~(c)).
First,} the baseline SRB motion, which is the tracking result of the reference SRB motion by the learned policy, is constructed.
Then the difference (delta $\Delta$) between baseline SRB motion and the full-body reference motion \textcolor{rev}{(such as SRB foot and full-body foot positional differences)} is calculated.
\textcolor{rev}{This delta is later used in the online full-body motion generation.}

\textcolor{rev}{In the online full-body motion generation,} the simulated SRB motion generated at runtime is converted into full-body motion by momentum-mapped inverse kinematics (Figure~\ref{fig:overview} (d)). 
The precomputed delta $\Delta$ is combined to produce a more realistic full-body motion by adding the detailed full-body data that were not expressed by the SRB motion.

\section{Policy representation}
\label{sec:policy_rep}

\textbf{State} $\mb{s} \in \mathbb R^{21}$ is composed of the following components.

\begin{itemize}

    \item \textbf{Center of mass state} $\mb s_c \in \mathbb R^{11}$ consists of the height, orientation (in unit quaternion), and generalized velocity (angular and linear velocities) of the center of mass of the SRB character, which are all expressed in the projected SRB frame.

    \item \textbf{Foot state} $\mb s_f^j \in \mathbb R^{4} (j \in \{\mathrm{left, right}\})$ consists of the center positions and rotations of each foot of the SRB character in the horizontal plane.
These values are expressed in the forward-facing SRB frame.

\item \textbf{Motion phase} $\psi(t) \in \left[ 0, 2\pi \right]$ indicates what point of the reference motion the current motion of the SRB character corresponds.
        This is actually stored in state $\mb{s}$ as a 2D-vector $\left(\mathrm{sin}\left(\psi(t)\right), \mathrm{cos}\left(\psi(t)\right)\right)$.

\end{itemize}

\textbf{Action} $\mb{a} \in \mathbb R^{10}$ consists of the following components.

\begin{itemize}

    \item \textbf{Desired foot landing position} $\mb a^j_s \in \mathbb R^{2} (j \in \{\mathrm{left, right}\})$ in the forward-facing SRB frame.
    Note that the vertical position is not included.

\item \textbf{Desired velocity of center of mass} $\mb a_v \in \mathbb R^{6}$ refers to the desired relative linear and angular velocities of the center of mass of the SRB character against that of the reference SRB motion.
    This is expressed with respect to the SRB frame.

\end{itemize}

\textbf{Reward} $r_t$ at each step $t$ is configured as below: %
\begin{equation}
\label{eq:reward}
r_t = w^s r_t^s - w^m r_t^m .
\end{equation}
Here, $r_t^s$ is the alive reward given when the episode does not end %
and $r_t^m$ is the mimic reward term that guides the SRB character to closely follow the reference SRB motion.

The mimic reward term $r_t^m$ is calculated as follows:
\begin{equation}
\label{eq:reward_mimic}
r_t^m = w^p r_t^p + w^e r_t^e. %
\end{equation}
$r_t^p$ is a posture reward term, which gives higher rewards when the states of the centers of mass of both SRB character and reference SRB motion are alike, which is calculated as follows:
\begin{equation}
\begin{aligned}
\label{eq:reward_posture}
    r_t^p = & w^{p_1}\| \Delta \mb p_t - \Delta \hat{\mb p}_{\psi(t)} \|+ w^{p_2}\| \Delta \mb R_t \ominus \Delta \hat{\mb R}_{\psi(t)} \| \\
        & + w^{p_3}\| \mb p_t - \hat{\mb p}_{\psi(t)} \| + w^{p_4}\| \mb R_t \ominus \hat{\mb R}_{\psi(t)} \| .
\end{aligned}
\end{equation}
Here, $\Delta \mb p_t$ and $\Delta \mb R_t $ respectively refer to the changes in the center of mass' position and orientation of the SRB character between time $t$ and $t +1$, while $\Delta \hat{\mb p}_{\psi(t)}$ and $\Delta \hat{\mb R}_{\psi(t)} $ refer to those changes in the reference SRB motion between the corresponding time points. 
$\mb p_t$ and $\mb R_t$ respectively refer to the position and orientation of the SRB character’s center of mass at time $t$, while  $\hat{\mb p}_{\psi(t)}$ and $\hat{\mb R}_{\psi(t)}$ refer to those of the reference SRB motion. %
$\|\mb{R}_1\ominus \mb{R}_2\|$ refers to the minimum angle of rotation between the two rotation matrices.
To compare the difference in the character height and the leaning angle of the body, the above terms are expressed in the projected frame of the respective character, that is, the projected SRB frame or the reference projected SRB frame.
\textcolor{rev}{The values of the weights $w^*$ used in our experiments are given in the supplementary material.}

$r_t^e$ is an end-effector reward, which has a higher value when the support foot positions of both the SRB character and reference SRB motion are alike. This reward is only applied to the foot in contact with the ground in the reference SRB motion and is calculated as follows:
\begin{equation}
\label{eq:reward_ee}
    r_t^e=\sum_j \sum_k \| \mb{c}^{jk}_t - \hat{\mb{c}}^{jk}_{\psi(t)} \|^2 ,
\end{equation}
where $\mb{c}^{jk}_t$ refers to the position of the $k^{th}$ contact point $(k \in \{\mathrm{toe, heel}\})$ of the $j^{th}$ foot $(j \in \{\mathrm{left, right}\})$ of the SRB character at time $t$ with respect to the SRB frame, while $\hat{\mb{c}}^{jk}_{\psi(t)}$ refers to that of the corresponding contact point of the reference SRB motion at the corresponding time, expressed with respect to the reference SRB frame. 
Note that the swing foot is excluded from this term because the SRB character does not directly depict the full-body character's actual swing motion of the foot.

\section{SRB simulation}
\label{sec:srb_simulation}

This section will explain the internal state of the SRB character at time $t$, where the input $\mb s_t$ for the RL policy is extracted, followed by the simulation process by QP that computes the internal state at time $t+1$. %

\subsection{Internal state of SRB character}
\label{sec:internal_state}

The internal state of the SRB character consists of the center of mass’ position and orientation, their time derivatives, and the position and orientation of the two feet.

\textbf{Center of mass state.} The global position $\mathbf p_t^{\{g\}}$ and global orientation $\mathbf R_t^{\{g\}}$ of the SRB character’s center of mass is used to calculate the SRB frame $\mb {T}_{t} \in \mathrm{SE}(3)$ at the center of mass as follows:
\begin{equation}
\label{eqn5}
	\mb T_t=\textrm{F}\left(\mathbf R_t^{\{g\}}, \mathbf p_t^{\{g\}} \right), 
\end{equation}
where $\textrm{F}$ is a function that outputs the rigid-body transformation from the position and rotation information.

The time derivatives of the center of mass position and orientation are expressed as the following generalized velocity  $\dot{\mb{q}}_t \in \mathrm{se}(3)$, which indicates the spatial velocity in body frame:
\begin{equation}
\label{eqn6}
    [\dot{\mb{q}}_t]=\mb{T}_{t}^{-1} \cdot \Dot{\mb{T}}_{t},
\end{equation}
where $\dot{\mb{T}}_{t}$ is the time derivative of $\mb{T}_{t}$, and $[\cdot]$ is the operator that converts se(3) in the 6-vector coordinates into a 4x4 matrix format.

\textbf{Foot state}
consists of the global position $\mb{f}^j_t$ and orientation $\mb{F}^j_t$ of each foot.
Their specific meaning varies depending on whether the $j^{th}$ foot is in the swing state, determined by the touch down and off timings of that foot in the reference SRB motion.
$\mb{f}^j_t$ is a horizontal position and
$\mb F^j_t$ is expressed as only the rotational component against the vertical axis.

When the $j^{th}$ foot is in a swing state, $\mb{f}^j_t$ moves continuously to the desired foot landing position $\mb{a}^j_s$.
This $\mb{a}^j_s$ is not directly suitable as a basis for reproducing the continuous movement of the swing foot since it changes discontinuously over time.
Therefore, we use a LQR filter presented in \cite{hwang2017performance}, to compute continuous $\mb{f}^j_t$ from the discontinuous  $\mb{a}^j_s$.
Similarly, by the LQR filter, $\mb{F}^j_t$ moves continuously to the desired foot landing orientation which is directly obtained from the reference SRB motion.
Further details of LQR filtering are described in the supplementary material.

When the $j^{th}$ foot is in a contact state, $\mb{f}^j_t$ refers to the actual global position of the foot, which remains fixed for the contact duration.
The moment the $j^{th}$ foot changes its state from swing to contact, the position of the foot and its rotation against the vertical axis (at the very last moment in the swing phase) become $\mb{f}^j_t$ and  $\mb F_t^j$ for the contact duration.
Action $\mb a^j_s$, which is output from the policy for the foot in the contact state, is ignored and not used.

\subsection{QP simulation}
\label{sec:simulation_qp}

The QP simulator takes the SRB character’s center of mass state $\mb T_t$, $\dot{\mb q}_t$, its relative desired velocity $\mb a_v$, and contact point $\mb{c}_t$ as inputs to calculate its internal state in the next time step.
To this end, the desired acceleration is calculated first based on the reference SRB motion and $\mb{a}_v$.
The actual achievable acceleration that maximally satisfies this desired acceleration and the corresponding contact force are calculated from the QP formulated as:
\begin{equation}
\label{eqn10}
    \min_{\ddot{\mb q}, \mb \lambda} Q \left(\ddot{\mb q}, \mb \lambda \right)
\end{equation}
such that
\begin{align}
\mathbf M \ddot{\mathbf q}+\mb {b} = \mb{J}_f^T\mb F_e+\mb{J}_c^T\mb{B}\mb{\lambda},\\
\mb \lambda \geq \mb 0.
\end{align}
Here, $\ddot{\mb{q}}$ \textcolor{rev}{:} the generalized center of mass acceleration, 
$\mb{\lambda}$ : the coefficient for the friction cone basis, 
$\mb F_e$ : the external force applied on the SRB character, 
$\mb{J}_c$ : the Jacobian matrices related to the velocity of the contact point $\mb{c}_t$, %
$\mb{J}_f$  : Jacobian matrices related to the velocity of the point where the external force is applied, and $\mb{B}$ : the friction cone basis vectors.
$\mb F_e$ is set to $\mb 0$ within the RL process, but is given non-zero values when external forces are applied during the runtime simulations. %
In short,
the acceleration and contact force that minimizes the objective function $Q$ are obtained while satisfying the equation of motion constraints and the constraints on the linearized basis of contact force \cite{kwon2017momentum,ellis2007cdm}.

The objective function $Q$ is defined as follows:
\begin{equation}
\label{eqn11}
	Q=\left\|\ddot{\mb{q}}-\ddot{\mb{q}}_d\right\|^2+ w_{\mb\lambda}\left\|\mb{\lambda}\right\|^2.
\end{equation}
Here, $\ddot{\mb{q}}_d$ refers to the desired acceleration.
$w_\mb{\lambda}$ refers to the weight for the contact force term, which was set to 0.001 to ensure robust control during our experiments.

Action $\mb a_v$ is used in the following to calculate the desired velocity $\dot{\mb{q}}_d$, through which the desired acceleration $\ddot{\mb{q}}_d$ is obtained.
\begin{equation}
\label{eq:desired_vel}
    \dot{\mb{q}}_d=\hat{\dot{\mb{q}}}_{\psi(t)}+\mb{a}_v ,
\end{equation}
where $\hat{\dot{\mb{q}}}_{\psi(t)}$ indicates the linear and angular velocities of the reference SRB motion’s center of mass, %
and action $\mb{a}_v$ refers to the desired relative linear and angular velocities of the SRB character’s center of mass against the reference SRB motion’s center of mass. %

 The desired acceleration $\ddot{\mb{q}}_d$ is calculated as follows by taking the difference between the desired and current position and velocity.
\begin{equation}
\label{eq:desired_acc}
	\ddot{\mb{q}}_d=a
	\log\left(
    \mb{T}_t^{-1}\hat{\mb{T}}_{\psi(t)}
	\right)
	+b\left(\dot{\mb{q}}_d-\dot{\mb{q}}_t\right),
\end{equation}
where $\mb{T}_t$ is the current SRB frame, while $\hat{\mb{T}}_{\psi(t)}$ is the reference SRB frame at the corresponding phase. 
$\dot{\mb{q}}_t$ is the current generalized velocity, while $\dot{\mb q}_d$ is the desired generalized velocity. 
The log function converts a rigid body transformation matrix $\in \mathrm{SE}(3)$ into a generalized velocity $\in \mathrm{se}(3)$.
In our experiments, PD gains $a$ and $b$ were set at 120 and 35, respectively.

By integrating the calculated $\Ddot{\mb{q}}$, $\mb{T}_t$ and $\dot{\mb{q}}_t$ are updated, and the foot state $\mb{f}^j_t$ and $\mb{F}^j_t$ are then updated as described in Section~\ref{sec:internal_state}.
The next time step’s state $\mb s_{t+1}$ can be derived from these values.
The simulation process through the QP solver helps the SRB character effectively find the contact force that best achieves the desired acceleration at each moment, helping to learn a robust policy.

\subsection{Motion phase adjustment}
\label{sec:motion_phase_adjustment}

At each SRB simulation timestep, the rate at which motion phase $\psi(t)$ changes is adjusted in two aspects.
First, the phase change rate is increased with increasing locomotion speed to prevent excessively long strides, which is inspired by \cite{kwon2017momentum}.
Second, when the character experiences a large unexpected force, it may deviate from the specified contact timings in the reference SRB motion, resulting in unstable control.
To address this, the phase rate is decreased to delay touchdown when it is expected to happen before the specified time, and increased to allow for an earlier touchdown when it is expected to happen later.
Further details of these adjustments are provided in the supplementary material.

\section{Full-body motion generation}
\label{sec:fullbody_mot_gen}

To generate a full-body motion from a simulated SRB character motion, we create a baseline SRB motion and calculate the kinematic and dynamic differences compared to the full-body reference motion.
During real-time simulation, these precomputed differences are applied to the simulated SRB motion, obtaining target values for momentum-mapped inverse kinematics (MMIK) to generate the full-body motion by solving MMIK.

\subsection{Baseline SRB motion construction}
\label{sec:baseline_motion}

\begin{figure}[h]
  \centering
  \includegraphics[width=1.0\linewidth]{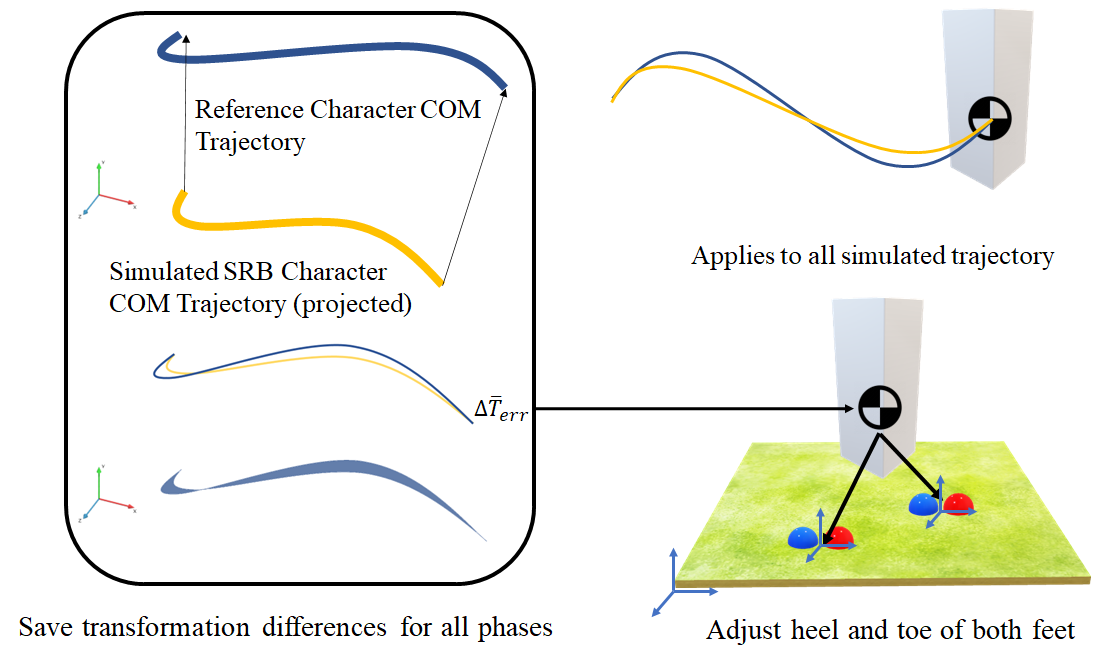}
  \caption{\label{fig:align_and_delta}
           Conceptual depiction of aligning COM trajectory and calculating COM frame delta.
            \textcolor{rev}{The frames on the plane are the center of each foot and blue and red balls are its heel and toe contact points.}
           }
\end{figure}

After obtaining a trained policy, it is used to simulate the SRB character to generate multiple simulated cycles, serving as the baseline SRB motion.
Then the COM trajectory and the feet states of the baseline SRB motion are transformed offline.
This transformation aligns the horizontal start and end points of the COM trajectory with those of the reference COM trajectory per motion cycle (Figure~\ref{fig:align_and_delta}), while ensuring no inclination of the trajectory by considering only the rotation against the y-axis.

\subsection{Delta computation} %
\label{sec:computing_delta}

The delta $\Delta$ between the baseline SRB motion and the full-body reference motion is computed to incorporate the reference full-body details into a reconstructed full-body motion during runtime. 
\textcolor{rev}{Our policy is capable of generating long SRB motions, but each cycle may exhibit slight variations.
To account for this, we take multiple simulated cycles of the baseline SRB motion for each reference clip and average the differences from the full-body reference motion over these cycles.
This averaging process is performed for each frame of every reference motion clip.
Specifically, the delta $\Delta_{\psi(t)}$ consists of the average differences in the center of mass frame, foot contact points, and centroidal velocity at the phase $\psi(t)$ of a reference motion clip.
Further details for computing $\Delta$ are described in the supplementary material.}

\subsection{Momentum-mapped inverse kinematics}
\label{sec:mmik}

During runtime, the simulated SRB motion is transformed into the full-body motion using momentum-mapped inverse kinematics (MMIK) presented in \cite{kwon2020fast}\textcolor{rev}{, which computes the fullbody pose to position its feet on planned SRB footstep locations while matching the planned SRB configuration and its time-derivative}.
We improve the previous form of MMIK to use the additional input of the precomputed delta $\Delta$.
The inputs are the full-body reference motion, $\Delta$, and simulated SRB states and it outputs the full-body pose closest to the given full-body reference pose, while satisfying the footstep position, COM configuration, and its time derivative of the error-adjusted simulated SRB motion using $\Delta$. 
Further details for \textcolor{rev}{MMIK} are described in the supplementary material.

\section{Experimental results}

\begin{figure}[h]
  \centering
    \vspace{-2\baselineskip}

    \subfigure[\textit{Walk}]{ %
                \includegraphics[width=0.47\columnwidth]{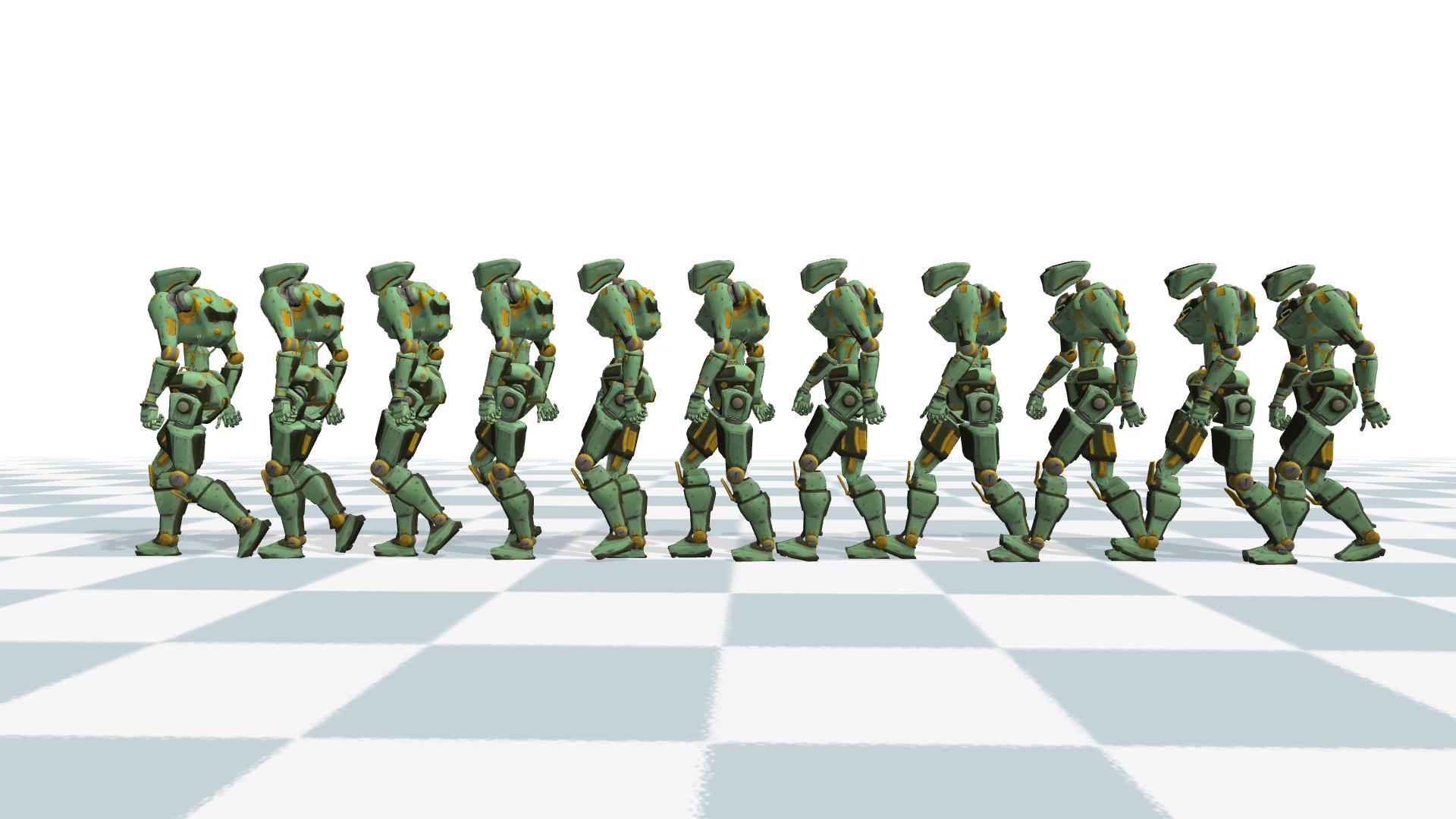}
                \label{fig:walk}
				}
    \subfigure[\textit{Fast walk}]{ %
                \includegraphics[width=0.47\columnwidth]{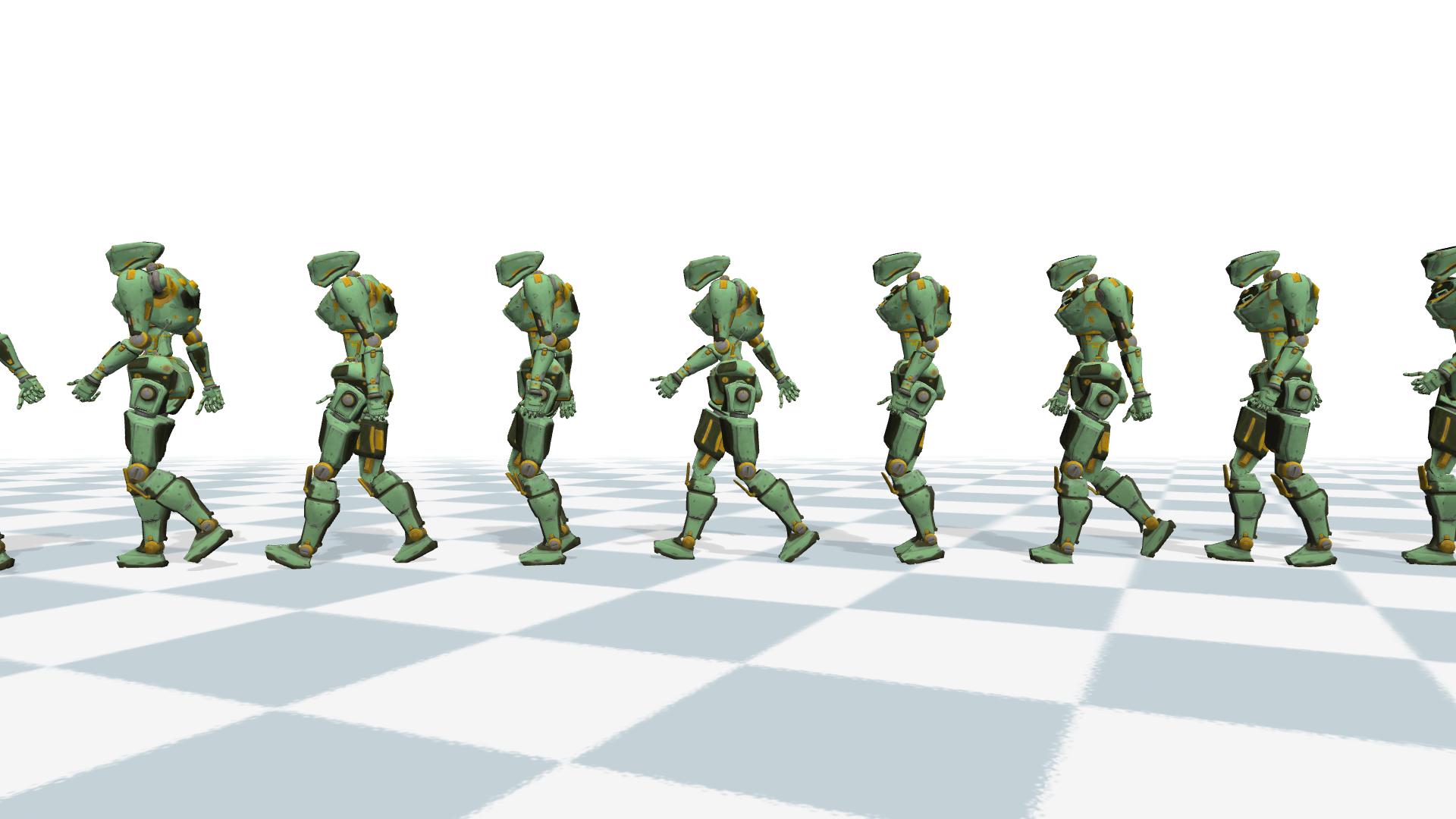}
                \label{fig:fastwalk}
				}
    \subfigure[\textit{Run}] { %
                \includegraphics[width=0.47\columnwidth]{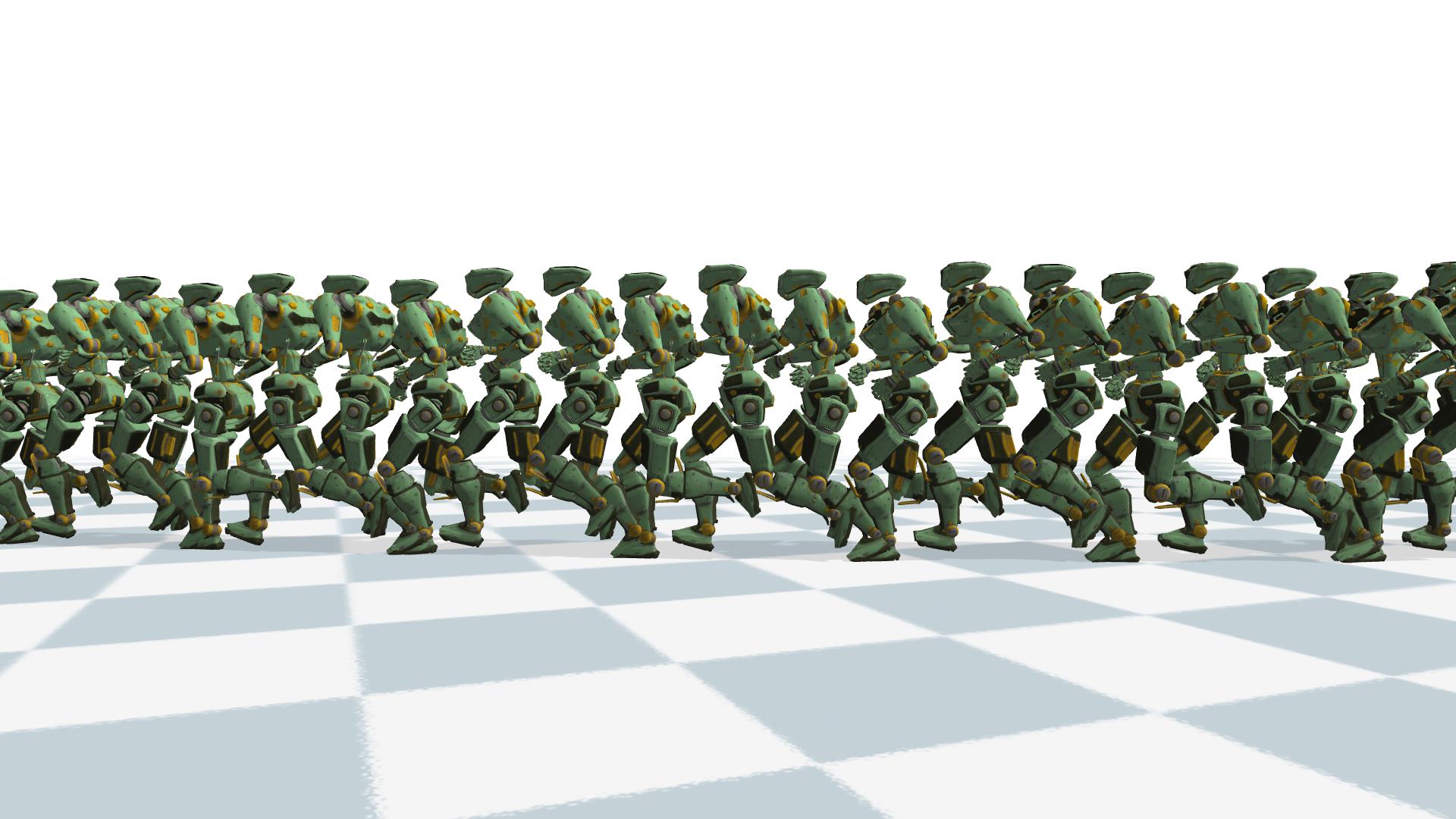}
                \label{fig:run}
				}
    \subfigure[\textit{Sprint}]{ %
                \includegraphics[width=0.47\columnwidth]{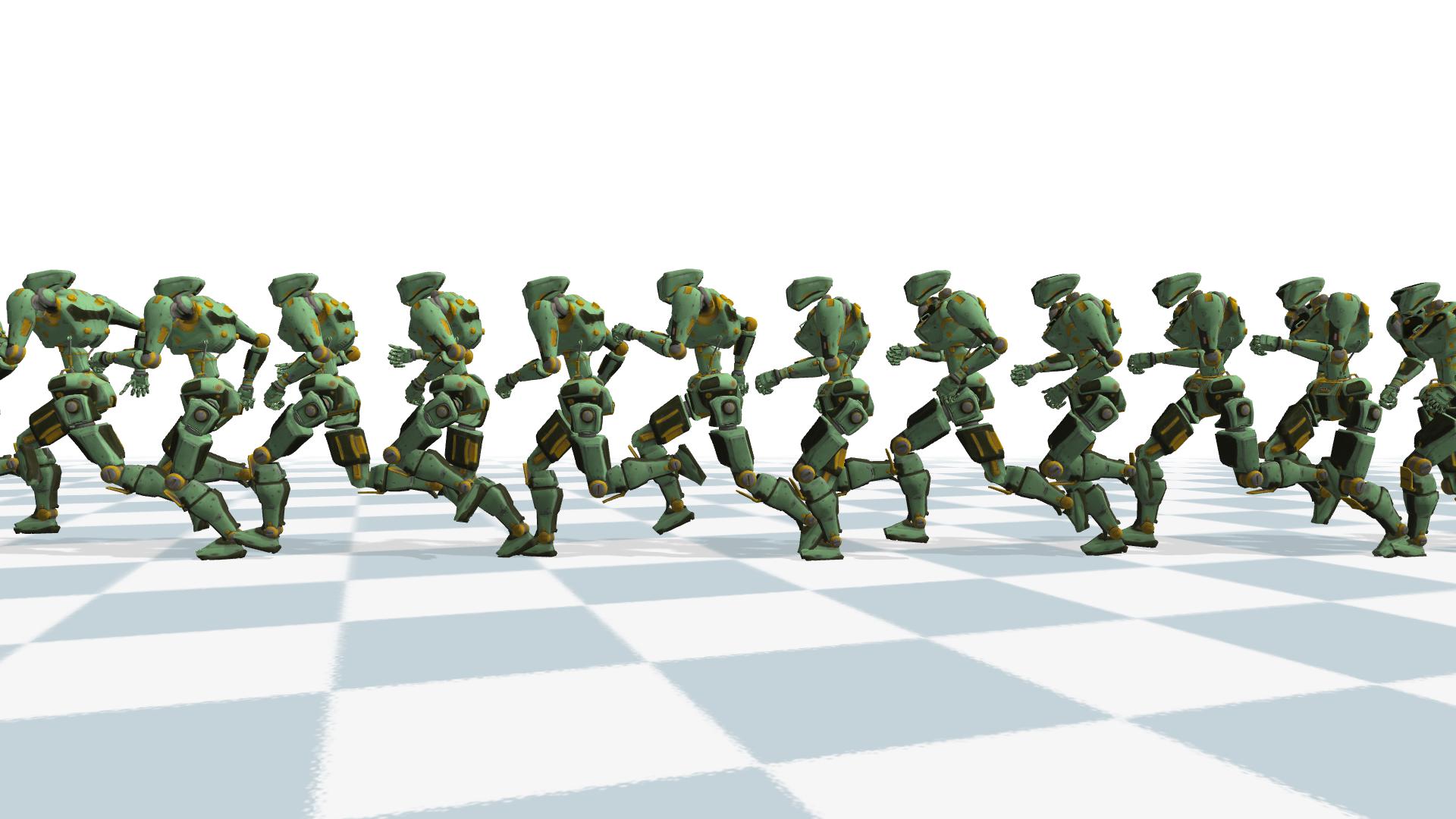}
                \label{fig:sprint}
				}
    \subfigure[\textit{Sprint jumps}]{ %
                \includegraphics[width=0.47\columnwidth]{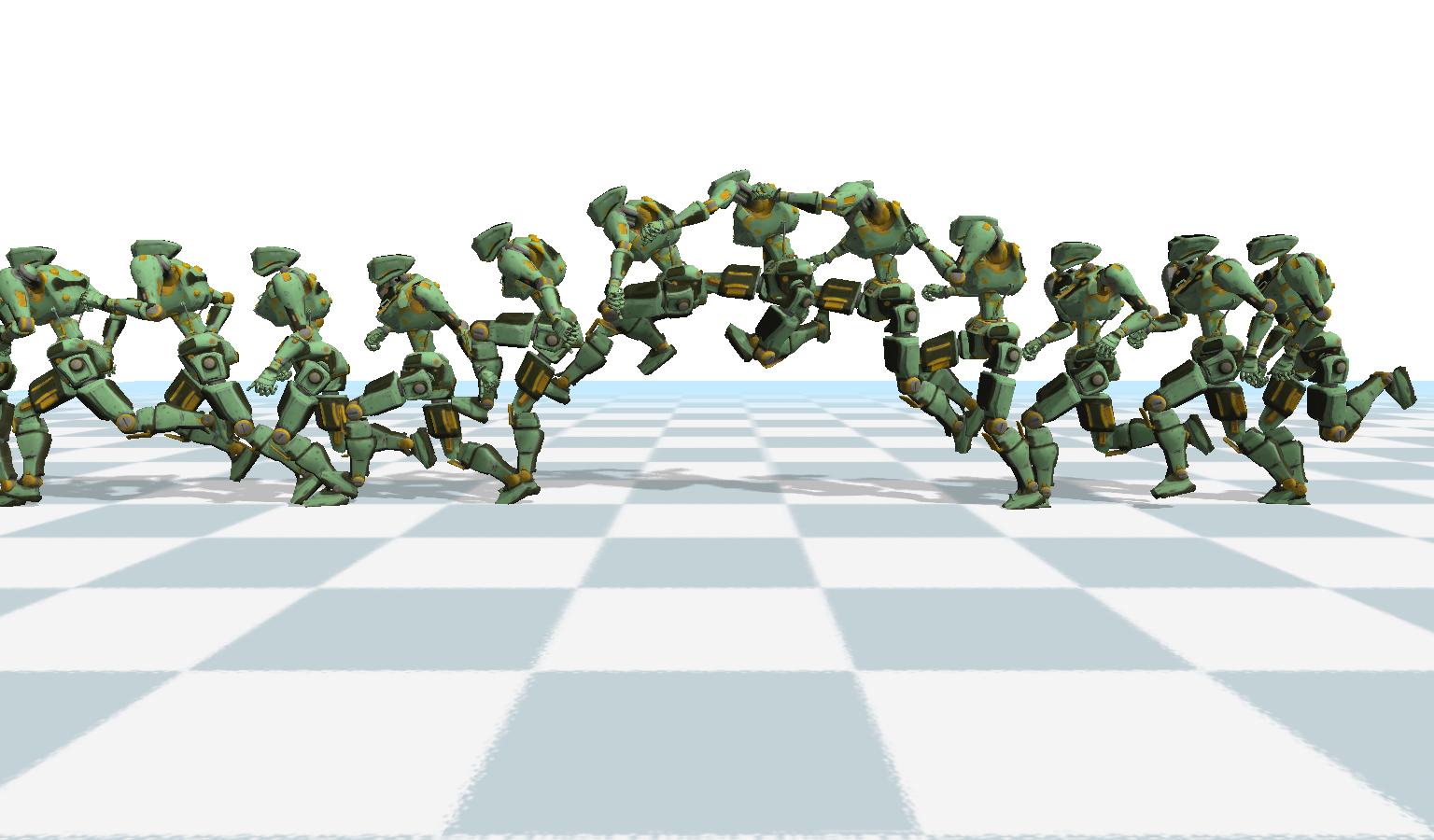}
                \label{fig:runjump}
				}
    \subfigure[\textit{Sharp turns} (180$^\circ$)]{
                \includegraphics[width=0.47\columnwidth]{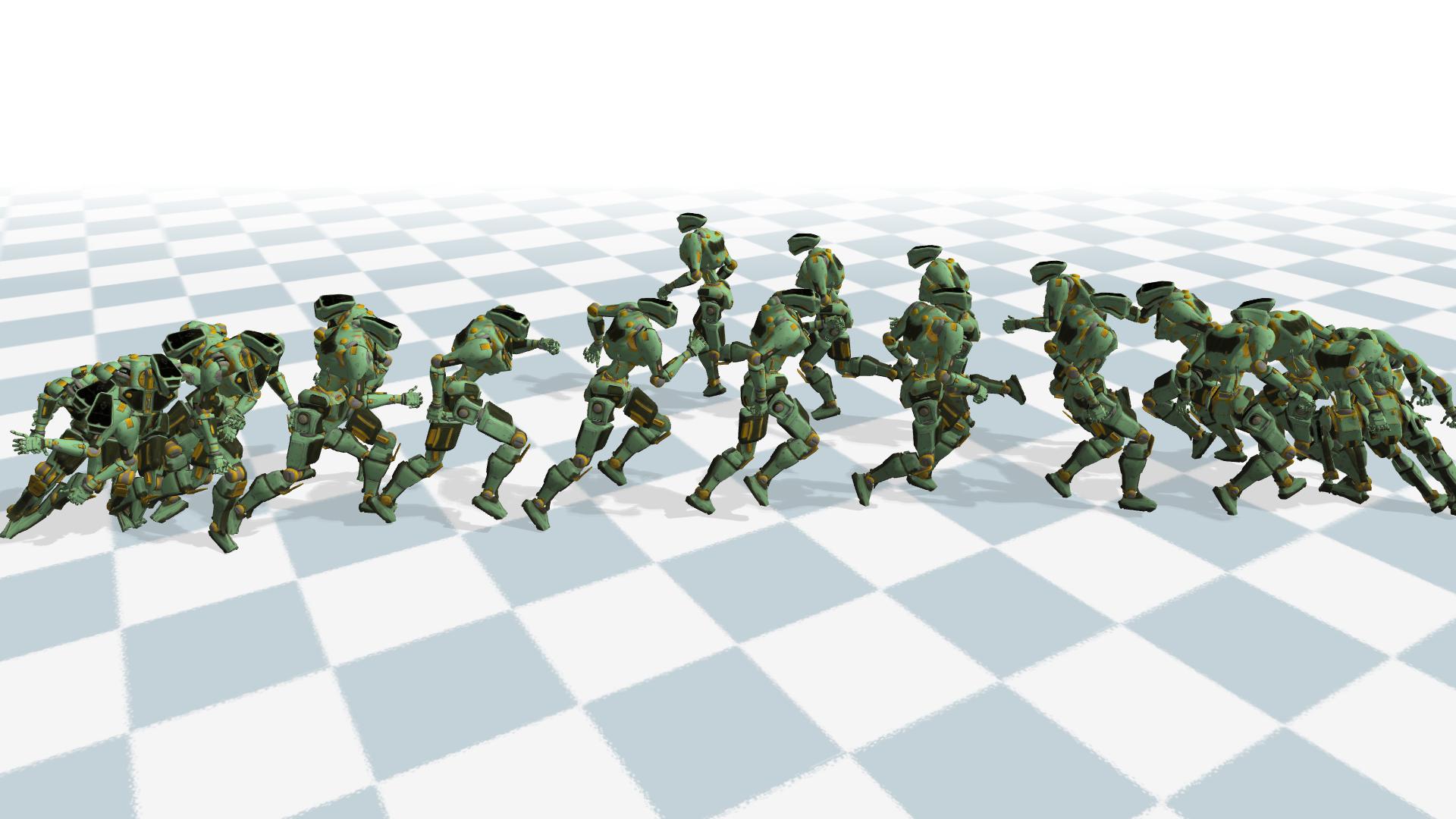}
                \label{fig:run180}
				}
    \subfigure[\textit{Sharp turns} (90$^\circ$)]{
                \includegraphics[width=0.47\columnwidth]{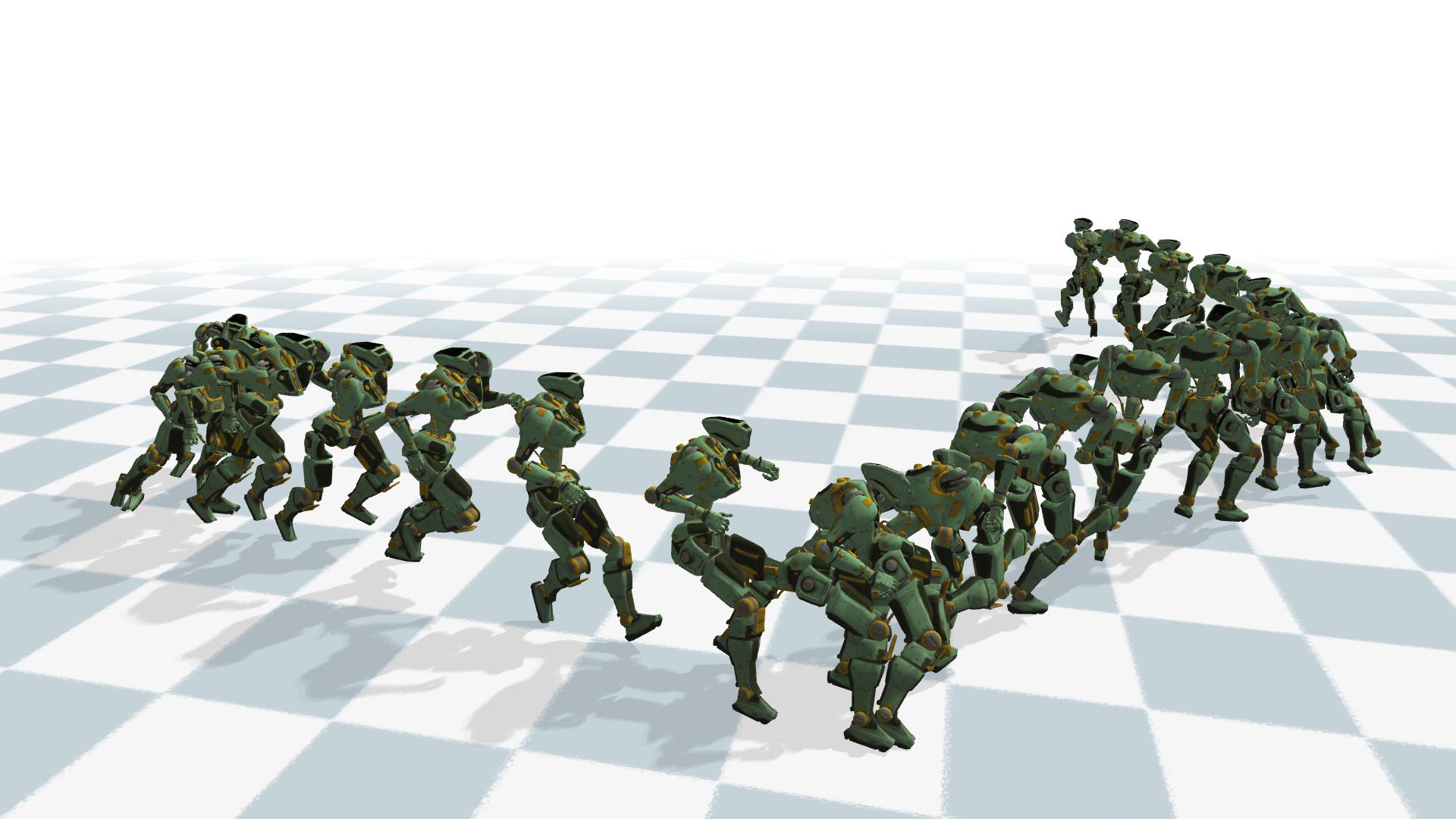}
                \label{fig:run90}
				}
    \subfigure[\textit{Backflips}]{
                \includegraphics[width=0.47\columnwidth]{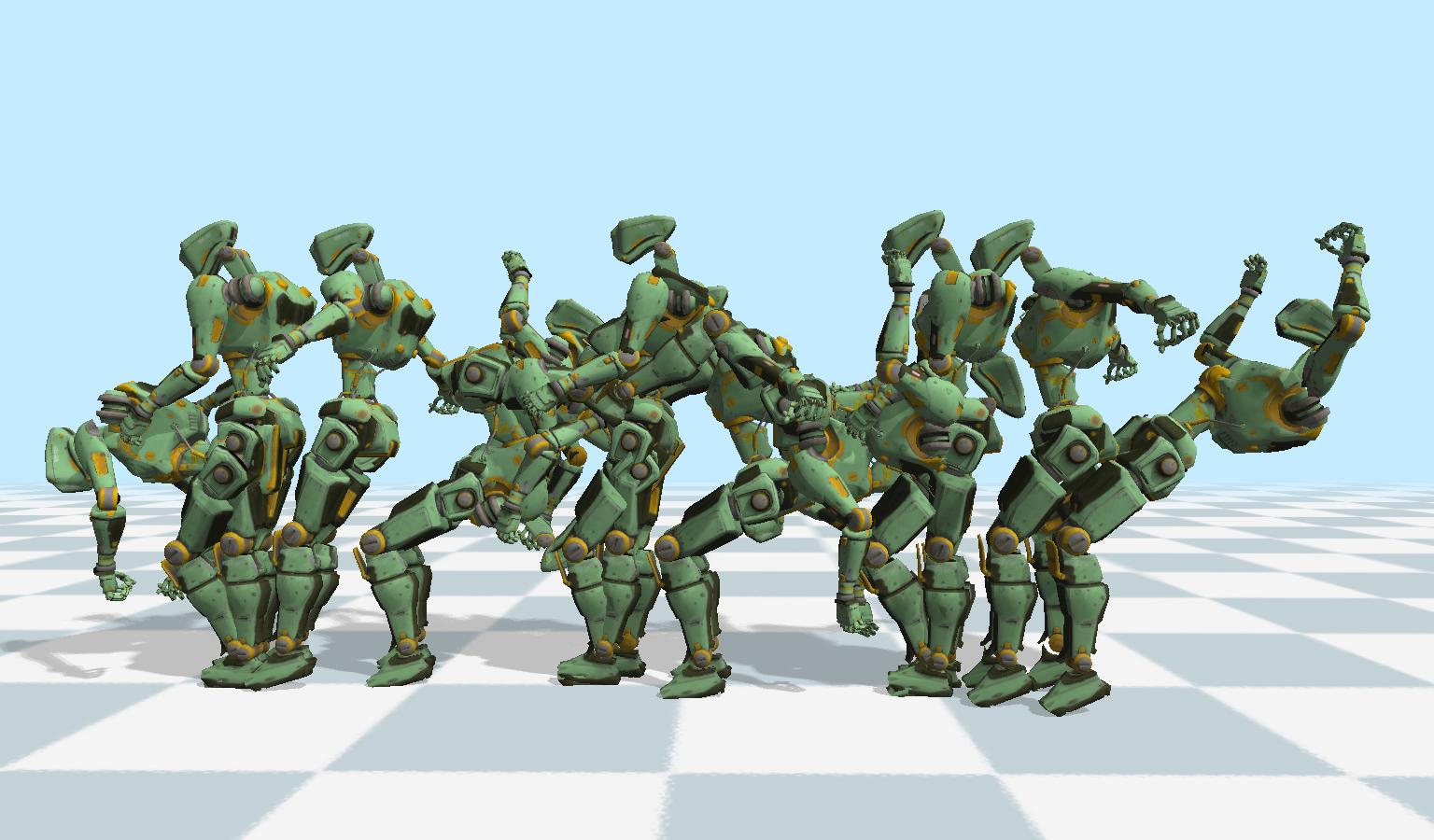}
                \label{fig:backflip}
                }

  \caption{\label{fig:various_motion}
           Tracking various motions.
		   }
\end{figure}

All experiments and policy training were conducted on a ultraportable laptop that runs on 16GB RAM, M1 CPU. %
All motions were simulated at \SI{60}{Hz}. 
Most controllers converged within 30 minutes, around 3M time steps, which is equivalent to about 14 hours of simulated time.
The most challenging \textit{Sprint jumps} controller converged within 1-hour, and
even after just 30 minutes, a policy performing the motion reliably was obtained.
Our approach enables the character to perform a range of reference motions,
including those involving drastic changes in direction, such as \textit{Sprint} or \textit{Sharp turns}, or those requiring strong force (Figure~\ref{fig:various_motion}).

We demonstrate how effectively our controllers respond to diverse, unobserved changes.
Note that all controllers used in the experiments were trained simply to mimic reference motions on flat ground (except Section~\ref{sec:results_interactive_control}), without any additional training for adapting to those changes.
In some experiments, we conducted comparisons with DeepMimic~\cite{peng2018deepmimic}.
DeepMimic has the advantage of simulating full-body dynamics, but it is designed to imitate detailed poses of a full-body reference motion, limiting its adaptability to unfamiliar situations.
As a result, DeepMimic exhibited less adaptability compared to our method.
However, it is important to note that DeepMimic and our method belong to different categories and have different purposes.
These comparative experiments were not intended to claim that our method is always superior to DeepMimic, but rather to showcase the advantages of our simplified physics-based approach compared to widely used full-body-based methods.
Further implementation and training details, and additional experimental results such as sample efficiency are provided in the supplementary material.

\subsection{Controller transitions and interpolations}

\begin{figure}[h]
  \centering

    \subfigure[\textit{Sprint} $\rightarrow$ \textit{Sprint jumps}]{ %
                \includegraphics[trim=0 300 0 250, clip, width=\columnwidth]{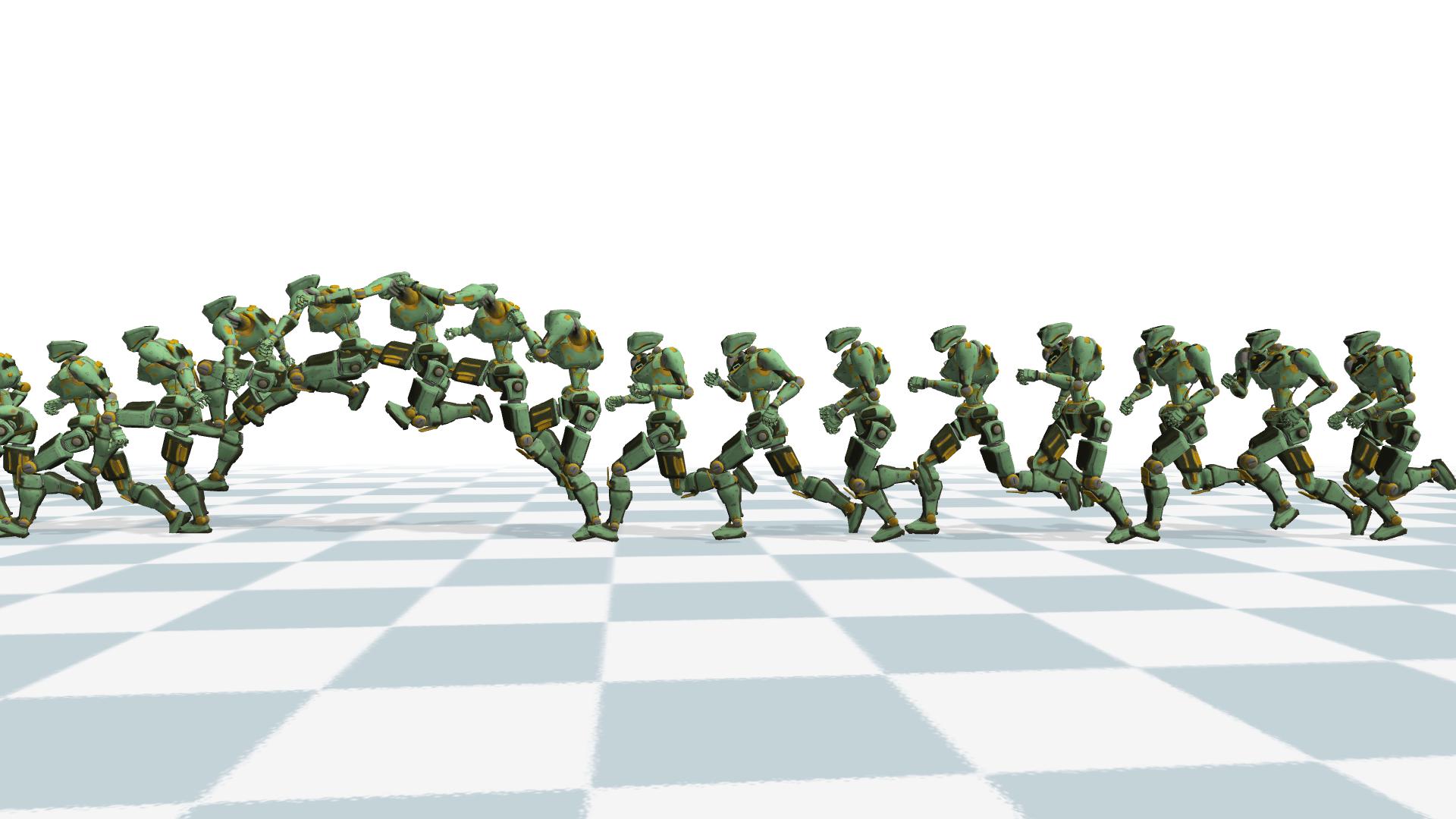}
                \label{fig:runjump}
				}
  
    \subfigure[\textit{Sharp turns} (180$^\circ$) $\rightarrow$ \textit{Sprint} ]{ %
                \includegraphics[trim=0 300 0 250, clip, width=\columnwidth]{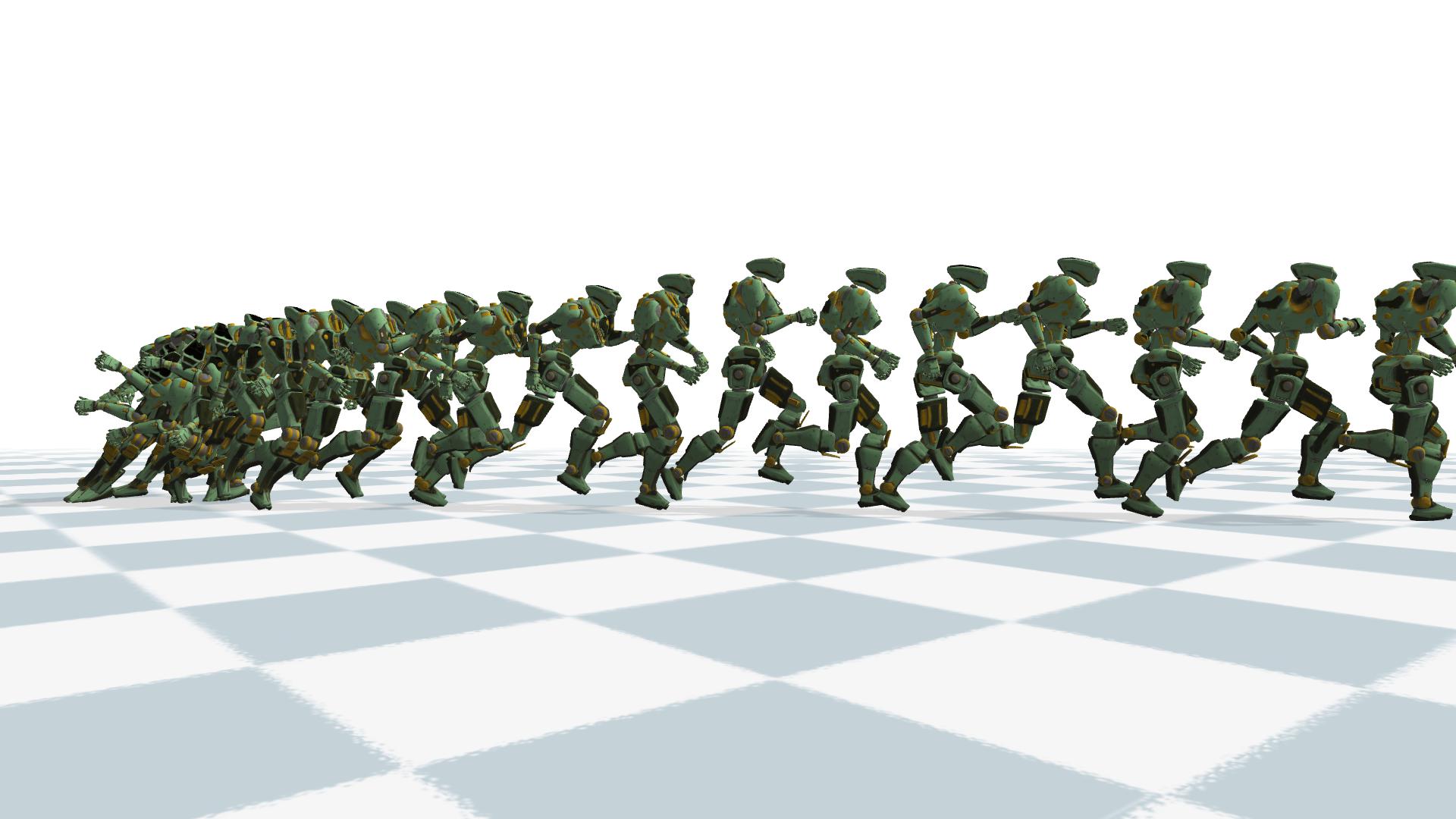}
                \label{fig:runjump}
				}
  
    \subfigure[\textit{Sharp turns} (-90$^\circ$) $\rightarrow$ \textit{Sprint} $\rightarrow$ \textit{Sharp turns} (90$^\circ$)]{ %
                \includegraphics[trim=0 300 0 250, clip, width=\columnwidth]{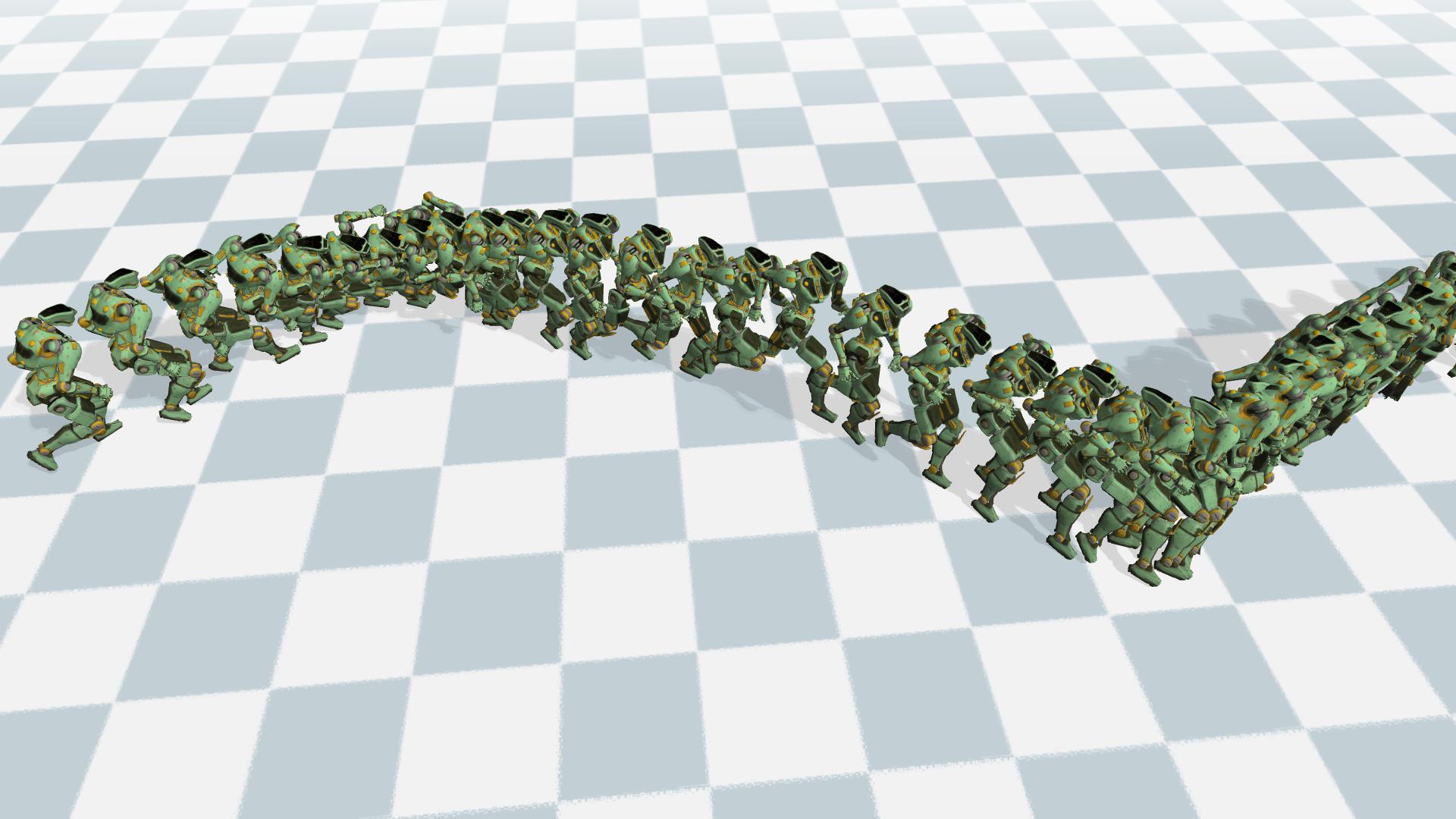}
                \label{fig:runjump}
				}

  \caption{\label{fig:transitions}
           Transitions by controller switching.
           }
\end{figure}

\textbf{Transitions by switching.}
Our method allows for transitioning between different controllers simply by switching pre-trained policies (Figure~\ref{fig:transitions}).
At the transition point, the policy $\pi_A$, reference SRB motion $\hat{\mb m}_A$, and precomputed delta $\Delta_A$, of the Controller A are immediately replaced with those of the Controller B. %
We use a motion stitching technique that gradually reduces the difference between the reference motions and deltas at the transition point and reflects it over time.
\textcolor{rev}{The transition point is predetermined as the moment when the contact state and posture are compatible.}
Note that the transition was successful (no falling within 20 seconds after the switch) when the difference in movement speed was within 30\%, such as \textit{Sprint} and \textit{Sprint jumps} or \textit{Sprint} and \textit{Sharp turns}, or when both controllers are exceptionally robust (e.g. \textit{Sprint} and \textit{Run}).

\begin{figure}
  \centering

    \includegraphics[trim=0 370 0 250, clip, width=\columnwidth]{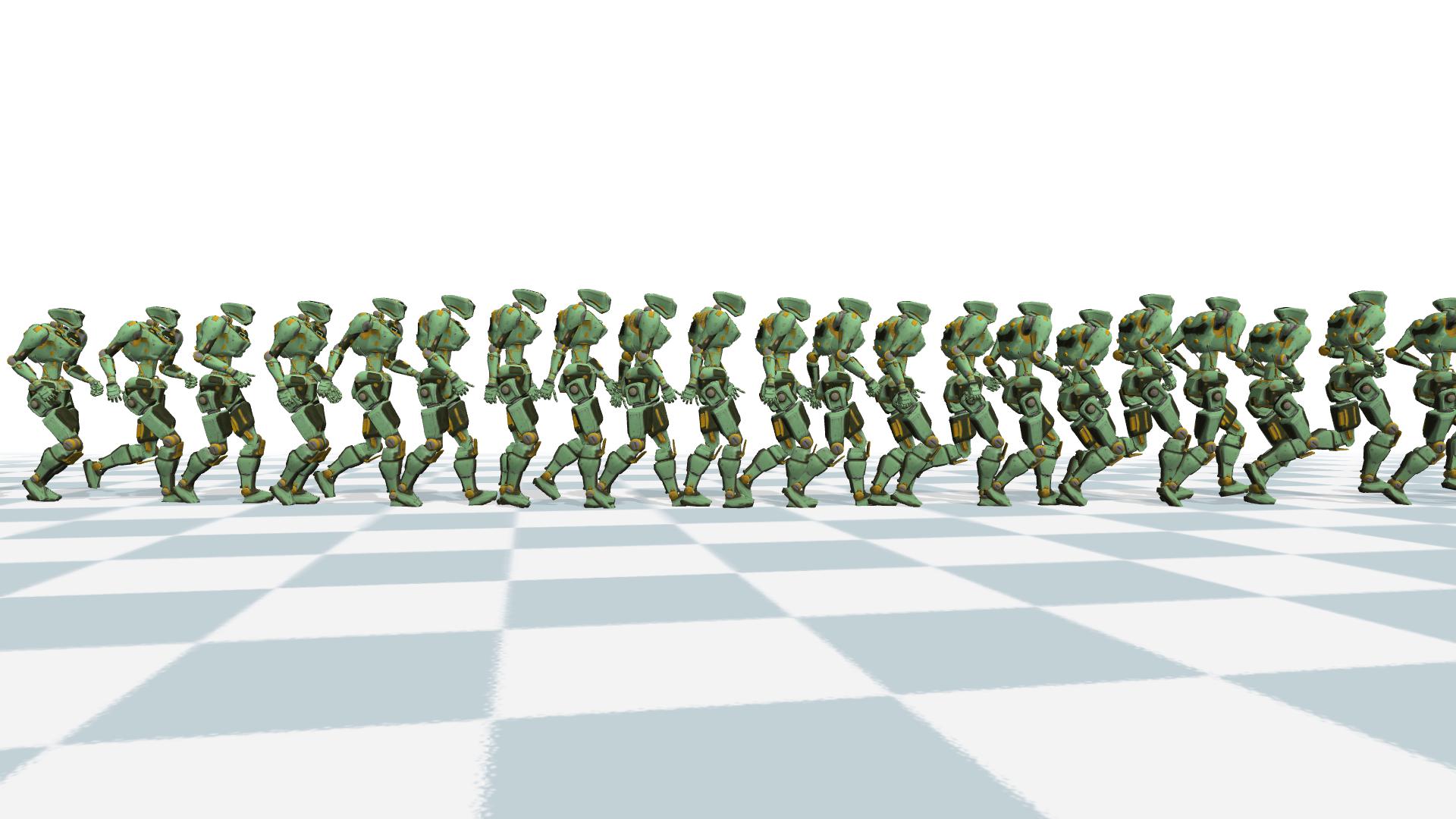}
      \caption{
          Blending of controllers. 
    Bi-directional transitions of motions are made between \textit{Fast walk} (\SI{1.8}{m/s}) and \textit{Run} (\SI{3.6}{m/s}) through the blended controller.
          }
    \label{fig:blending}
\end{figure}

\textbf{Transitions by blending.}
Transitions between controllers with larger speed differences can be achieved through blending, where Controller A gradually transitions to Controller B by linearly increasing the interpolation weight $t$ from 0 to 1 (for 1 sec).
The actions from $\pi_A$ and $\pi_B$, $\hat{\mb m}_A$ and $\hat{\mb m}_B$, $\Delta_A$ and $\Delta_B$, as well as the change rates of $\psi_A(t)$ and $\psi_B(t)$, the remaining time in $\hat{\mb m}_A$ and $\hat{\mb m}_B$ until each $j^{th}$ foot touches down $t_A^{d,j}$, $t_B^{d ,j}$ and until \textcolor{rev}{lifts} off $t_A^{o,j}$, $t_B^{o,j}$ are interpolated.
\textcolor{rev}{The phase or contact times do not have to be precisely aligned during a blended transition because $t_A^{d,j}$, $t_B^{d ,j}$, $t_A^{o,j}$, and $t_B^{o,j}$ are expressed as signed distances so that interpolation is possible when the contact states of Controller A and Controller B are different.
The contact state changes \textcolor{rev}{(touch-down or lift-off)} in the blended controller occurs at the point when the interpolated distance becomes 0.}
With this method, smooth transitions were achieved between controllers with significant differences in movement speeds and styles,
such as \textit{Fast walk} (\SI{1.8}{m/s}) and \textit{Run} (\SI{3.6}{m/s}) (Figure~\ref{fig:blending}).

\begin{figure}[h]
  \centering
    \subfigure[\textit{Run} controller ($t=0$, \SI{3.6}{m/s})]{ %
                \includegraphics[trim=0 350 0 370, clip, width=\columnwidth]{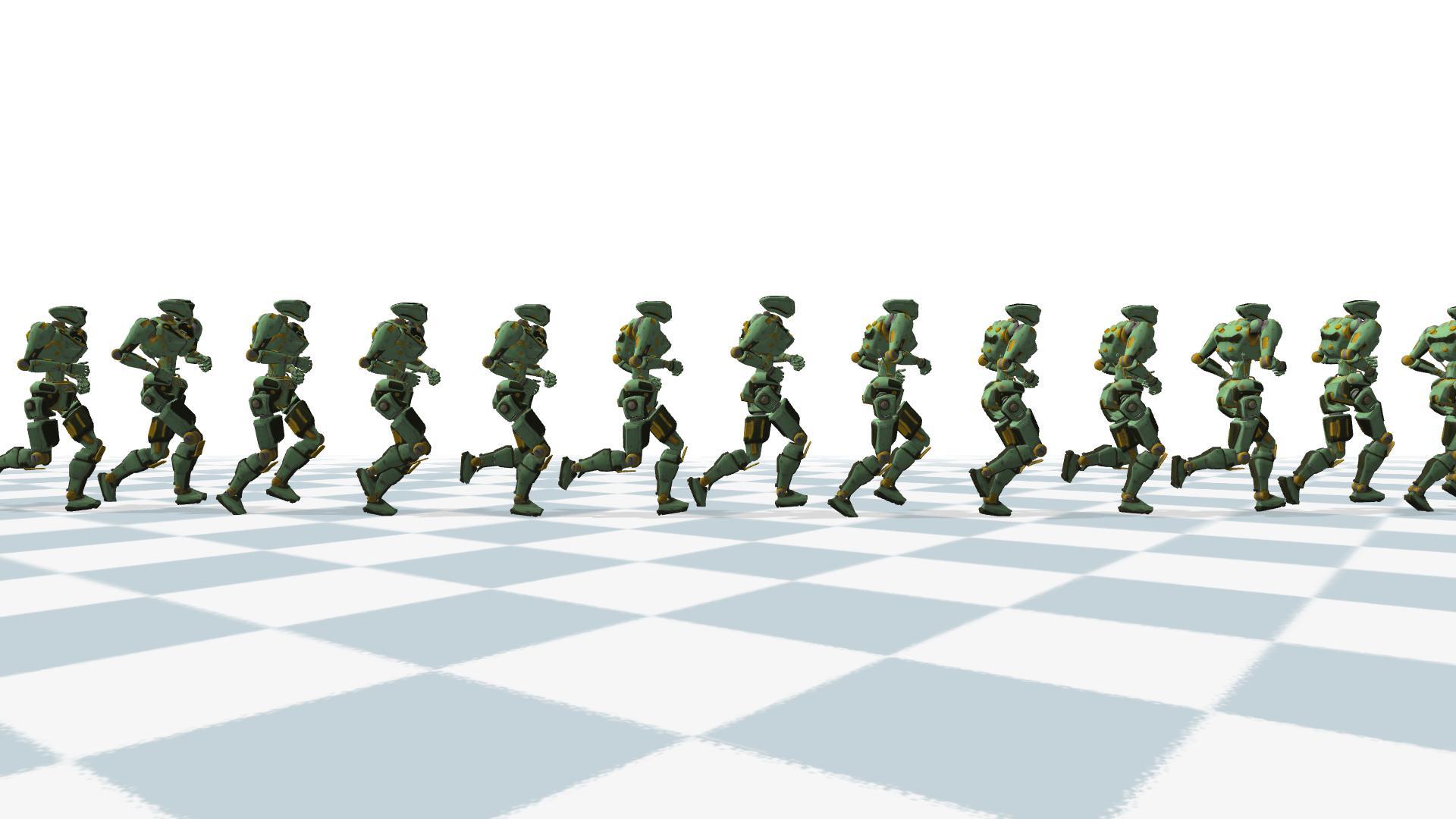}
                \label{fig:runjump}
				}
  \subfigure[Interpolated controller 1 ($t=0.25$, \SI{4.2}{m/s})]{ %
                \includegraphics[trim=0 350 0 370, clip, width=\columnwidth]{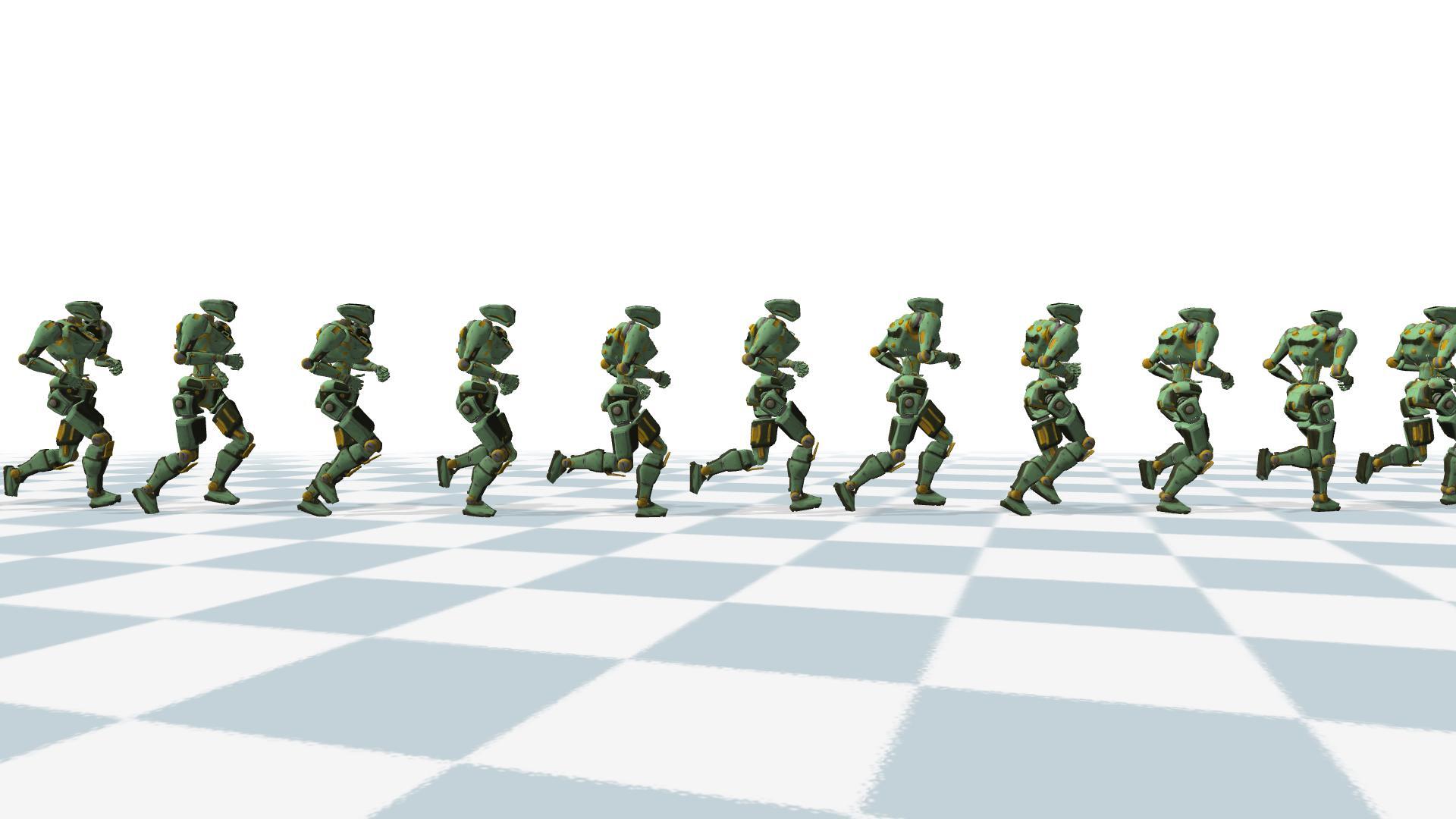}
                \label{fig:runjump}
				}
  \subfigure[Interpolated controller 2 ($t=0.5$, \SI{4.9}{m/s})]{ %
                \includegraphics[trim=0 350 0 370, clip, width=\columnwidth]{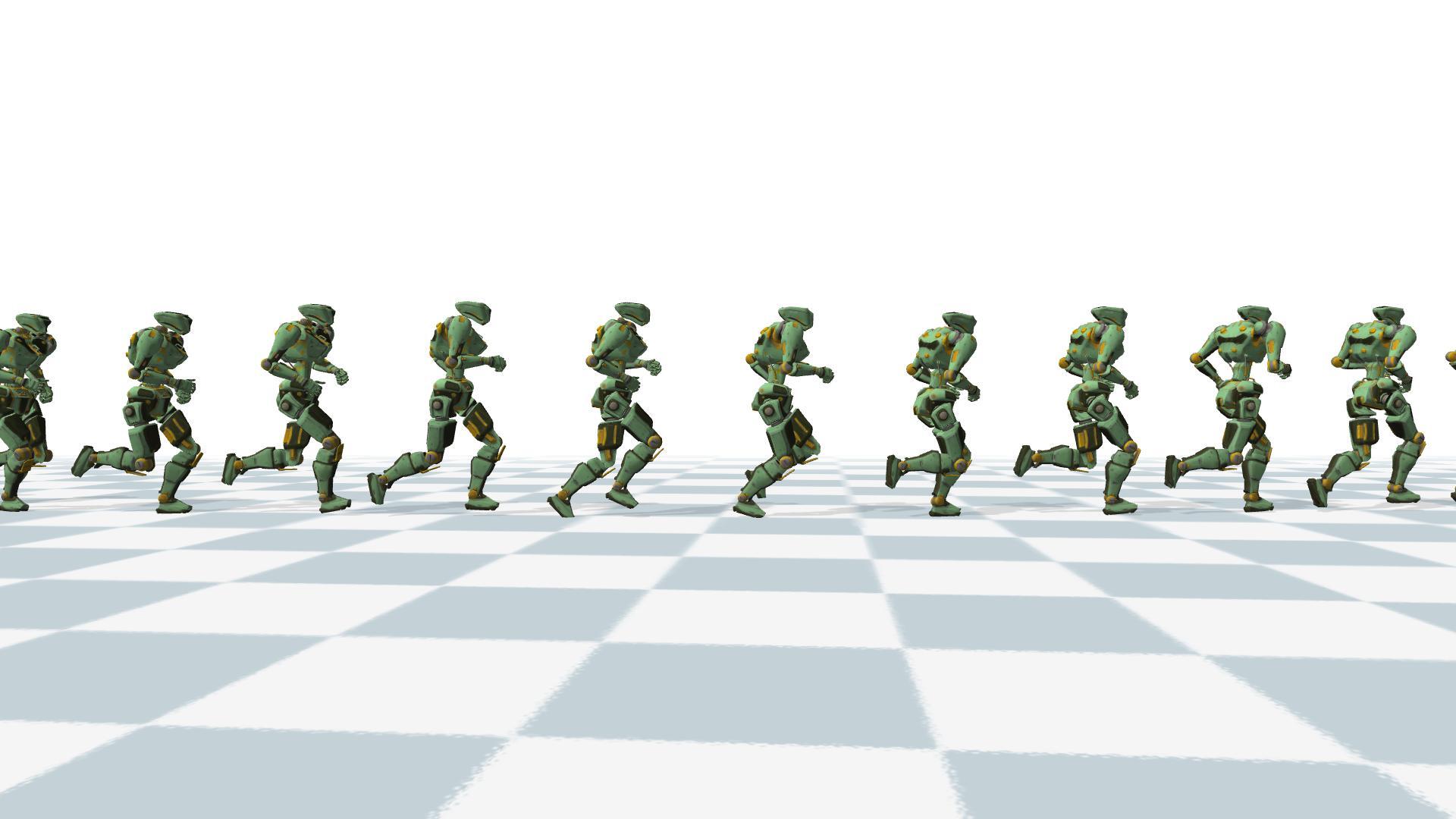}
                \label{fig:runjump}
				}
  \subfigure[Interpolated controller 3 ($t=0.75$, \SI{5.3}{m/s})]{ %
                \includegraphics[trim=0 350 0 370, clip, width=\columnwidth]{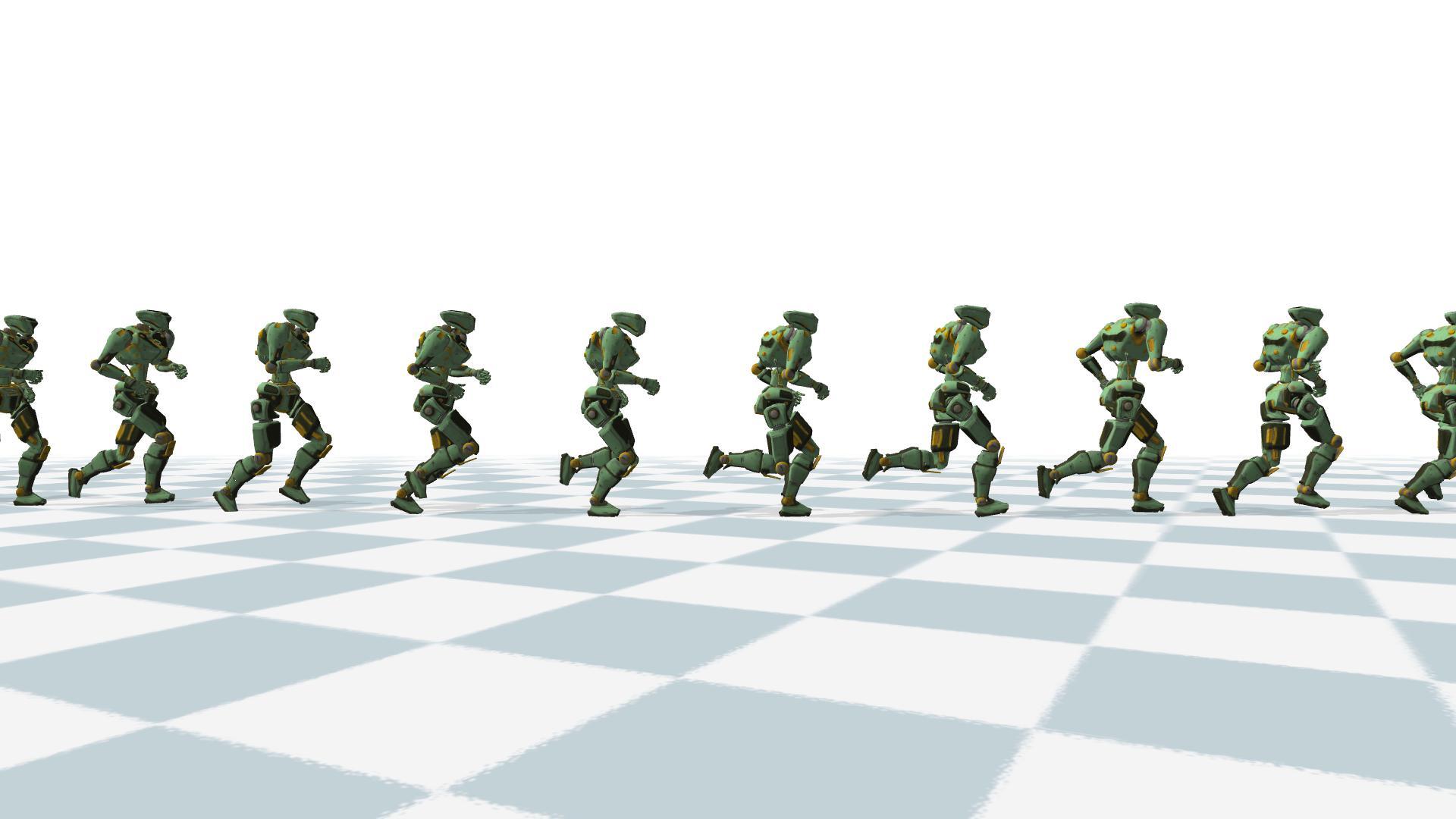}
                \label{fig:runjump}
				}
    \subfigure[\textit{Sprint} controller ($t=1$, \SI{6.0}{m/s})]{ %
                \includegraphics[trim=0 350 0 370, clip, width=\columnwidth]{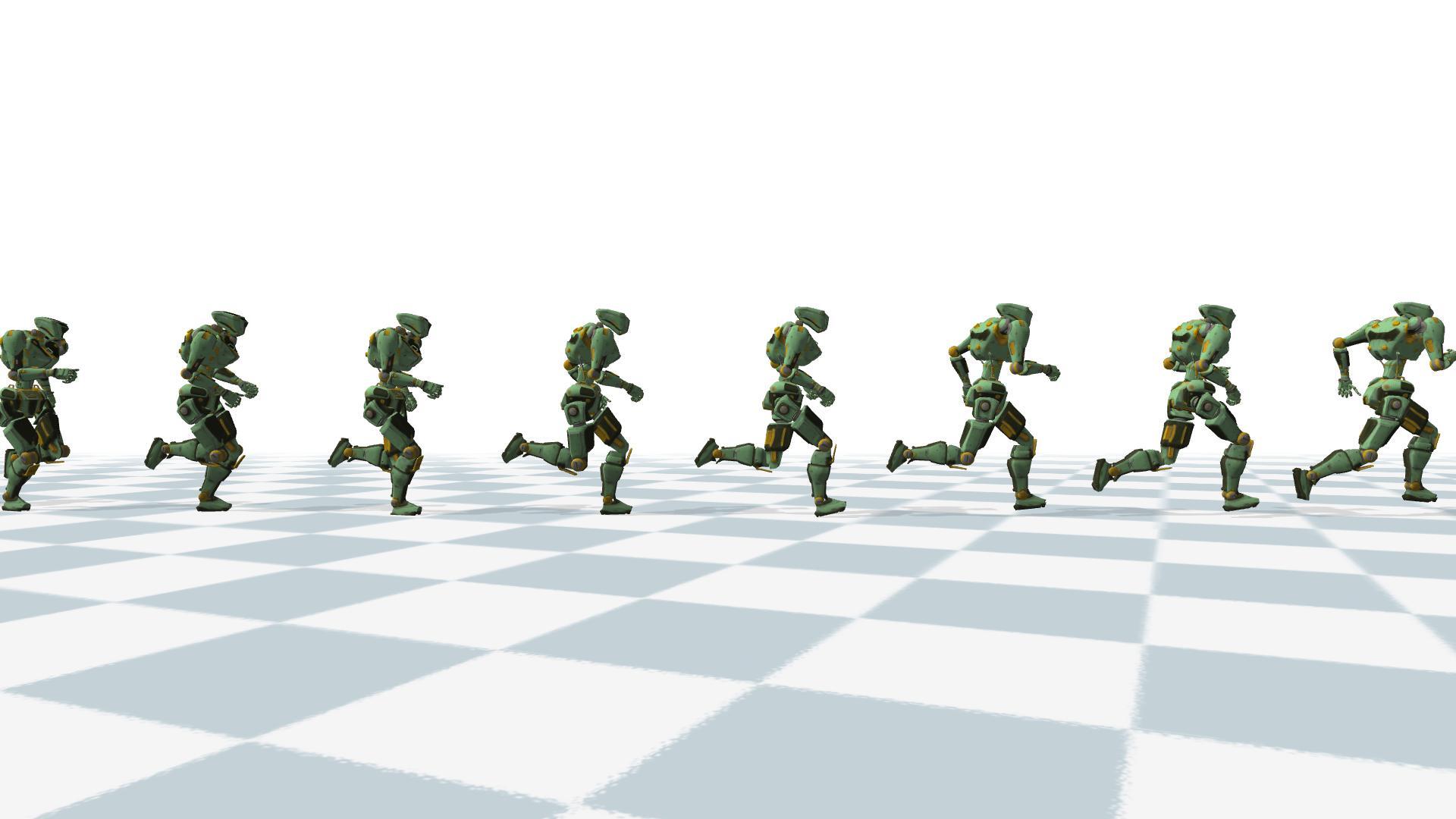}
                \label{fig:runjump}
				}
  
  \caption{\label{fig:interpolation}
           Interpolated controllers.
           This figure shows the simulated motions from \textit{Run} and \textit{Sprint} controllers and the interpolated controllers between the two.
The speed in parentheses is the moving speed of the simulated character.
           }
\end{figure}

\textbf{Interpolation.}
The interpolation method allows for creating new interpolated controllers at constant ratios between two controllers, apart from transitioning between two controllers (Figure~\ref{fig:interpolation}).

\subsection{Environmental adaptations}

\begin{figure}[b]
  \centering
    \subfigure[\textit{Walk} on uneven terrain]{
	  \includegraphics[trim=0 360 60 240, clip, width=\columnwidth]{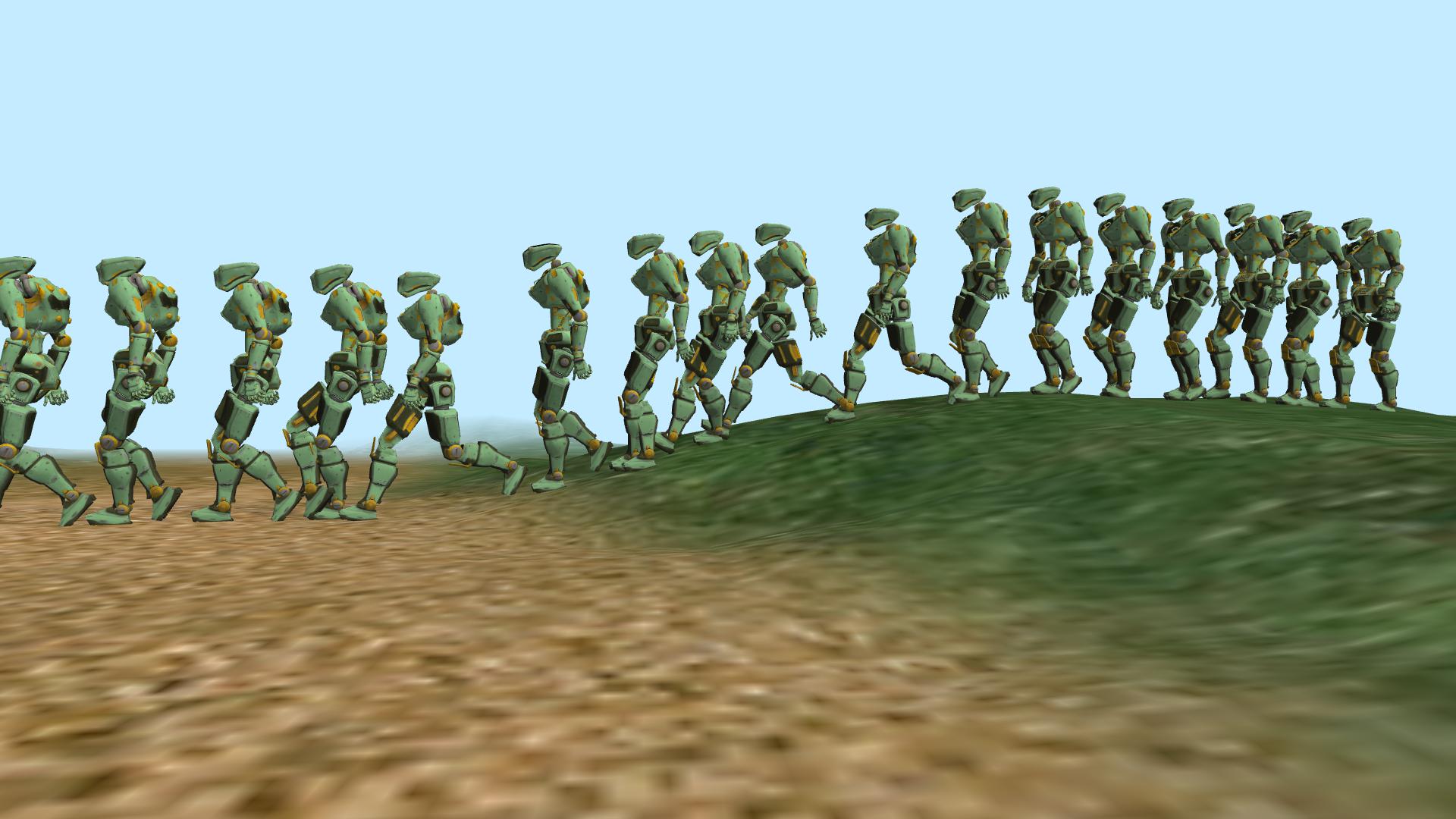}
				}
  \subfigure[\textit{Sprint} on uneven terrain]{
	  \includegraphics[trim=60 320 0 280, clip, width=\columnwidth]{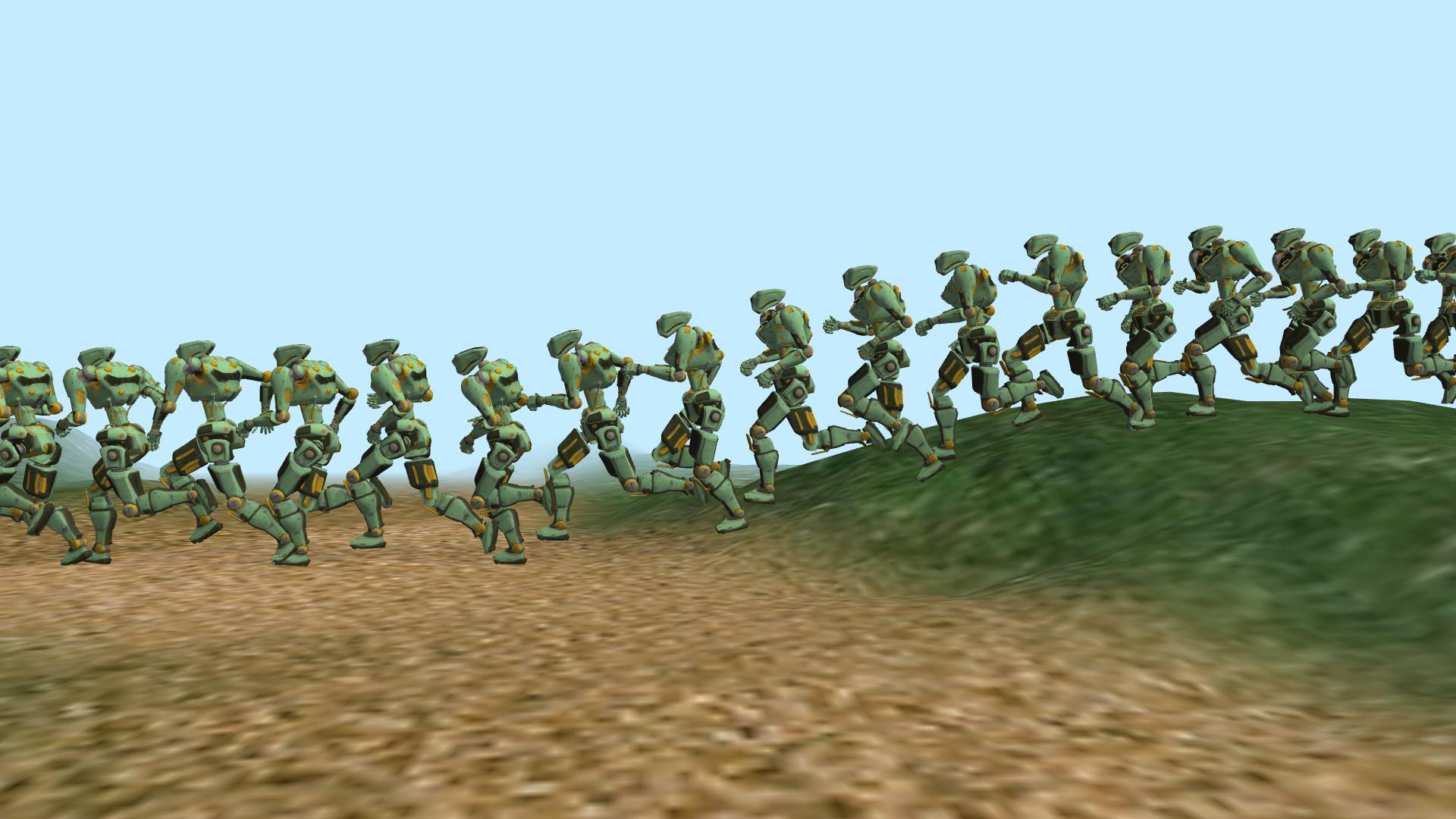}
				}
  \caption{\label{fig:terrain_adaptation}
    Adaptation for uneven terrain.
		   }
\end{figure}

\textbf{Uneven terrain.}
Our controllers trained on even terrain was able to create locomotion on uneven terrain with a gradient of a maximum 30 degrees (Figure~\ref{fig:terrain_adaptation}).
This is achieved without the need for any additional modifications to the algorithm, and the results are based on the inherent adaptability of the SRB-based policy.
In contrast, DeepMimic was observed to struggle even on gentle slopes, as it focuses on tracking full-body poses, resulting in the swing foot tripping.

\begin{figure}[t]
  \centering
    \subfigure[\SI{10}{kg}]{
	  \includegraphics[trim=550 50 450 250, clip, width=0.25\columnwidth]{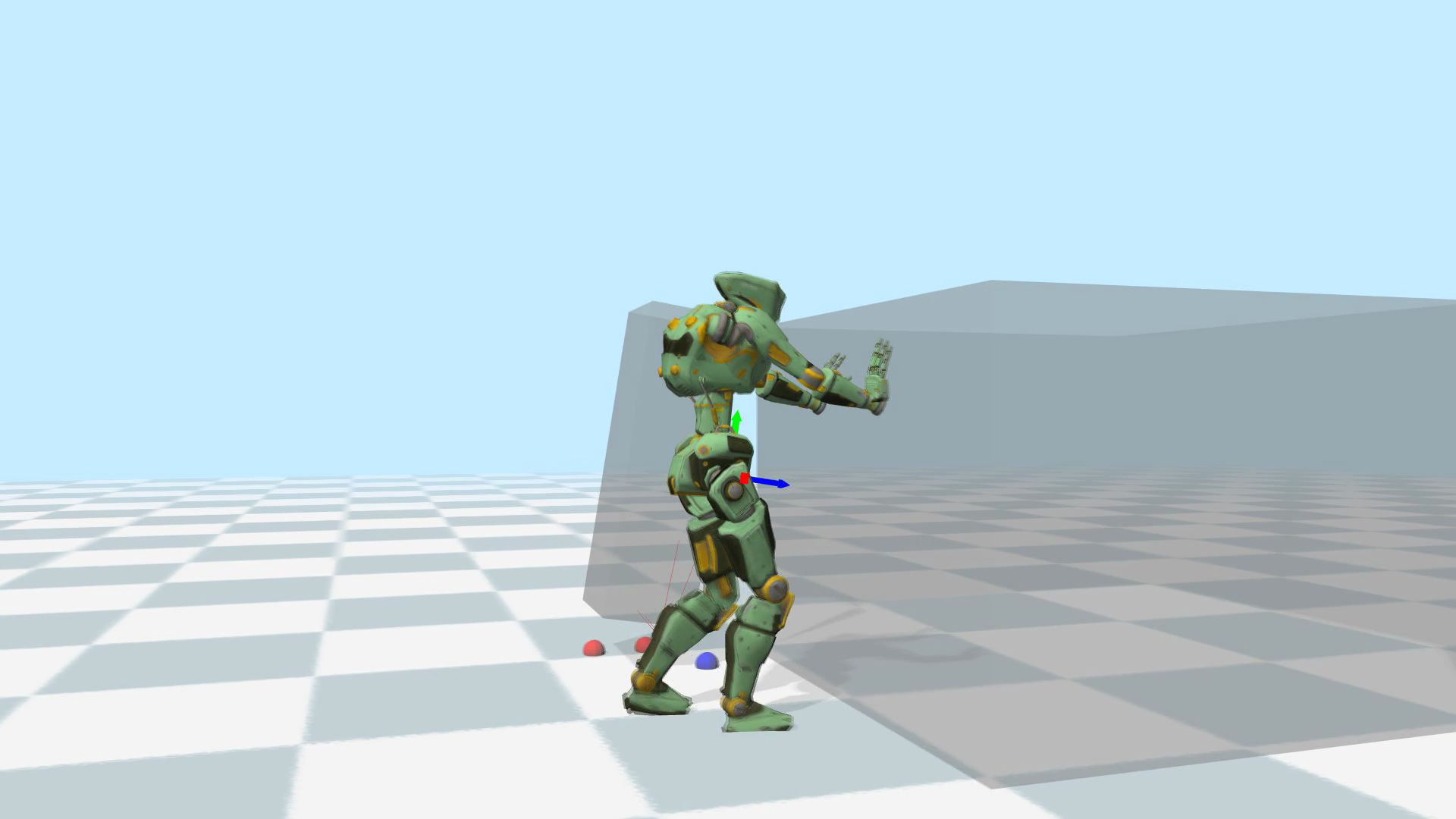}
	  \includegraphics[trim=550 50 450 250, clip, width=0.25\columnwidth]{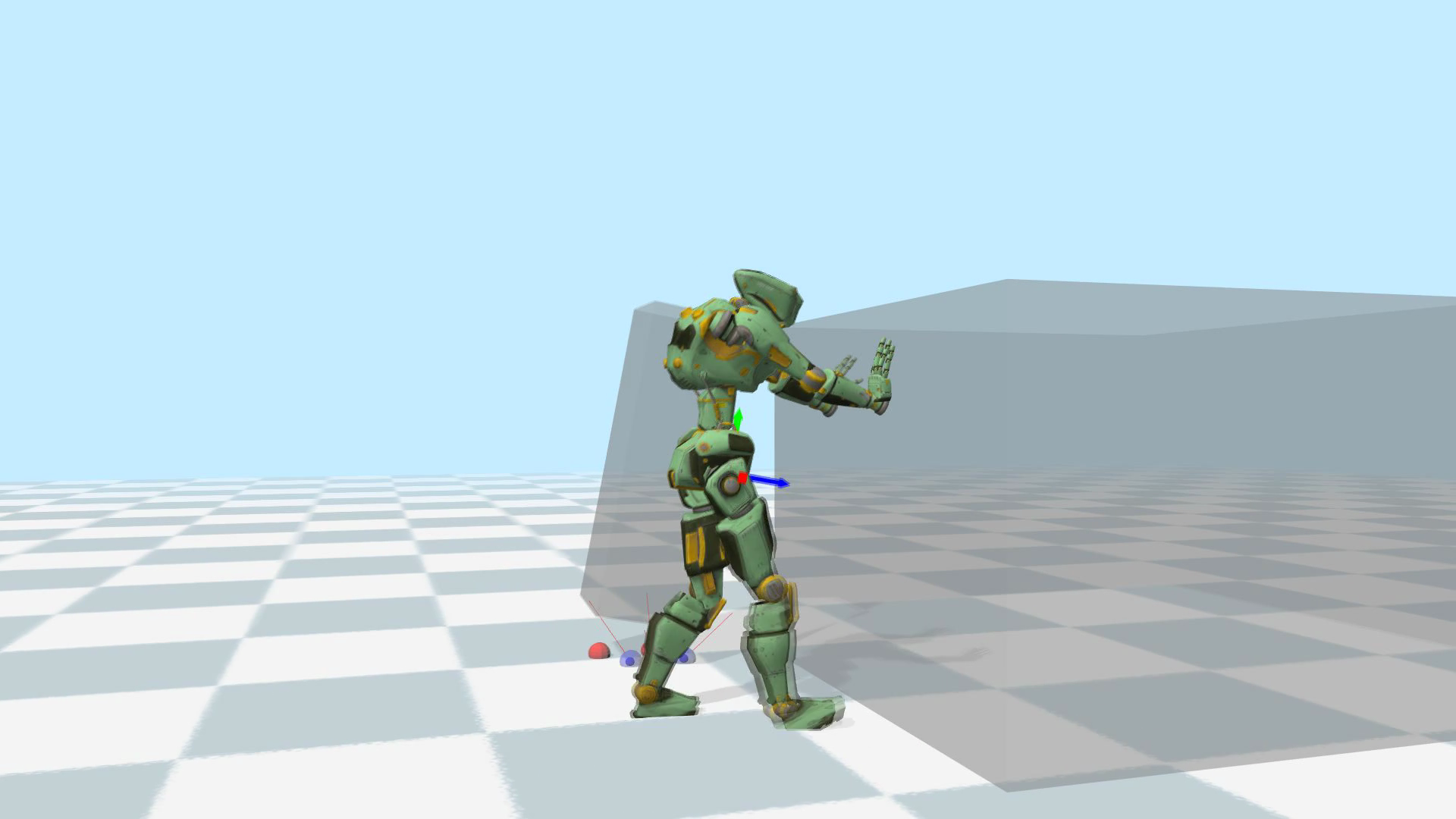}
	  \includegraphics[trim=550 50 450 250, clip, width=0.25\columnwidth]{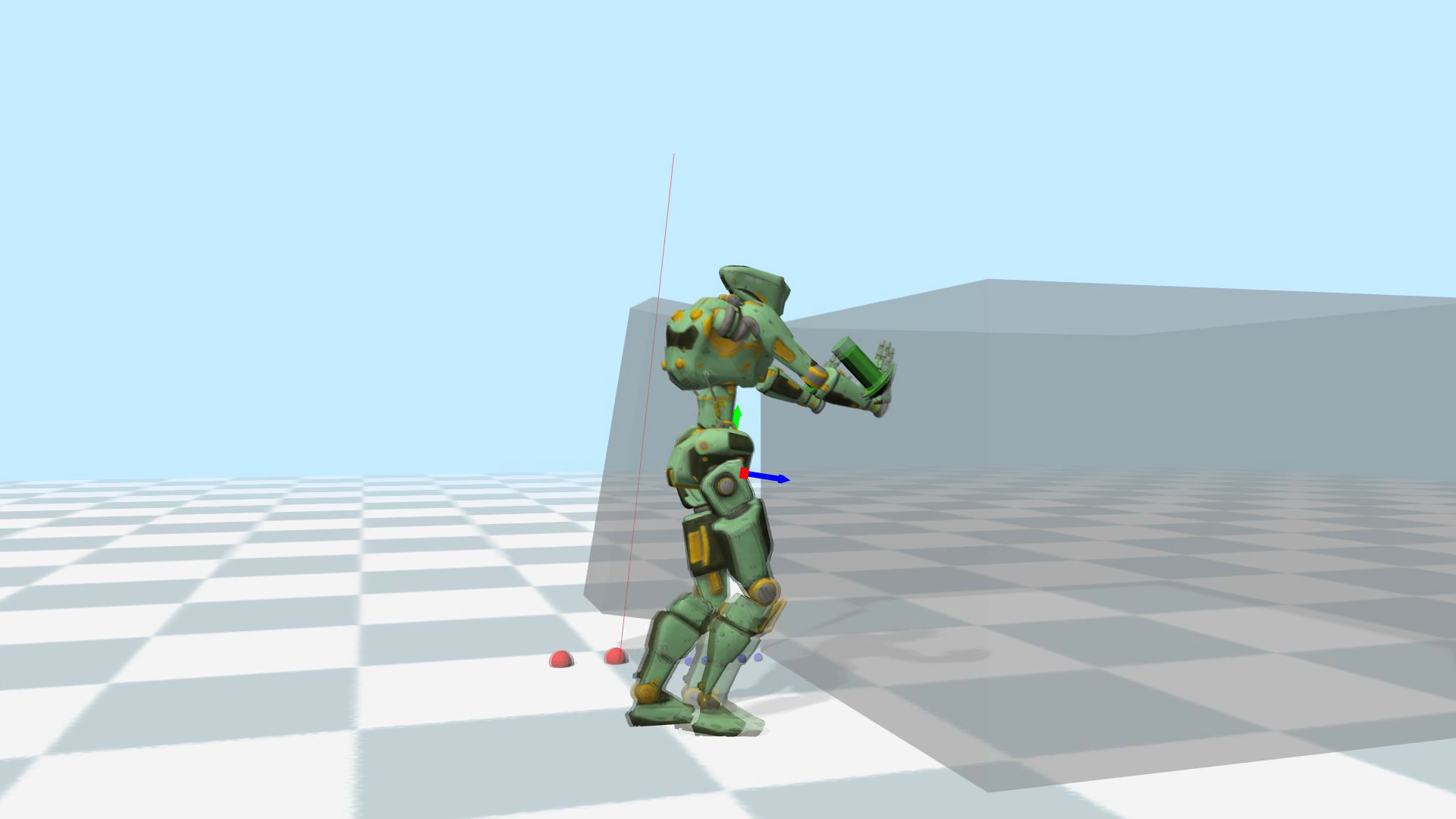}
	  \includegraphics[trim=550 50 450 250, clip, width=0.25\columnwidth]{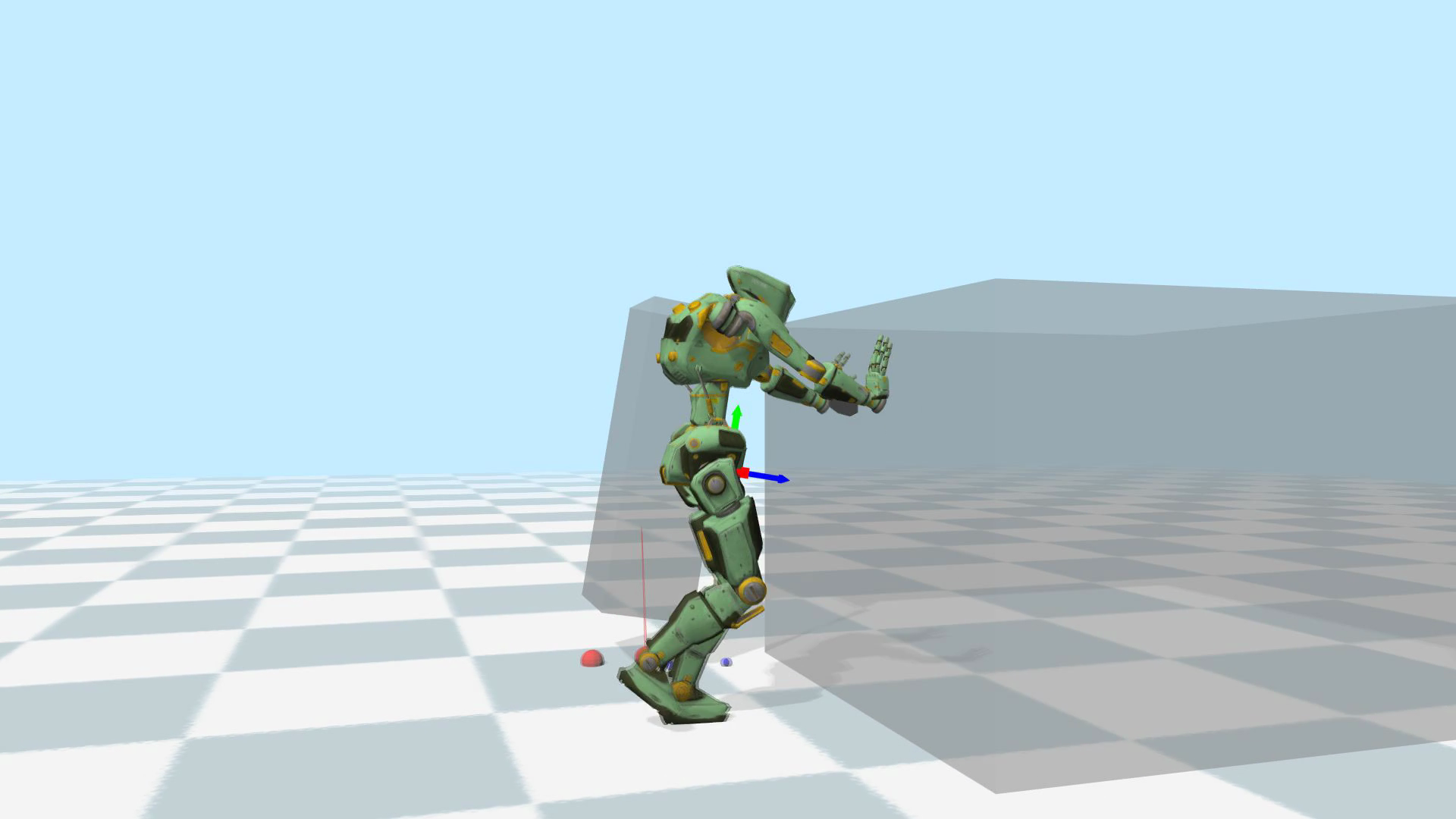}
        \label{fig:push_box_a}
				}
    \subfigure[\SI{30}{kg}]{
	  \includegraphics[trim=550 50 450 250, clip, width=0.25\columnwidth]{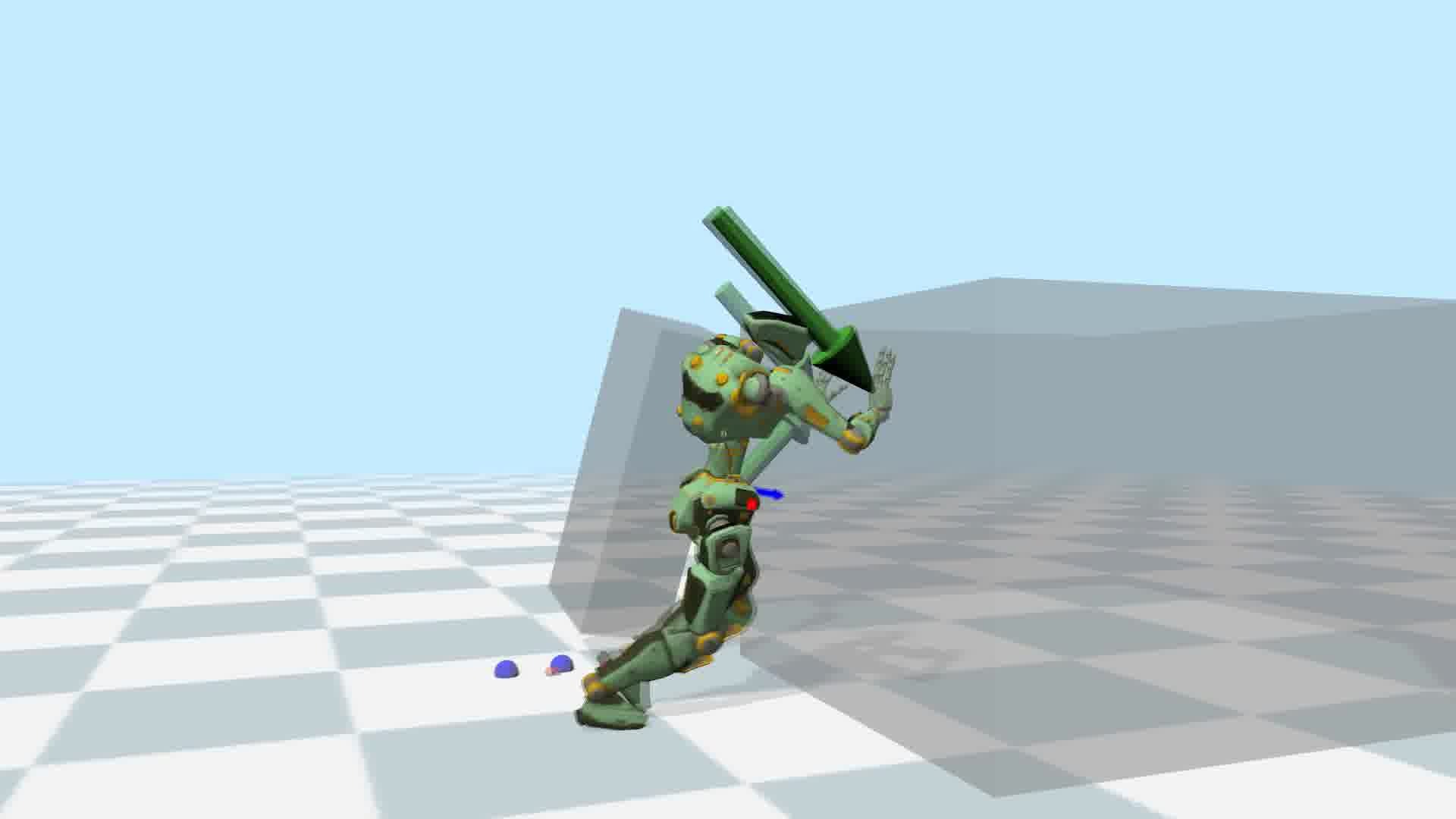}
	  \includegraphics[trim=550 50 450 250, clip, width=0.25\columnwidth]{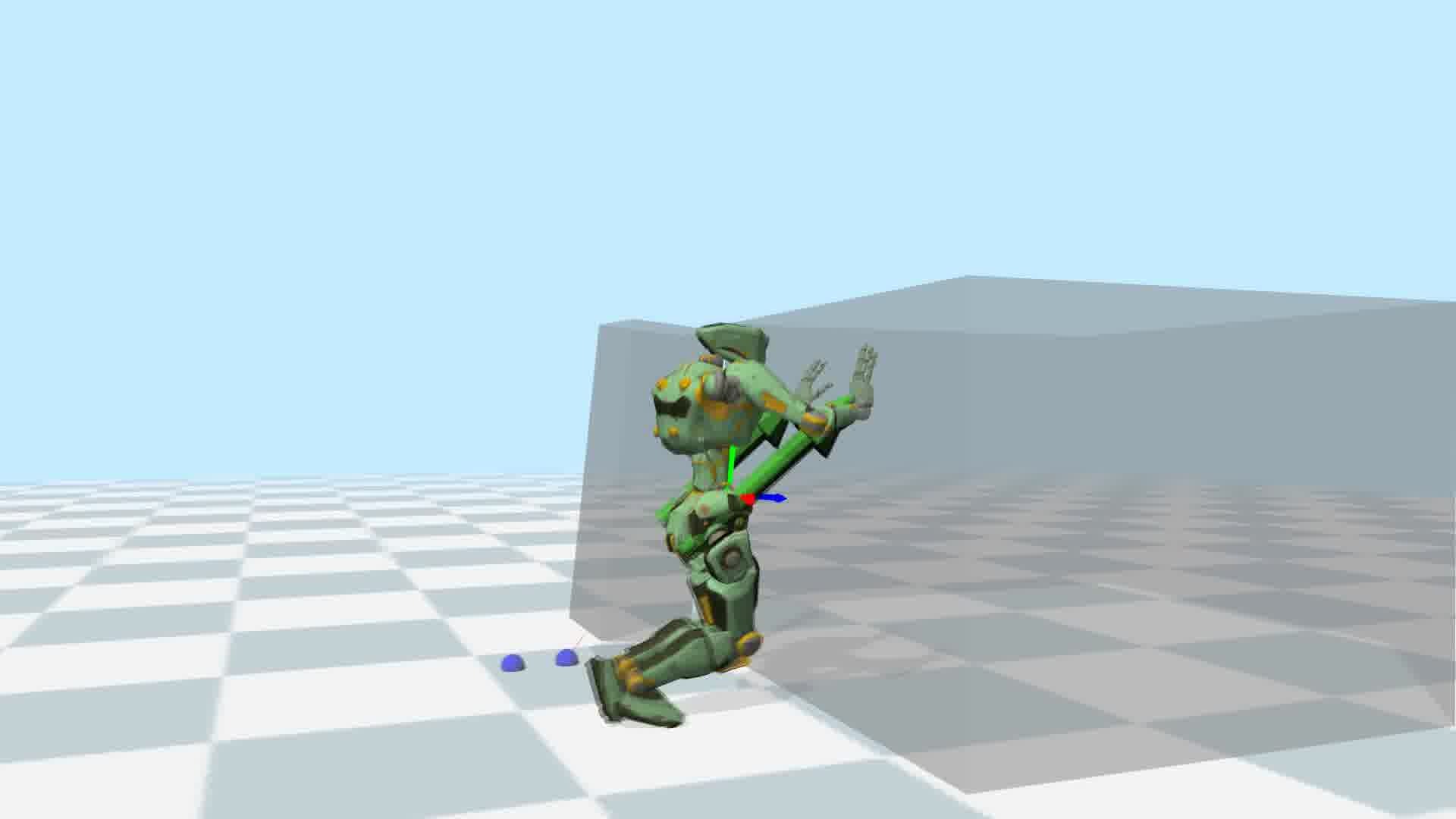}
	  \includegraphics[trim=550 50 450 250, clip, width=0.25\columnwidth]{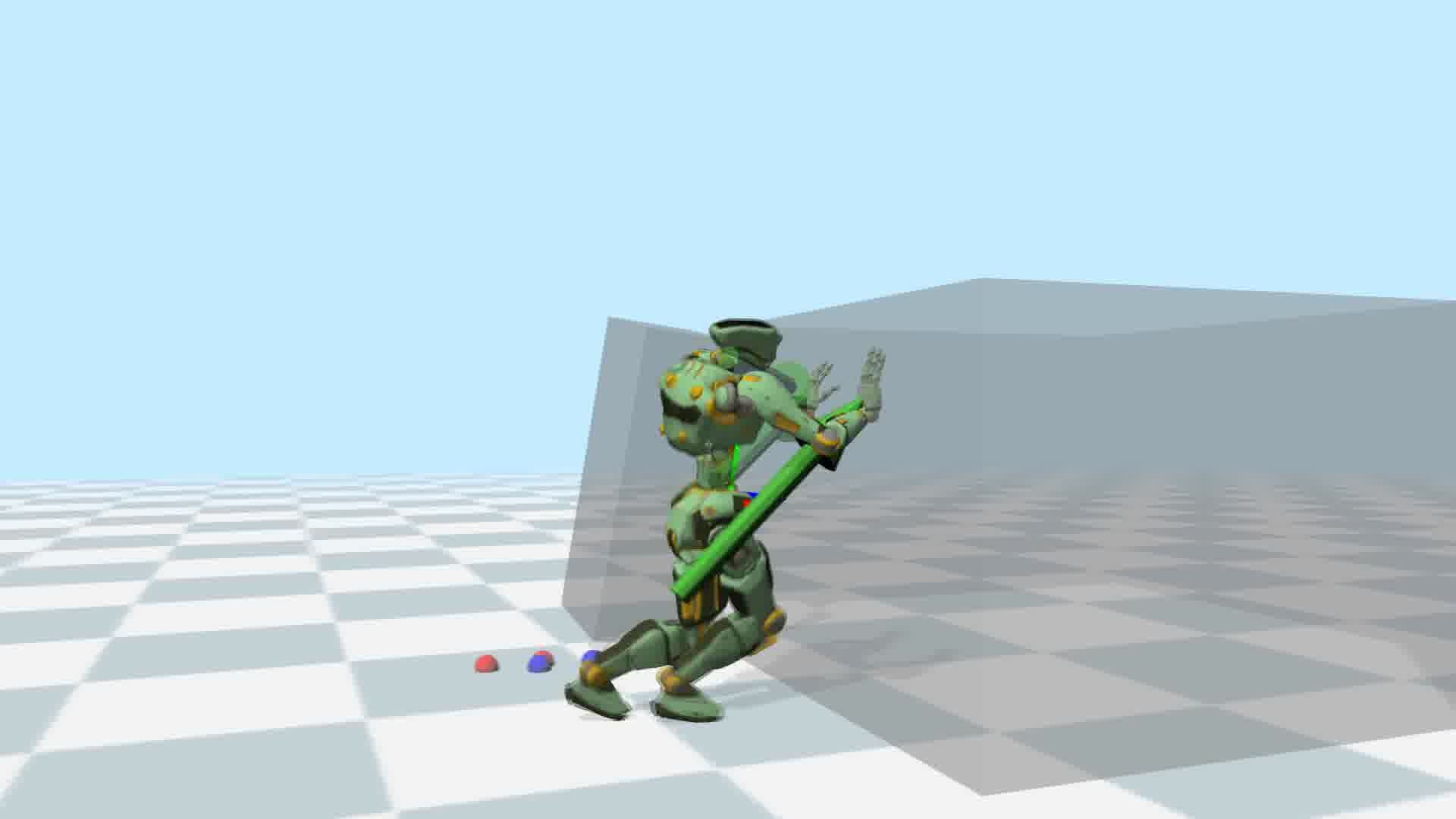}
	  \includegraphics[trim=550 50 450 250, clip, width=0.25\columnwidth]{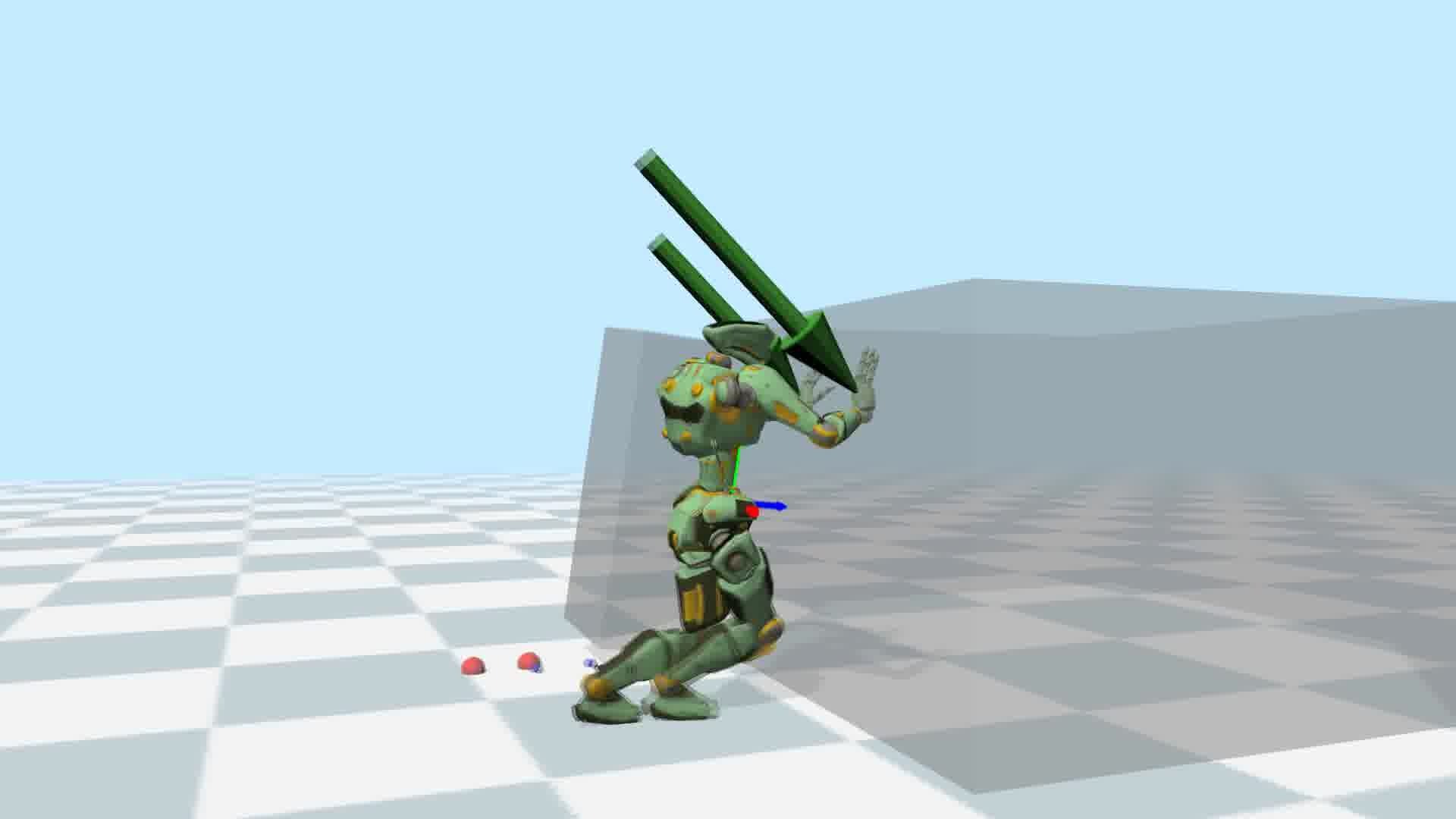}
        \label{fig:push_box_b}
				}
    \subfigure[\SI{10}{kg} - DeepMimic]{
	  \includegraphics[trim=550 50 450 250, clip, width=0.25\columnwidth]{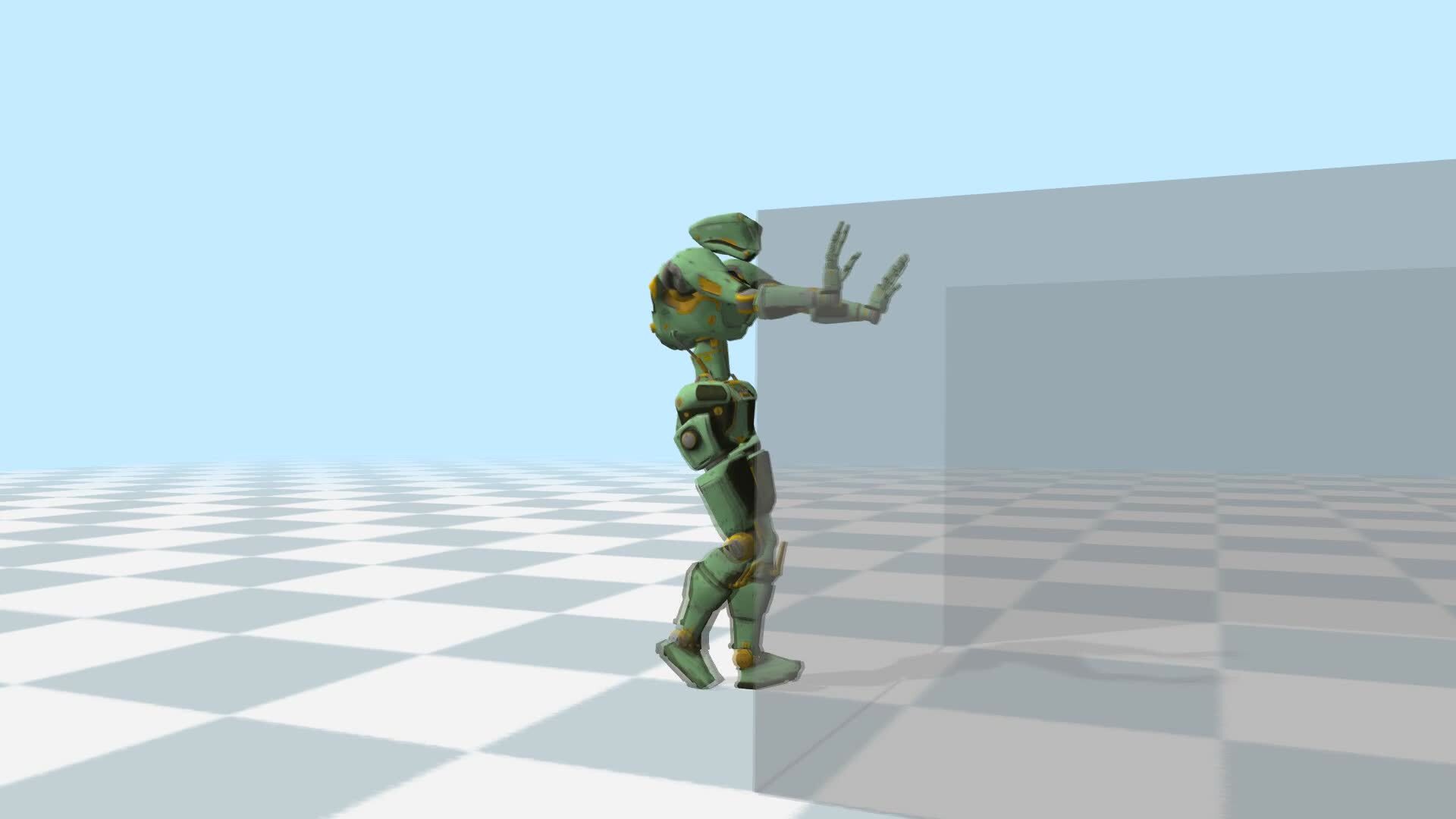}
	  \includegraphics[trim=550 50 450 250, clip, width=0.25\columnwidth]{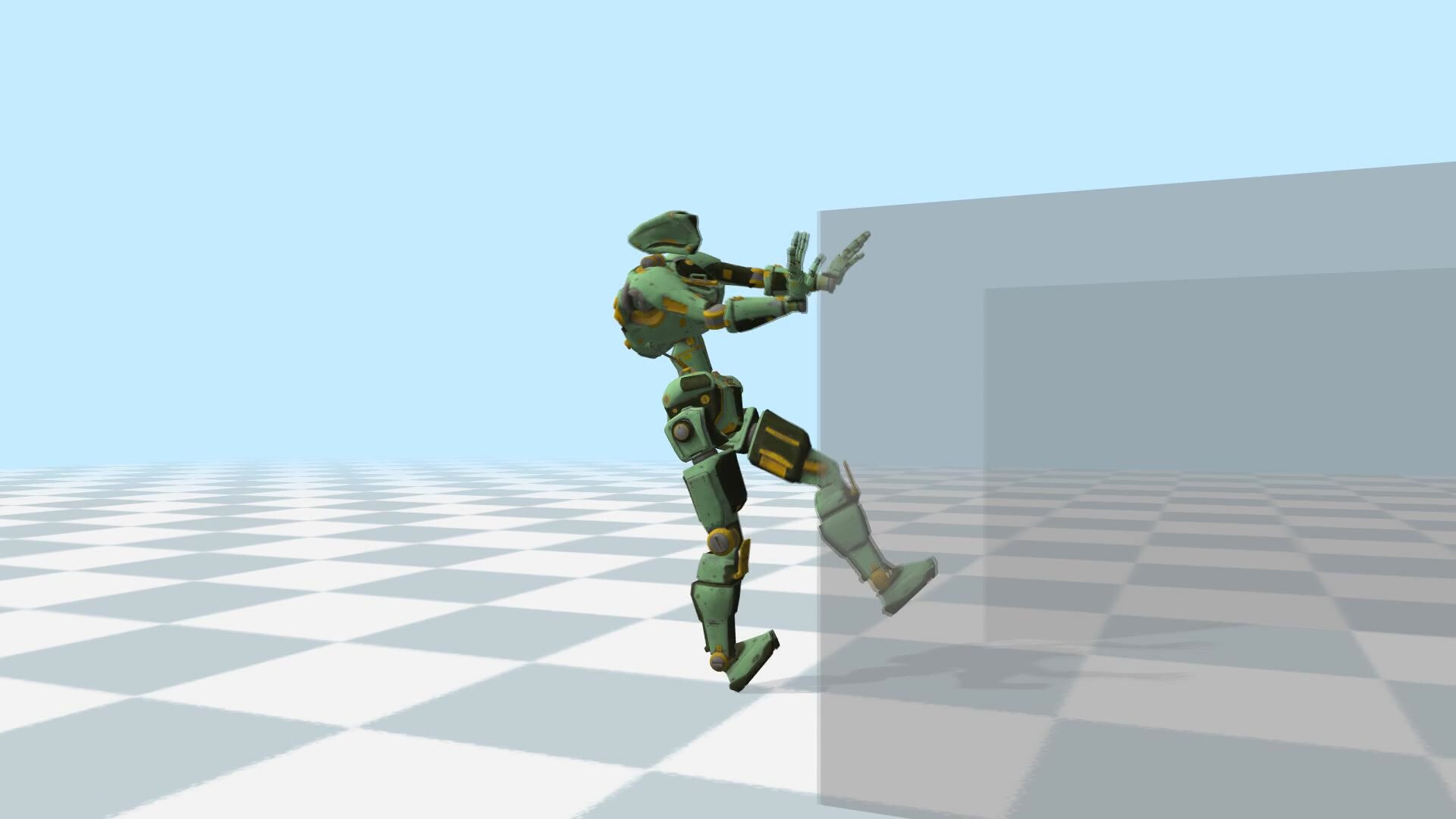}
	  \includegraphics[trim=550 50 450 250, clip, width=0.25\columnwidth]{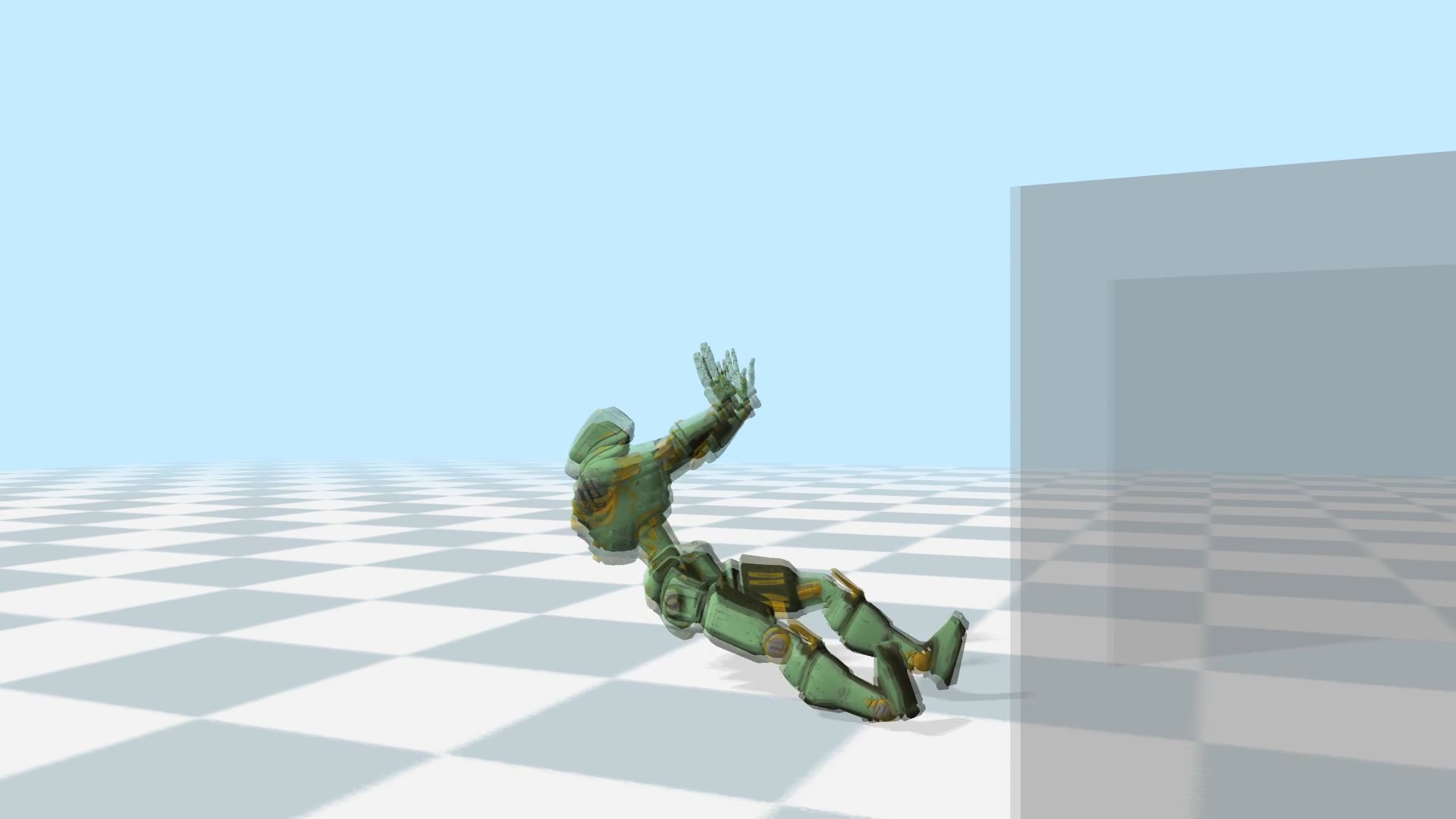}
	  \includegraphics[trim=550 50 450 250, clip, width=0.25\columnwidth]{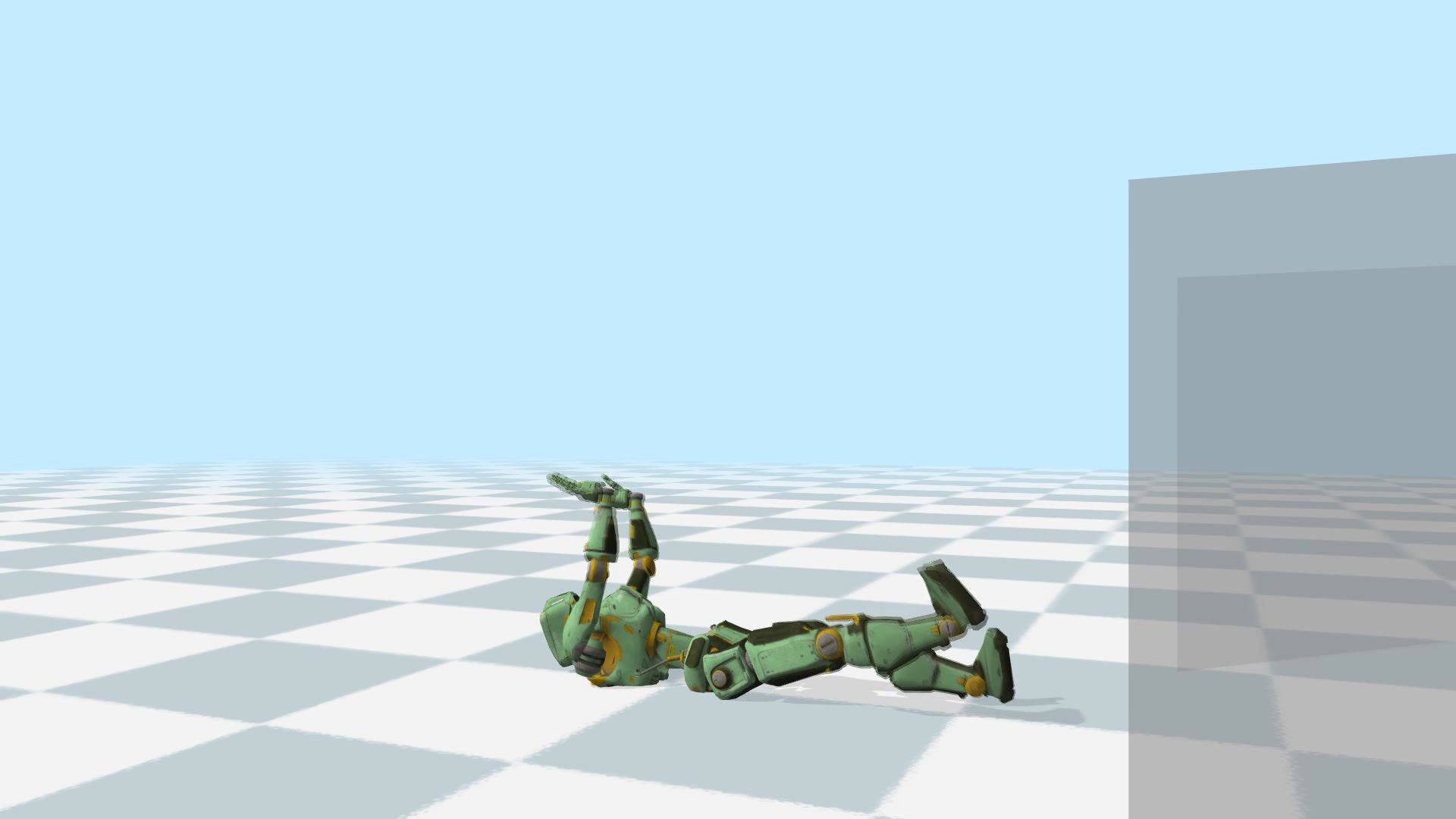}
        \label{fig:push_box_c}
				}
  \caption{\label{fig:push_box}
    Pushing a box of varying weights.
		   }
\end{figure}

\textbf{Push boxes.}
The policy was trained to track an edited version of the \textit{Walk} reference motion, where both arms were raised forward.
The learning environment did not include any boxes, and therefore, the policy did not receive any box information.
\textcolor{rev}{The penalty-based contact force exerted by the box is provided to the QP solver as external force information. }
As shown in Figure~\ref{fig:push_box}, the character was able to push boxes of different weights using different motions, by leaning forward to exert more force when the box is heavier.
Adjusting the foot landing position slightly behind the desired landing position in the action output of the policy resulted in a stronger pushing behavior when dealing with heavy boxes.
In contrast, a DeepMimic controller trained to track the same reference motion was unable to push the box.
Our SRB-based model achieved comparable results to a previous study \cite{Lee:2021:Parameterized} without the need for a policy parameterized by box weights.

\subsection{Comparision for external pushes}

\begin{figure}[htb]
  \centering
	\subfigure[Push from side]{
  \includegraphics[width=1\linewidth]{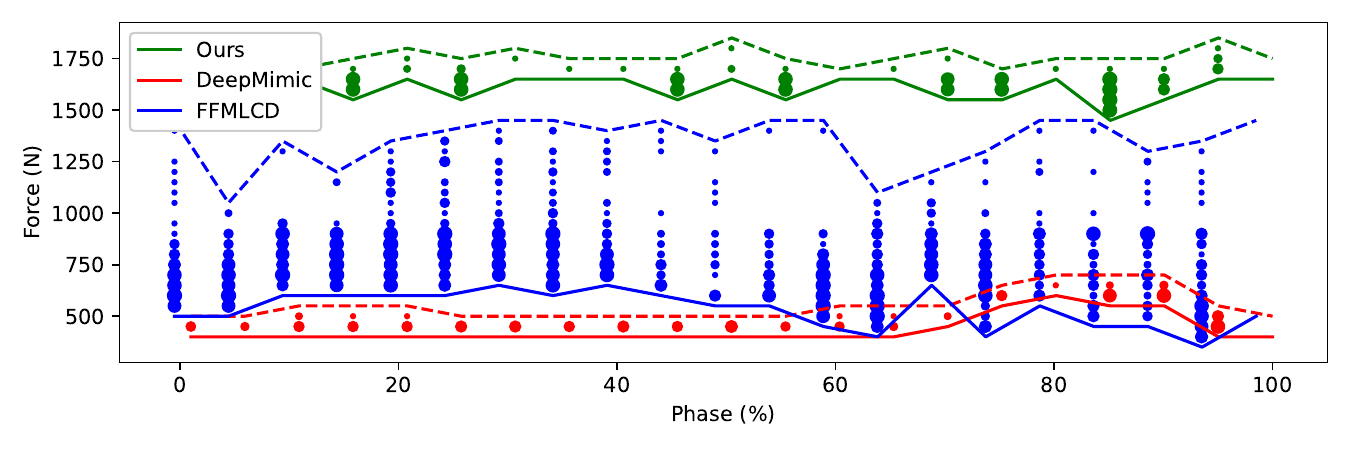}
  }
	\subfigure[Push from back]{
  \includegraphics[width=1\linewidth]{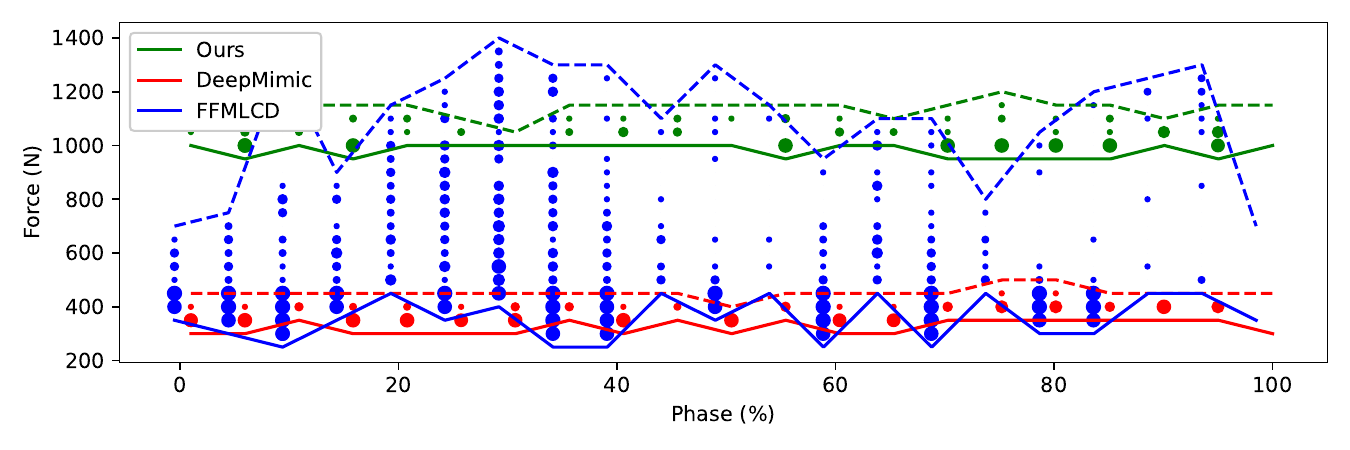}
  }
  \caption{\label{fig:external_forces}
        Plots for external pushes.
Dotted line: the smallest external force at which the character always loses balance.
Solid line: the largest external force at which the character maintains balance at each phase without falling.
Circles: 
    the ratio of successful balance maintenance (no falling within 20 seconds after the push) out of 10 trials at points where both balance maintenance and loss occur, with its size representing the proportion of successful balance maintenance.
           }
\end{figure}

We compared the balance maintenance level of characters controlled by our method, FFMLCD~\cite{kwon2020fast}, and DeepMimic~\cite{peng2018deepmimic}.
As reported in a previous study~\cite{lee_push-recovery_2015}, the response to external force can vary depending on the push timing, so we performed push experiments throughout the entire motion phase rather than at a specific point.
The experiments used \textit{Sprint} controllers and were performed at 20 evenly spaced phase points.
External force was applied to the character's root from the side or behind for 0.2 seconds, ranging from \SI{100}{N} to \SI{2000}{N} in \SI{50}{N} increments.
Each force level was applied 10 times at each point.

As depicted in Figure~\ref{fig:external_forces}, our method shows a higher minimum force at which the character always loses balance (dotted lines) in side pushes, while it is comparable to FFMLCD in back pushes.
However, from a practical standpoint, the important factor is the force magnitude at which the character maintains balance without falling (solid lines).
In this aspect, our method exhibits a force magnitude approximately three times higher than FFMLCD in both push cases.
This is because our method, utilizing RL, generates desired footstep positions at each frame to maintain balance, while FFMLCD relies on non-linear optimization-based planning that may occasionally struggle to find optimal solutions in challenging situations.
Moreover, FFMLCD performs this planning at each half cycle instead of each frame, which further limits its responsiveness.
Due to the limited adaptability caused by full-body tracking, DeepMimic easily fell even with weaker forces.
All controllers were more stable when pushed from the side compared to the back, because pushing from the back would cause the body to lean forward and the swing foot to suddenly hit the ground, resulting in a fall or excessive contact force.

\subsection{Interactive control}
\label{sec:results_interactive_control}

We also conducted experiments to enable more useful applications by allowing users to interactively control the facing direction of the character. For this purpose, we added a scalar $\delta_y$, representing the difference between the desired facing direction and the current SRB facing direction, to the inputs of the policy network. The policy is then trained using the following reward $r^c_t$:

\begin{equation}
\label{eq:reward_interactive}
	r^c_t = \mathrm{exp}(-r_t) \cdot \mathrm{exp}(-w^c \cdot (\mathrm{cos}(\delta_y)-1)) ,
\end{equation}
where $r_t$ represents the reward defined in Equation~\ref{eq:reward}, and $w^c$ denotes the weight.

\section{Discussion}

In this paper, we have presented a framework that can learn an adaptive tracking policy in a sample-efficient manner using a SRB character simplifying a full-body character. 
The QP-based SRB simulation allows efficient learning of a policy,  
and the simulated SRB motion %
can be converted back into a full-body motion using precomputed delta and a momentum-mapped inverse kinematics solver.

Thanks to the less detailed SRB model-based learning, our method enables fast policy learning with fewer samples compared to traditional full-body methods.
Moreover, the learned policy exhibits adaptability in various environments.
However, it is worth noting that our method has limitations in accurately representing contact-rich scenarios, %
unlike full-body physics simulations.
For example, it does not explicitly represent collisions between body parts and the ground when the character falls.
In our method, "falling" refers to when the center of mass is too close to the ground, making the solution of MMIK meaningless.
In practice, we can partially address this limitation by switching to a full-body simulation using PD-servo to generate a more natural fall when the character begins to lose balance.
\textcolor{rev}{While our assumed constant inertia for the SRB character does not precisely mimic the time-varying inertia of a full-body, it successfully generated plausible locomotion and jumping movements, as demonstrated.
As discussed in \cite{kwon2020fast}, modeling time-varying inertia may improve the accuracy of motions that depend on rotational speed.}

In our method, each controller is trained based on a single short reference motion clip.
Instead of relying on a large number of reference motions, our approach prioritizes maximizing the number of situations that the same controller can handle without additional learning.
Once such an adaptability is achieved, we believe that it can be straightforwardly extended in the direction of enabling a wider range of motions by using more complex network structures supporting unorganized motion capture datasets, such as VAE or RNN~\cite{Park2019,controlvae22}.  %
The proposed framework allows only indirect adjustment of the contact timing based on the adjustment of the speed of phase progression.
This can negatively affect robustness because a limited range of contact timing adjustment interferes with more flexible adaptation of the character to external force. 
However, unlike DeepMimic-style controllers which use fixed contact timing, 
the contact timing is easily adjustable in our framework that employs virtual legs, and thus it would help to improve the robustness of the controllers by adjusting the contact timing in a similar way as tested in \cite{Brachiation2022}.  %
One of the computational bottlenecks in our learning process is solving the QP.    %
As future work, it would be beneficial to explore the use of a GPU-based QP solver, such as the one presented in \cite{xie2022glide}, to further accelerate learning.

\begin{acks}
    This work was supported by the National Research Foundation of Korea (NRF) grant (NRF-2019R1C1C1006778, RS-2023-00222776) and Institute of Information \& communications Technology Planning \& Evaluation (IITP) grant (No. 2021-0-00320), funded by the Korea Government (MSIT).
\end{acks}

\bibliographystyle{ACM-Reference-Format} 
\bibliography{egbibsample2}


\begin{thebibliography}{50}


\ifx \showCODEN    \undefined \def \showCODEN     #1{\unskip}     \fi
\ifx \showDOI      \undefined \def \showDOI       #1{#1}\fi
\ifx \showISBNx    \undefined \def \showISBNx     #1{\unskip}     \fi
\ifx \showISBNxiii \undefined \def \showISBNxiii  #1{\unskip}     \fi
\ifx \showISSN     \undefined \def \showISSN      #1{\unskip}     \fi
\ifx \showLCCN     \undefined \def \showLCCN      #1{\unskip}     \fi
\ifx \shownote     \undefined \def \shownote      #1{#1}          \fi
\ifx \showarticletitle \undefined \def \showarticletitle #1{#1}   \fi
\ifx \showURL      \undefined \def \showURL       {\relax}        \fi
\providecommand\bibfield[2]{#2}
\providecommand\bibinfo[2]{#2}
\providecommand\natexlab[1]{#1}
\providecommand\showeprint[2][]{arXiv:#2}

\bibitem[Abe et~al\mbox{.}(2007)]%
        {abe2007multiobjective}
\bibfield{author}{\bibinfo{person}{Yeuhi Abe}, \bibinfo{person}{Marco
  Da~Silva}, {and} \bibinfo{person}{Jovan Popovi{\'c}}.}
  \bibinfo{year}{2007}\natexlab{}.
\newblock \showarticletitle{Multiobjective control with frictional contacts}.
  In \bibinfo{booktitle}{\emph{Proceedings of the 2007 ACM
  SIGGRAPH/Eurographics symposium on Computer animation}}.
  \bibinfo{pages}{249--258}.
\newblock


\bibitem[Agrawal et~al\mbox{.}(2013)]%
        {agrawal2013diverse}
\bibfield{author}{\bibinfo{person}{Shailen Agrawal}, \bibinfo{person}{Shuo
  Shen}, {and} \bibinfo{person}{Michiel van~de Panne}.}
  \bibinfo{year}{2013}\natexlab{}.
\newblock \showarticletitle{Diverse motion variations for physics-based
  character animation}. In \bibinfo{booktitle}{\emph{Proceedings of the 12th
  ACM SIGGRAPH/Eurographics Symposium on Computer Animation}}.
  \bibinfo{pages}{37--44}.
\newblock


\bibitem[Bergamin et~al\mbox{.}(2019)]%
        {bergamin_drecon_2019}
\bibfield{author}{\bibinfo{person}{Kevin Bergamin}, \bibinfo{person}{Simon
  Clavet}, \bibinfo{person}{Daniel Holden}, {and}
  \bibinfo{person}{James~Richard Forbes}.} \bibinfo{year}{2019}\natexlab{}.
\newblock \showarticletitle{{DReCon}: data-driven responsive control of
  physics-based characters}.
\newblock \bibinfo{journal}{\emph{ACM Transactions on Graphics (TOG)}}
  \bibinfo{volume}{38}, \bibinfo{number}{6} (\bibinfo{date}{Nov.}
  \bibinfo{year}{2019}), \bibinfo{pages}{206:1--206:11}.
\newblock
\showISSN{0730-0301}


\bibitem[Brockman et~al\mbox{.}(2016)]%
        {brockman2016openai}
\bibfield{author}{\bibinfo{person}{Greg Brockman}, \bibinfo{person}{Vicki
  Cheung}, \bibinfo{person}{Ludwig Pettersson}, \bibinfo{person}{Jonas
  Schneider}, \bibinfo{person}{John Schulman}, \bibinfo{person}{Jie Tang},
  {and} \bibinfo{person}{Wojciech Zaremba}.} \bibinfo{year}{2016}\natexlab{}.
\newblock \showarticletitle{Openai gym}.
\newblock \bibinfo{journal}{\emph{arXiv preprint arXiv:1606.01540}}
  (\bibinfo{year}{2016}).
\newblock


\bibitem[Chentanez et~al\mbox{.}(2018)]%
        {Chentanez2018}
\bibfield{author}{\bibinfo{person}{Nuttapong Chentanez},
  \bibinfo{person}{Matthias M\"{u}ller}, \bibinfo{person}{Miles Macklin},
  \bibinfo{person}{Viktor Makoviychuk}, {and} \bibinfo{person}{Stefan
  Jeschke}.} \bibinfo{year}{2018}\natexlab{}.
\newblock \showarticletitle{Physics-Based Motion Capture Imitation with Deep
  Reinforcement Learning}. In \bibinfo{booktitle}{\emph{Proceedings of the 11th
  ACM SIGGRAPH Conference on Motion, Interaction and Games}}
  \emph{(\bibinfo{series}{MIG '18})}. Article \bibinfo{articleno}{1},
  \bibinfo{numpages}{10}~pages.
\newblock
\showISBNx{9781450360159}


\bibitem[Cho et~al\mbox{.}(2021)]%
        {cho_motion_2021}
\bibfield{author}{\bibinfo{person}{Kyungmin Cho}, \bibinfo{person}{Chaelin
  Kim}, \bibinfo{person}{Jungjin Park}, \bibinfo{person}{Joonkyu Park}, {and}
  \bibinfo{person}{Junyong Noh}.} \bibinfo{year}{2021}\natexlab{}.
\newblock \showarticletitle{Motion recommendation for online character
  control}.
\newblock \bibinfo{journal}{\emph{ACM Transactions on Graphics}}
  \bibinfo{volume}{40}, \bibinfo{number}{6} (\bibinfo{year}{2021}).
\newblock
\showISSN{0730-0301}


\bibitem[Coros et~al\mbox{.}(2009)]%
        {coros2009robust}
\bibfield{author}{\bibinfo{person}{Stelian Coros}, \bibinfo{person}{Philippe
  Beaudoin}, {and} \bibinfo{person}{Michiel van~de Panne}.}
  \bibinfo{year}{2009}\natexlab{}.
\newblock \showarticletitle{Robust task-based control policies for
  physics-based characters}.
\newblock \bibinfo{journal}{\emph{ACM Trans. Graph. (Proc. SIGGRAPH Asia)}}
  \bibinfo{volume}{28}, \bibinfo{number}{5} (\bibinfo{year}{2009}),
  \bibinfo{pages}{1--9}.
\newblock


\bibitem[Coros et~al\mbox{.}(2010)]%
        {coros2010generalized}
\bibfield{author}{\bibinfo{person}{Stelian Coros}, \bibinfo{person}{Philippe
  Beaudoin}, {and} \bibinfo{person}{Michiel Van~de Panne}.}
  \bibinfo{year}{2010}\natexlab{}.
\newblock \showarticletitle{Generalized biped walking control}.
\newblock \bibinfo{journal}{\emph{ACM Transactions On Graphics (TOG)}}
  \bibinfo{volume}{29}, \bibinfo{number}{4} (\bibinfo{year}{2010}),
  \bibinfo{pages}{1--9}.
\newblock


\bibitem[Da~Silva et~al\mbox{.}(2008)]%
        {da2008simulation}
\bibfield{author}{\bibinfo{person}{Marco Da~Silva}, \bibinfo{person}{Yeuhi
  Abe}, {and} \bibinfo{person}{Jovan Popovi{\'c}}.}
  \bibinfo{year}{2008}\natexlab{}.
\newblock \showarticletitle{Simulation of human motion data using short-horizon
  model-predictive control}. In \bibinfo{booktitle}{\emph{Computer Graphics
  Forum}}, Vol.~\bibinfo{volume}{27}. \bibinfo{pages}{371--380}.
\newblock


\bibitem[Duan et~al\mbox{.}(2016)]%
        {duan2016benchmarking}
\bibfield{author}{\bibinfo{person}{Yan Duan}, \bibinfo{person}{Xi Chen},
  \bibinfo{person}{Rein Houthooft}, \bibinfo{person}{John Schulman}, {and}
  \bibinfo{person}{Pieter Abbeel}.} \bibinfo{year}{2016}\natexlab{}.
\newblock \showarticletitle{Benchmarking deep reinforcement learning for
  continuous control}. In \bibinfo{booktitle}{\emph{Proceedings of the 33rd
  {International} {Conference} on {International} {Conference} on {Machine}
  {Learning} - {Volume} 48}} \emph{(\bibinfo{series}{{ICML}'16})}.
  \bibinfo{pages}{1329--1338}.
\newblock


\bibitem[Ellis et~al\mbox{.}(2007)]%
        {ellis2007cdm}
\bibfield{author}{\bibinfo{person}{Jane Ellis}, \bibinfo{person}{Harald
  Winkler}, \bibinfo{person}{Jan Corfee-Morlot}, {and}
  \bibinfo{person}{Fr{\'e}d{\'e}ric Gagnon-Lebrun}.}
  \bibinfo{year}{2007}\natexlab{}.
\newblock \showarticletitle{CDM: Taking stock and looking forward}.
\newblock \bibinfo{journal}{\emph{Energy policy}} \bibinfo{volume}{35},
  \bibinfo{number}{1} (\bibinfo{year}{2007}), \bibinfo{pages}{15--28}.
\newblock


\bibitem[Ha and Liu(2014)]%
        {ha2014iterative}
\bibfield{author}{\bibinfo{person}{Sehoon Ha} {and} \bibinfo{person}{C~Karen
  Liu}.} \bibinfo{year}{2014}\natexlab{}.
\newblock \showarticletitle{Iterative training of dynamic skills inspired by
  human coaching techniques}.
\newblock \bibinfo{journal}{\emph{ACM Transactions on Graphics (TOG)}}
  \bibinfo{volume}{34}, \bibinfo{number}{1} (\bibinfo{year}{2014}),
  \bibinfo{pages}{1--11}.
\newblock


\bibitem[H{\"a}m{\"a}l{\"a}inen et~al\mbox{.}(2015)]%
        {hamalainen2015online}
\bibfield{author}{\bibinfo{person}{Perttu H{\"a}m{\"a}l{\"a}inen},
  \bibinfo{person}{Joose Rajam{\"a}ki}, {and} \bibinfo{person}{C~Karen Liu}.}
  \bibinfo{year}{2015}\natexlab{}.
\newblock \showarticletitle{Online control of simulated humanoids using
  particle belief propagation}.
\newblock \bibinfo{journal}{\emph{ACM Transactions on Graphics (TOG)}}
  \bibinfo{volume}{34}, \bibinfo{number}{4} (\bibinfo{year}{2015}),
  \bibinfo{pages}{1--13}.
\newblock


\bibitem[Holden et~al\mbox{.}(2017)]%
        {holden_phase_2017}
\bibfield{author}{\bibinfo{person}{Daniel Holden}, \bibinfo{person}{Taku
  Komura}, {and} \bibinfo{person}{Jun Saito}.} \bibinfo{year}{2017}\natexlab{}.
\newblock \showarticletitle{Phase-Functioned Neural Networks for Character
  Control}.
\newblock \bibinfo{journal}{\emph{ACM Transactions on Graphics}}
  \bibinfo{volume}{36}, \bibinfo{number}{4}, Article \bibinfo{articleno}{42}
  (\bibinfo{year}{2017}), \bibinfo{numpages}{13}~pages.
\newblock
\showISSN{0730-0301}


\bibitem[Hwang et~al\mbox{.}(2017)]%
        {hwang2017performance}
\bibfield{author}{\bibinfo{person}{Jaepyung Hwang}, \bibinfo{person}{Kwanguk
  Kim}, \bibinfo{person}{Il~Hong Suh}, {and} \bibinfo{person}{Taesoo Kwon}.}
  \bibinfo{year}{2017}\natexlab{}.
\newblock \showarticletitle{Performance-based animation using constraints for
  virtual object manipulation}.
\newblock \bibinfo{journal}{\emph{IEEE computer graphics and applications}}
  \bibinfo{volume}{37}, \bibinfo{number}{4} (\bibinfo{year}{2017}),
  \bibinfo{pages}{95--102}.
\newblock


\bibitem[Kwon and Hodgins(2017)]%
        {kwon2017momentum}
\bibfield{author}{\bibinfo{person}{Taesoo Kwon} {and}
  \bibinfo{person}{Jessica~K Hodgins}.} \bibinfo{year}{2017}\natexlab{}.
\newblock \showarticletitle{Momentum-mapped inverted pendulum models for
  controlling dynamic human motions}.
\newblock \bibinfo{journal}{\emph{ACM Transactions on Graphics (TOG)}}
  \bibinfo{volume}{36}, \bibinfo{number}{1} (\bibinfo{year}{2017}),
  \bibinfo{pages}{1--14}.
\newblock


\bibitem[Kwon et~al\mbox{.}(2020)]%
        {kwon2020fast}
\bibfield{author}{\bibinfo{person}{Taesoo Kwon}, \bibinfo{person}{Yoonsang
  Lee}, {and} \bibinfo{person}{Michiel Van De~Panne}.}
  \bibinfo{year}{2020}\natexlab{}.
\newblock \showarticletitle{Fast and flexible multilegged locomotion using
  learned centroidal dynamics}.
\newblock \bibinfo{journal}{\emph{ACM Transactions on Graphics (TOG)}}
  \bibinfo{volume}{39}, \bibinfo{number}{4} (\bibinfo{year}{2020}),
  \bibinfo{pages}{46--1}.
\newblock


\bibitem[Lee et~al\mbox{.}(2022)]%
        {chimeras22}
\bibfield{author}{\bibinfo{person}{Seyoung Lee}, \bibinfo{person}{Jiye Lee},
  {and} \bibinfo{person}{Jehee Lee}.} \bibinfo{year}{2022}\natexlab{}.
\newblock \showarticletitle{Learning {Virtual} {Chimeras} by {Dynamic} {Motion}
  {Reassembly}}.
\newblock \bibinfo{journal}{\emph{ACM Transactions on Graphics}}
  \bibinfo{volume}{41}, \bibinfo{number}{6} (\bibinfo{year}{2022}),
  \bibinfo{pages}{182:1--182:13}.
\newblock
\showISSN{0730-0301}


\bibitem[Lee et~al\mbox{.}(2021)]%
        {Lee:2021:Parameterized}
\bibfield{author}{\bibinfo{person}{Seyoung Lee}, \bibinfo{person}{Sunmin Lee},
  \bibinfo{person}{Yongwoo Lee}, {and} \bibinfo{person}{Jehee Lee}.}
  \bibinfo{year}{2021}\natexlab{}.
\newblock \showarticletitle{Learning a family of motor skills from a single
  motion clip}.
\newblock \bibinfo{journal}{\emph{ACM Transactions on Graphics}}
  \bibinfo{volume}{40}, \bibinfo{number}{4} (\bibinfo{date}{July}
  \bibinfo{year}{2021}), \bibinfo{pages}{93:1--93:13}.
\newblock
\showISSN{0730-0301}


\bibitem[Lee et~al\mbox{.}(2010)]%
        {lee2010data}
\bibfield{author}{\bibinfo{person}{Yoonsang Lee}, \bibinfo{person}{Sungeun
  Kim}, {and} \bibinfo{person}{Jehee Lee}.} \bibinfo{year}{2010}\natexlab{}.
\newblock \showarticletitle{Data-driven biped control}.
\newblock \bibinfo{journal}{\emph{ACM Trans. Graph.}} \bibinfo{volume}{29},
  \bibinfo{number}{4} (\bibinfo{year}{2010}), \bibinfo{pages}{1--8}.
\newblock


\bibitem[Lee et~al\mbox{.}(2015)]%
        {lee_push-recovery_2015}
\bibfield{author}{\bibinfo{person}{Yoonsang Lee}, \bibinfo{person}{Kyungho
  Lee}, \bibinfo{person}{Soon-Sun Kwon}, \bibinfo{person}{Jiwon Jeong},
  \bibinfo{person}{Carol O'Sullivan}, \bibinfo{person}{Moon~Seok Park}, {and}
  \bibinfo{person}{Jehee Lee}.} \bibinfo{year}{2015}\natexlab{}.
\newblock \showarticletitle{Push-recovery {Stability} of {Biped} {Locomotion}}.
\newblock \bibinfo{journal}{\emph{ACM Transactions on Graphics (TOG)}}
  \bibinfo{volume}{34}, \bibinfo{number}{6} (\bibinfo{year}{2015}),
  \bibinfo{pages}{180:1--180:9}.
\newblock


\bibitem[Ling et~al\mbox{.}(2020)]%
        {Ling_2020}
\bibfield{author}{\bibinfo{person}{Hung~Yu Ling}, \bibinfo{person}{Fabio
  Zinno}, \bibinfo{person}{George Cheng}, {and} \bibinfo{person}{Michiel Van
  De~Panne}.} \bibinfo{year}{2020}\natexlab{}.
\newblock \showarticletitle{Character controllers using motion VAEs}.
\newblock \bibinfo{journal}{\emph{ACM Transactions on Graphics}}
  \bibinfo{volume}{39}, \bibinfo{number}{4} (\bibinfo{year}{2020}).
\newblock
\showISSN{1557-7368}


\bibitem[Liu and Hodgins(2017)]%
        {liu2017learning}
\bibfield{author}{\bibinfo{person}{Libin Liu} {and} \bibinfo{person}{Jessica
  Hodgins}.} \bibinfo{year}{2017}\natexlab{}.
\newblock \showarticletitle{Learning to schedule control fragments for
  physics-based characters using deep q-learning}.
\newblock \bibinfo{journal}{\emph{ACM Transactions on Graphics (TOG)}}
  \bibinfo{volume}{36}, \bibinfo{number}{3} (\bibinfo{year}{2017}),
  \bibinfo{pages}{1--14}.
\newblock


\bibitem[Macchietto et~al\mbox{.}(2009)]%
        {Macchietto09}
\bibfield{author}{\bibinfo{person}{Adriano Macchietto}, \bibinfo{person}{Victor
  Zordan}, {and} \bibinfo{person}{Christian~R. Shelton}.}
  \bibinfo{year}{2009}\natexlab{}.
\newblock \showarticletitle{Momentum control for balance}.
\newblock \bibinfo{journal}{\emph{ACM Transactions on Graphics}}
  \bibinfo{volume}{28}, \bibinfo{number}{3} (\bibinfo{date}{July}
  \bibinfo{year}{2009}), \bibinfo{pages}{80:1--80:8}.
\newblock
\showISSN{0730-0301}


\bibitem[Mordatch et~al\mbox{.}(2012)]%
        {mordatch2012discovery}
\bibfield{author}{\bibinfo{person}{Igor Mordatch}, \bibinfo{person}{Emanuel
  Todorov}, {and} \bibinfo{person}{Zoran Popovi{\'c}}.}
  \bibinfo{year}{2012}\natexlab{}.
\newblock \showarticletitle{Discovery of complex behaviors through
  contact-invariant optimization}.
\newblock \bibinfo{journal}{\emph{ACM Transactions on Graphics (TOG)}}
  \bibinfo{volume}{31}, \bibinfo{number}{4} (\bibinfo{year}{2012}),
  \bibinfo{pages}{1--8}.
\newblock


\bibitem[Park et~al\mbox{.}(2019)]%
        {Park2019}
\bibfield{author}{\bibinfo{person}{Soohwan Park}, \bibinfo{person}{Hoseok Ryu},
  \bibinfo{person}{Seyoung Lee}, \bibinfo{person}{Sunmin Lee}, {and}
  \bibinfo{person}{Jehee Lee}.} \bibinfo{year}{2019}\natexlab{}.
\newblock \showarticletitle{Learning predict-and-simulate policies from
  unorganized human motion data}.
\newblock \bibinfo{journal}{\emph{ACM Transactions on Graphics}}
  \bibinfo{volume}{38}, \bibinfo{number}{6} (\bibinfo{year}{2019}),
  \bibinfo{pages}{205:1--205:11}.
\newblock
\showISSN{0730-0301}


\bibitem[Peng et~al\mbox{.}(2018a)]%
        {peng2018deepmimic}
\bibfield{author}{\bibinfo{person}{Xue~Bin Peng}, \bibinfo{person}{Pieter
  Abbeel}, \bibinfo{person}{Sergey Levine}, {and} \bibinfo{person}{Michiel
  van~de Panne}.} \bibinfo{year}{2018}\natexlab{a}.
\newblock \showarticletitle{Deepmimic: Example-guided deep reinforcement
  learning of physics-based character skills}.
\newblock \bibinfo{journal}{\emph{ACM Transactions on Graphics (TOG)}}
  \bibinfo{volume}{37}, \bibinfo{number}{4} (\bibinfo{year}{2018}),
  \bibinfo{pages}{1--14}.
\newblock


\bibitem[Peng et~al\mbox{.}(2015)]%
        {peng2015dynamic}
\bibfield{author}{\bibinfo{person}{Xue~Bin Peng}, \bibinfo{person}{Glen
  Berseth}, {and} \bibinfo{person}{Michiel Van~de Panne}.}
  \bibinfo{year}{2015}\natexlab{}.
\newblock \showarticletitle{Dynamic terrain traversal skills using
  reinforcement learning}.
\newblock \bibinfo{journal}{\emph{ACM Transactions on Graphics (TOG)}}
  \bibinfo{volume}{34}, \bibinfo{number}{4} (\bibinfo{year}{2015}),
  \bibinfo{pages}{1--11}.
\newblock


\bibitem[Peng et~al\mbox{.}(2016)]%
        {peng2016terrain}
\bibfield{author}{\bibinfo{person}{Xue~Bin Peng}, \bibinfo{person}{Glen
  Berseth}, {and} \bibinfo{person}{Michiel Van~de Panne}.}
  \bibinfo{year}{2016}\natexlab{}.
\newblock \showarticletitle{Terrain-adaptive locomotion skills using deep
  reinforcement learning}.
\newblock \bibinfo{journal}{\emph{ACM Transactions on Graphics (TOG)}}
  \bibinfo{volume}{35}, \bibinfo{number}{4} (\bibinfo{year}{2016}),
  \bibinfo{pages}{1--12}.
\newblock


\bibitem[Peng et~al\mbox{.}(2017)]%
        {2017-TOG-deepLoco}
\bibfield{author}{\bibinfo{person}{Xue~Bin Peng}, \bibinfo{person}{Glen
  Berseth}, \bibinfo{person}{KangKang Yin}, {and} \bibinfo{person}{Michiel
  van~de Panne}.} \bibinfo{year}{2017}\natexlab{}.
\newblock \showarticletitle{DeepLoco: Dynamic Locomotion Skills Using
  Hierarchical Deep Reinforcement Learning}.
\newblock \bibinfo{journal}{\emph{ACM Transactions on Graphics (Proc. SIGGRAPH
  2017)}} \bibinfo{volume}{36}, \bibinfo{number}{4} (\bibinfo{year}{2017}).
\newblock


\bibitem[Peng et~al\mbox{.}(2022)]%
        {peng_ase_2022}
\bibfield{author}{\bibinfo{person}{Xue~Bin Peng}, \bibinfo{person}{Yunrong
  Guo}, \bibinfo{person}{Lina Halper}, \bibinfo{person}{Sergey Levine}, {and}
  \bibinfo{person}{Sanja Fidler}.} \bibinfo{year}{2022}\natexlab{}.
\newblock \showarticletitle{{ASE}: large-scale reusable adversarial skill
  embeddings for physically simulated characters}.
\newblock \bibinfo{journal}{\emph{ACM Transactions on Graphics}}
  \bibinfo{volume}{41}, \bibinfo{number}{4} (\bibinfo{date}{July}
  \bibinfo{year}{2022}), \bibinfo{pages}{94:1--94:17}.
\newblock
\showISSN{0730-0301}


\bibitem[Peng et~al\mbox{.}(2018b)]%
        {peng18}
\bibfield{author}{\bibinfo{person}{Xue~Bin Peng}, \bibinfo{person}{Angjoo
  Kanazawa}, \bibinfo{person}{Jitendra Malik}, \bibinfo{person}{Pieter Abbeel},
  {and} \bibinfo{person}{Sergey Levine}.} \bibinfo{year}{2018}\natexlab{b}.
\newblock \showarticletitle{SFV: Reinforcement Learning of Physical Skills from
  Videos}.
\newblock \bibinfo{journal}{\emph{ACM Trans. Graph.}} \bibinfo{volume}{37},
  \bibinfo{number}{6}, Article \bibinfo{articleno}{178} (\bibinfo{date}{dec}
  \bibinfo{year}{2018}), \bibinfo{numpages}{14}~pages.
\newblock
\showISSN{0730-0301}
\urldef\tempurl%
\url{https://doi.org/10.1145/3272127.3275014}
\showDOI{\tempurl}


\bibitem[Peng et~al\mbox{.}(2021)]%
        {peng_amp_2021}
\bibfield{author}{\bibinfo{person}{Xue~Bin Peng}, \bibinfo{person}{Ze Ma},
  \bibinfo{person}{Pieter Abbeel}, \bibinfo{person}{Sergey Levine}, {and}
  \bibinfo{person}{Angjoo Kanazawa}.} \bibinfo{year}{2021}\natexlab{}.
\newblock \showarticletitle{{AMP}: adversarial motion priors for stylized
  physics-based character control}.
\newblock \bibinfo{journal}{\emph{ACM Transactions on Graphics}}
  \bibinfo{volume}{40}, \bibinfo{number}{4} (\bibinfo{year}{2021}),
  \bibinfo{pages}{144:1--144:20}.
\newblock
\showISSN{0730-0301}


\bibitem[Rajeswaran et~al\mbox{.}(2017)]%
        {rajeswaran2017learning}
\bibfield{author}{\bibinfo{person}{Aravind Rajeswaran}, \bibinfo{person}{Vikash
  Kumar}, \bibinfo{person}{Abhishek Gupta}, \bibinfo{person}{Giulia Vezzani},
  \bibinfo{person}{John Schulman}, \bibinfo{person}{Emanuel Todorov}, {and}
  \bibinfo{person}{Sergey Levine}.} \bibinfo{year}{2017}\natexlab{}.
\newblock \showarticletitle{Learning complex dexterous manipulation with deep
  reinforcement learning and demonstrations}.
\newblock \bibinfo{journal}{\emph{arXiv preprint arXiv:1709.10087}}
  (\bibinfo{year}{2017}).
\newblock


\bibitem[Reda et~al\mbox{.}(2022)]%
        {Brachiation2022}
\bibfield{author}{\bibinfo{person}{Daniele Reda}, \bibinfo{person}{Hung~Yu
  Ling}, {and} \bibinfo{person}{Michiel van~de Panne}.}
  \bibinfo{year}{2022}\natexlab{}.
\newblock \showarticletitle{Learning to Brachiate via Simplified Model
  Imitation}. In \bibinfo{booktitle}{\emph{ACM SIGGRAPH 2022 Conference
  Proceedings}} \emph{(\bibinfo{series}{SIGGRAPH '22})}. Article
  \bibinfo{articleno}{24}, \bibinfo{numpages}{9}~pages.
\newblock
\showISBNx{9781450393379}


\bibitem[Starke et~al\mbox{.}(2019)]%
        {starke_neural_2019}
\bibfield{author}{\bibinfo{person}{Sebastian Starke}, \bibinfo{person}{He
  Zhang}, \bibinfo{person}{Taku Komura}, {and} \bibinfo{person}{Jun Saito}.}
  \bibinfo{year}{2019}\natexlab{}.
\newblock \showarticletitle{Neural State Machine for Character-Scene
  Interactions}.
\newblock \bibinfo{journal}{\emph{ACM Transactions on Graphics}}
  \bibinfo{volume}{38}, \bibinfo{number}{6}, Article \bibinfo{articleno}{209}
  (\bibinfo{year}{2019}), \bibinfo{numpages}{14}~pages.
\newblock
\showISSN{0730-0301}


\bibitem[Tassa et~al\mbox{.}(2012)]%
        {tassa2012synthesis}
\bibfield{author}{\bibinfo{person}{Yuval Tassa}, \bibinfo{person}{Tom Erez},
  {and} \bibinfo{person}{Emanuel Todorov}.} \bibinfo{year}{2012}\natexlab{}.
\newblock \showarticletitle{Synthesis and stabilization of complex behaviors
  through online trajectory optimization}. In \bibinfo{booktitle}{\emph{2012
  IEEE/RSJ International Conference on Intelligent Robots and Systems}}. IEEE,
  \bibinfo{pages}{4906--4913}.
\newblock


\bibitem[Tsounis et~al\mbox{.}(2020)]%
        {tsounis2020deepgait}
\bibfield{author}{\bibinfo{person}{Vassilios Tsounis}, \bibinfo{person}{Mitja
  Alge}, \bibinfo{person}{Joonho Lee}, \bibinfo{person}{Farbod Farshidian},
  {and} \bibinfo{person}{Marco Hutter}.} \bibinfo{year}{2020}\natexlab{}.
\newblock \showarticletitle{Deepgait: Planning and control of quadrupedal gaits
  using deep reinforcement learning}.
\newblock \bibinfo{journal}{\emph{IEEE Robotics and Automation Letters}}
  \bibinfo{volume}{5}, \bibinfo{number}{2} (\bibinfo{year}{2020}),
  \bibinfo{pages}{3699--3706}.
\newblock


\bibitem[Viereck and Righetti(2021)]%
        {viereck2021learning}
\bibfield{author}{\bibinfo{person}{Julian Viereck} {and}
  \bibinfo{person}{Ludovic Righetti}.} \bibinfo{year}{2021}\natexlab{}.
\newblock \showarticletitle{Learning a centroidal motion planner for legged
  locomotion}. In \bibinfo{booktitle}{\emph{2021 IEEE International Conference
  on Robotics and Automation (ICRA)}}. IEEE, \bibinfo{pages}{4905--4911}.
\newblock


\bibitem[Wampler et~al\mbox{.}(2014)]%
        {wampler2014generalizing}
\bibfield{author}{\bibinfo{person}{Kevin Wampler}, \bibinfo{person}{Zoran
  Popovi{\'c}}, {and} \bibinfo{person}{Jovan Popovi{\'c}}.}
  \bibinfo{year}{2014}\natexlab{}.
\newblock \showarticletitle{Generalizing locomotion style to new animals with
  inverse optimal regression}.
\newblock \bibinfo{journal}{\emph{ACM Transactions on Graphics (TOG)}}
  \bibinfo{volume}{33}, \bibinfo{number}{4} (\bibinfo{year}{2014}),
  \bibinfo{pages}{1--11}.
\newblock


\bibitem[Wang et~al\mbox{.}(2012)]%
        {wang2012optimizing}
\bibfield{author}{\bibinfo{person}{Jack~M Wang}, \bibinfo{person}{Samuel~R
  Hamner}, \bibinfo{person}{Scott~L Delp}, {and} \bibinfo{person}{Vladlen
  Koltun}.} \bibinfo{year}{2012}\natexlab{}.
\newblock \showarticletitle{Optimizing locomotion controllers using
  biologically-based actuators and objectives}.
\newblock \bibinfo{journal}{\emph{ACM Transactions on Graphics (TOG)}}
  \bibinfo{volume}{31}, \bibinfo{number}{4} (\bibinfo{year}{2012}),
  \bibinfo{pages}{1--11}.
\newblock


\bibitem[Winkler et~al\mbox{.}(2018)]%
        {winkler2018gait}
\bibfield{author}{\bibinfo{person}{Alexander~W Winkler},
  \bibinfo{person}{C~Dario Bellicoso}, \bibinfo{person}{Marco Hutter}, {and}
  \bibinfo{person}{Jonas Buchli}.} \bibinfo{year}{2018}\natexlab{}.
\newblock \showarticletitle{Gait and trajectory optimization for legged systems
  through phase-based end-effector parameterization}.
\newblock \bibinfo{journal}{\emph{IEEE Robotics and Automation Letters}}
  \bibinfo{volume}{3}, \bibinfo{number}{3} (\bibinfo{year}{2018}),
  \bibinfo{pages}{1560--1567}.
\newblock


\bibitem[Won et~al\mbox{.}(2022)]%
        {won_physics-based_2022}
\bibfield{author}{\bibinfo{person}{Jungdam Won}, \bibinfo{person}{Deepak
  Gopinath}, {and} \bibinfo{person}{Jessica Hodgins}.}
  \bibinfo{year}{2022}\natexlab{}.
\newblock \showarticletitle{Physics-based character controllers using
  conditional {VAEs}}.
\newblock \bibinfo{journal}{\emph{ACM Transactions on Graphics}}
  \bibinfo{volume}{41}, \bibinfo{number}{4} (\bibinfo{year}{2022}),
  \bibinfo{pages}{96:1--96:12}.
\newblock
\showISSN{0730-0301}


\bibitem[Xie et~al\mbox{.}(2022)]%
        {xie2022glide}
\bibfield{author}{\bibinfo{person}{Zhaoming Xie}, \bibinfo{person}{Xingye Da},
  \bibinfo{person}{Buck Babich}, \bibinfo{person}{Animesh Garg}, {and}
  \bibinfo{person}{Michiel~van de Panne}.} \bibinfo{year}{2022}\natexlab{}.
\newblock \showarticletitle{Glide: Generalizable quadrupedal locomotion in
  diverse environments with a centroidal model}. In
  \bibinfo{booktitle}{\emph{Algorithmic Foundations of Robotics XV: Proceedings
  of the Fifteenth Workshop on the Algorithmic Foundations of Robotics}}.
  Springer, \bibinfo{pages}{523--539}.
\newblock


\bibitem[Xie et~al\mbox{.}(2020)]%
        {allsteps20}
\bibfield{author}{\bibinfo{person}{Zhaoming Xie}, \bibinfo{person}{Hung~Yu
  Ling}, \bibinfo{person}{Nam~Hee Kim}, {and} \bibinfo{person}{Michiel van~de
  Panne}.} \bibinfo{year}{2020}\natexlab{}.
\newblock \showarticletitle{ALLSTEPS: Curriculum-driven Learning of Stepping
  Stone Skills}.
\newblock \bibinfo{journal}{\emph{Computer Graphics Forum}}
  \bibinfo{volume}{39}, \bibinfo{number}{8} (\bibinfo{year}{2020}),
  \bibinfo{pages}{213--224}.
\newblock


\bibitem[Yao et~al\mbox{.}(2022)]%
        {controlvae22}
\bibfield{author}{\bibinfo{person}{Heyuan Yao}, \bibinfo{person}{Zhenhua Song},
  \bibinfo{person}{Baoquan Chen}, {and} \bibinfo{person}{Libin Liu}.}
  \bibinfo{year}{2022}\natexlab{}.
\newblock \showarticletitle{{ControlVAE}: {Model}-{Based} {Learning} of
  {Generative} {Controllers} for {Physics}-{Based} {Characters}}.
\newblock \bibinfo{journal}{\emph{ACM Transactions on Graphics}}
  \bibinfo{volume}{41}, \bibinfo{number}{6} (\bibinfo{year}{2022}),
  \bibinfo{pages}{183:1--183:16}.
\newblock
\showISSN{0730-0301}


\bibitem[Ye and Liu(2010)]%
        {ye2010optimal}
\bibfield{author}{\bibinfo{person}{Yuting Ye} {and} \bibinfo{person}{C.~Karen
  Liu}.} \bibinfo{year}{2010}\natexlab{}.
\newblock \showarticletitle{Optimal feedback control for character animation
  using an abstract model}.
\newblock \bibinfo{journal}{\emph{ACM Trans. Graph.}} \bibinfo{volume}{29},
  \bibinfo{number}{4} (\bibinfo{year}{2010}), \bibinfo{pages}{1--9}.
\newblock


\bibitem[Yin et~al\mbox{.}(2007)]%
        {yin2007simbicon}
\bibfield{author}{\bibinfo{person}{KangKang Yin}, \bibinfo{person}{Kevin
  Loken}, {and} \bibinfo{person}{Michiel Van~de Panne}.}
  \bibinfo{year}{2007}\natexlab{}.
\newblock \showarticletitle{Simbicon: Simple biped locomotion control}.
\newblock \bibinfo{journal}{\emph{ACM Transactions on Graphics (TOG)}}
  \bibinfo{volume}{26}, \bibinfo{number}{3} (\bibinfo{year}{2007}),
  \bibinfo{pages}{105--es}.
\newblock


\bibitem[Yin et~al\mbox{.}(2021)]%
        {jump2021}
\bibfield{author}{\bibinfo{person}{Zhiqi Yin}, \bibinfo{person}{Zeshi Yang},
  \bibinfo{person}{Michiel Van De~Panne}, {and} \bibinfo{person}{Kangkang
  Yin}.} \bibinfo{year}{2021}\natexlab{}.
\newblock \showarticletitle{Discovering diverse athletic jumping strategies}.
\newblock \bibinfo{journal}{\emph{ACM Transactions on Graphics}}
  \bibinfo{volume}{40}, \bibinfo{number}{4} (\bibinfo{date}{July}
  \bibinfo{year}{2021}), \bibinfo{pages}{91:1--91:17}.
\newblock
\showISSN{0730-0301}


\bibitem[Zhang et~al\mbox{.}(2018)]%
        {zhang_2018}
\bibfield{author}{\bibinfo{person}{He Zhang}, \bibinfo{person}{Sebastian
  Starke}, \bibinfo{person}{Taku Komura}, {and} \bibinfo{person}{Jun Saito}.}
  \bibinfo{year}{2018}\natexlab{}.
\newblock \showarticletitle{Mode-Adaptive Neural Networks for Quadruped Motion
  Control}.
\newblock \bibinfo{journal}{\emph{ACM Transactions on Graphics}}
  \bibinfo{volume}{37}, \bibinfo{number}{4} (\bibinfo{year}{2018}).
\newblock
\showISSN{0730-0301}


\end{thebibliography}



\begin{thebibliography}{5}


\ifx \showCODEN    \undefined \def \showCODEN     #1{\unskip}     \fi
\ifx \showDOI      \undefined \def \showDOI       #1{#1}\fi
\ifx \showISBNx    \undefined \def \showISBNx     #1{\unskip}     \fi
\ifx \showISBNxiii \undefined \def \showISBNxiii  #1{\unskip}     \fi
\ifx \showISSN     \undefined \def \showISSN      #1{\unskip}     \fi
\ifx \showLCCN     \undefined \def \showLCCN      #1{\unskip}     \fi
\ifx \shownote     \undefined \def \shownote      #1{#1}          \fi
\ifx \showarticletitle \undefined \def \showarticletitle #1{#1}   \fi
\ifx \showURL      \undefined \def \showURL       {\relax}        \fi
\providecommand\bibfield[2]{#2}
\providecommand\bibinfo[2]{#2}
\providecommand\natexlab[1]{#1}
\providecommand\showeprint[2][]{arXiv:#2}

\bibitem[Dorato et~al\mbox{.}(1994)]%
        {LQRbook}
\bibfield{author}{\bibinfo{person}{Peter Dorato}, \bibinfo{person}{Vito
  Cerone}, {and} \bibinfo{person}{Chaouki Abdallah}.}
  \bibinfo{year}{1994}\natexlab{}.
\newblock \bibinfo{booktitle}{\emph{Linear-Quadratic Control: An
  Introduction}}.
\newblock \bibinfo{publisher}{Simon \& Schuster}.
\newblock


\bibitem[Kwon and Hodgins(2017)]%
        {kwon2017momentum}
\bibfield{author}{\bibinfo{person}{Taesoo Kwon} {and}
  \bibinfo{person}{Jessica~K Hodgins}.} \bibinfo{year}{2017}\natexlab{}.
\newblock \showarticletitle{Momentum-mapped inverted pendulum models for
  controlling dynamic human motions}.
\newblock \bibinfo{journal}{\emph{ACM Transactions on Graphics (TOG)}}
  \bibinfo{volume}{36}, \bibinfo{number}{1} (\bibinfo{year}{2017}),
  \bibinfo{pages}{1--14}.
\newblock


\bibitem[Kwon et~al\mbox{.}(2020)]%
        {kwon2020fast}
\bibfield{author}{\bibinfo{person}{Taesoo Kwon}, \bibinfo{person}{Yoonsang
  Lee}, {and} \bibinfo{person}{Michiel Van De~Panne}.}
  \bibinfo{year}{2020}\natexlab{}.
\newblock \showarticletitle{Fast and flexible multilegged locomotion using
  learned centroidal dynamics}.
\newblock \bibinfo{journal}{\emph{ACM Transactions on Graphics (TOG)}}
  \bibinfo{volume}{39}, \bibinfo{number}{4} (\bibinfo{year}{2020}),
  \bibinfo{pages}{46--1}.
\newblock


\bibitem[Lee et~al\mbox{.}(2010)]%
        {lee2010data}
\bibfield{author}{\bibinfo{person}{Yoonsang Lee}, \bibinfo{person}{Sungeun
  Kim}, {and} \bibinfo{person}{Jehee Lee}.} \bibinfo{year}{2010}\natexlab{}.
\newblock \showarticletitle{Data-driven biped control}.
\newblock \bibinfo{journal}{\emph{ACM Trans. Graph.}} \bibinfo{volume}{29},
  \bibinfo{number}{4} (\bibinfo{year}{2010}), \bibinfo{pages}{1--8}.
\newblock


\bibitem[Peng et~al\mbox{.}(2018)]%
        {peng2018deepmimic}
\bibfield{author}{\bibinfo{person}{Xue~Bin Peng}, \bibinfo{person}{Pieter
  Abbeel}, \bibinfo{person}{Sergey Levine}, {and} \bibinfo{person}{Michiel
  van~de Panne}.} \bibinfo{year}{2018}\natexlab{}.
\newblock \showarticletitle{Deepmimic: Example-guided deep reinforcement
  learning of physics-based character skills}.
\newblock \bibinfo{journal}{\emph{ACM Transactions on Graphics (TOG)}}
  \bibinfo{volume}{37}, \bibinfo{number}{4} (\bibinfo{year}{2018}),
  \bibinfo{pages}{1--14}.
\newblock


\end{thebibliography}

\clearpage

\clearpage

\end{document}


\title{Supplementary Material for Adaptive Tracking of a Single-Rigid-Body Character in Various Environments}

\maketitle

\section{Details of LQR filtering for swing foot state}
\label{sec:swing_foot_state}

The global desired foot landing position $\check{\mb{f}}^j_t$ is calculated from $\mb{a}^j_s$, the desired foot landing position in the forward-facing SRB frame, as follows:
\begin{equation}
\label{eq:foot_state}
    \check{\mb{f}}^j_t= \textrm{projY}\left( \mathbf p_t^{\{g\}}+ \mathbf R_t^{y\{g\}}\left(\left(\mb{a}^j_s\right)_t+\mb{o}^j_t\right) \right) ,
\end{equation}
where $\mathbf R_t^{y\{g\}}$ refers to the orientation of the center of mass consisting only of rotation about the global vertical axis.
$\mb o_t$ is the foot offset vector that aligns the desired landing position $\check{\mb{f}}^j_t$ with the reference SRB motion's landing position when action $\mb{a}^j_s = \mb 0$ (refer to Section~\ref{sec:foot_offset} for the calculation of $\mb o_t$).
Function $\textrm{projY}(\cdot)$ projects the 3D global position onto the horizontal ground plane.
The global desired foot landing orientation $\check{\mb{F}}^j_t$ is directly obtained from the reference SRB motion.

Then, the foot state of the SRB character in a swing state is calculated as follows:
\begin{align}
\label{eq:smooth_foot_traj_p}
    \mb{f}^j_{t},\dot{\mb{f}}^j_{t}=\textrm{LQRfilter}\left(\mb{f}^j_{t-1},\dot{\mb{f}}^j_{t-1}, \check{\mb{f}}^j_{t},\alpha\right), \\
\label{eq:smooth_foot_traj_R}
    \mb{F}^j_{t},\dot{\mb{F}}^j_{t}=\textrm{LQRfilter}\left(\mb{F}^j_{t-1},\dot{\mb{F}}^j_{t-1}, \check{\mb{F}}^j_{t} ,\alpha \right)  .
\end{align}

Here, %
$\mb{f}^j_{t}$ and $\mb{F}^j_{t}$ are the continuously changing global position and orientation of each foot, which are LQR-filtered values of $\check{\mb{f}}^j_{t}$ and $\check{\mb{F}}^j_{t}$ (refer to Section~\ref{sec:lqr} for LQRFilter),
and $\dot{\mb{f}}^j_{t}$ and $\dot{\mb{F}}^j_{t}$ denote their time derivatives.
The parameter $\alpha$, which adjusts the strength of the LQR filtering, is experimentally set to 8 in our experiments.

\subsection{Foot offset vector $\mb o_t$}
\label{sec:foot_offset}

\begin{figure}[H]
  \centering
  \includegraphics[trim=130 50 70 40, clip, width=.2\linewidth]{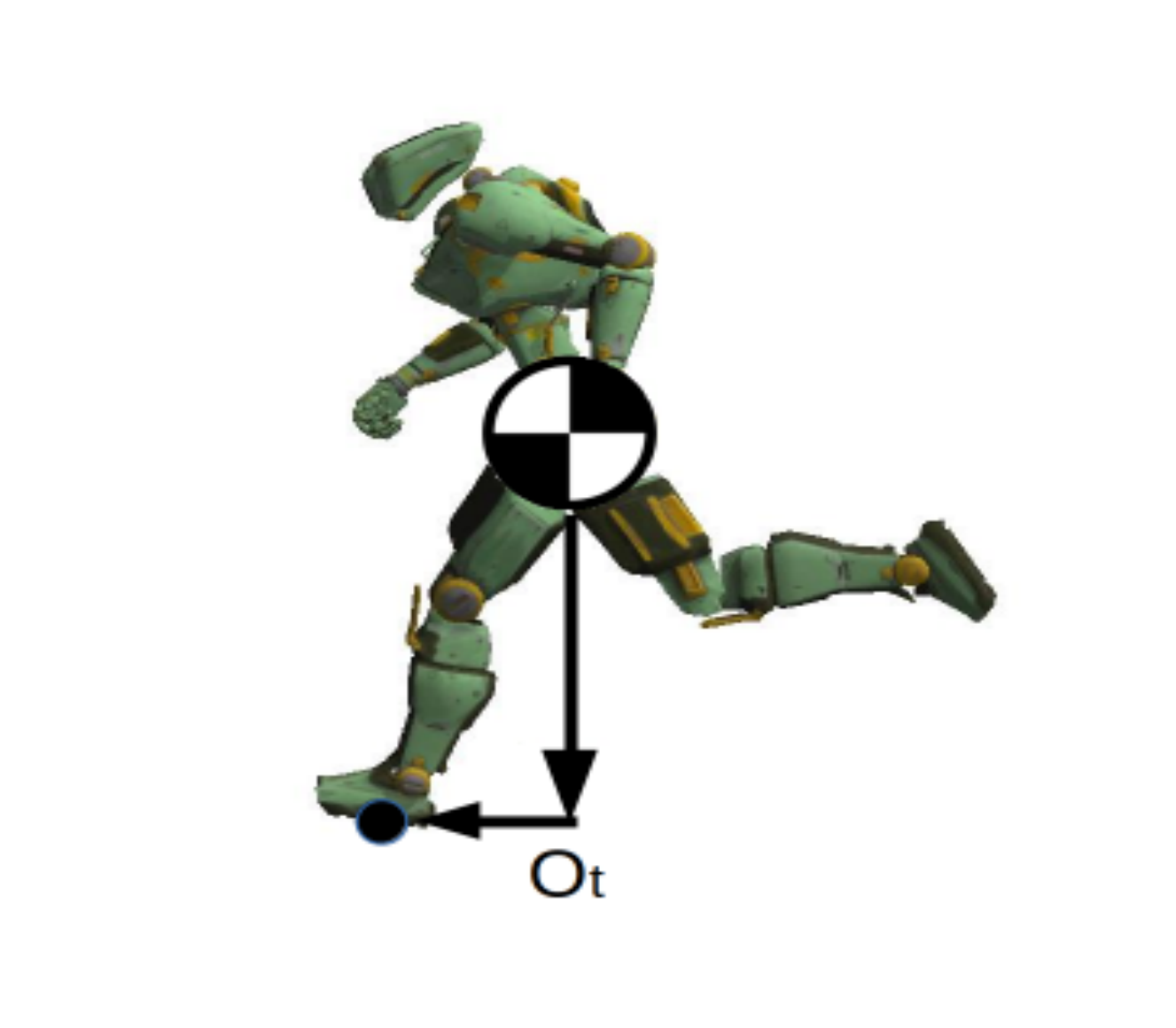}
  \caption{\label{fig:foot_offset}
           Foot offset vector $\mb o_t$.}
\end{figure}

The foot offset vector $\mb o_t$ used in Equation~\ref{eq:foot_state} adjusts the desired landing position of the $j^{th}$ foot at touch down to the landing position of the reference SRB motion when action $\mb{a}^j_s = 0$ (Figure~\ref{fig:foot_offset}). This is calculated as follows:
\begin{equation}
\label{eq:foot_offset}
    \mb{o}^j_t =   \left( \hat{\mathbf R}_{\psi(t)}^{y\{g\}} \right)^{-1}  \cdot \left(  \hat{\mb{f}}^j_{\psi(t)}  - \textrm{projY}\left( \hat{\mb{p}}^{\{g\}}_{\psi(t)} \right) \right)  ,
\end{equation}
where $\hat{\mathbf R}_{\psi(t)}^{y\{g\}}$ is a rotational matrix that only contains the rotational components of the reference SRB motion’s center of mass with respect to the global vertical axis, $\hat{\mb{f}}^j_{\psi(t)}$ is the global position of the $j^{th}$ foot of the reference SRB motion, and $\hat{\mb{p}}^{\{g\}}_{\psi(t)}$ is the global position of the reference SRB motion’s center of mass. All values are obtained for time ${\psi(t)}$ of the reference SRB motion that corresponds to the simulation time $t$. 
For reference, most motions can be learned when no offset is given ($\mb o_t = 0$), but the learning speed is faster when $\mb o_t$ calculated from Equation~\ref{eq:foot_offset} is given. Additionally, some motions may not be well learned when no offset is given.

\subsection{LQRFilter \label{sec:lqr}}

Our LQRFilter dynamically controls the mass particles, which represent the continuously changing positions and orientation of the feet, to follow the discontinuously changing desired landing positions and orientation.

Specifically, 1D mass particle dynamics simulation is used in our LQRFilter.
Three 1D mass particles are used for each leg to express the foot position (2D) and orientation (1D). 
The foot orientation is converted by a single scalar rotation angle $\theta$ to be expressed as a single mass particle, and used as an input to LQRFilter, and the filtered output is converted back into a rotation matrix. 
Here, $\theta$ is expressed as an angle relative to the frontal direction of the SRB to avoid singularity.

Linear quadratic regulators are used to control the mass particles. 
The equation of motion for each particle can be expressed as the state space equation of a continuous-time linear system: $\dot{\mathbf s} = \mathbf A \mathbf s + \mathbf B \mathbf u$.
For example, $\mathbf s$ is a 2D vector that stores the position and speed of a particle. 
The problem of physically controlling this particle using continuously changing external force $\mb u$ can be expressed as an optimal control problem using a linear quadratic regulator (LQR),
and the control results in the particle moving along a smooth $C^2$ continuous curve.
If the control objective function $J$ for guiding the particle to the origin is defined as
$J=\int _{0} ^{\infty}  
	 \left(
	 \mathbf s^T \mathbf Q\mathbf s +\mathbf u^T\mathbf R\mathbf u 
	 \right) dt
     $
     , the linear feedback matrix $\mb K$ that minimizes $J$ has an analytic solution~\cite{LQRbook}. 
By applying the linear feedback law $\mathbf u =-\mathbf K\left(\mathbf s-\mathbf s_d\right)$ to calculate the external force to guide the particle to the desired position, it is possible to calculate the continuously changing foot position and orientation (represented as particle states $\left\{\mathbf s\right\}$) of the SRB character that follows the discontinuous desired foot landing position and orientation (represented as desired particle states $\left\{\mathbf s_d\right\}$).

For instance, when considering a 1D particle representing the x-position of a foot, the desired particle state $\mb s_d$ is a two-dimensional vector, where the x-coordinate of the desired foot landing position is its position and its speed is set to 0.
By changing the positional component $Q_{0}=10^\alpha$ of the diagonal weighting matrix $\mathbf Q$, the speed at which how fast the particle moves towards its destination can be controlled. 
By using a large value for $\alpha$, the mass particle closely follows the desired foot landing position with minimal filtering.

\section{Motion phase adjustment details}

\textbf{Stride adjustment.}
The phase change rate is adjusted relative to the locomotion speed to prevent excessively lengthy strides. 
Let $\Delta v$ be the difference between the velocities of the reference and simulated centers of mass of the SRB character ($\Delta v=\left\| \Delta \mb p_t -\Delta \hat{\mb p} \right\|$).
Where the average moving speed of the reference motion is $\bar v$ and the reference phase change rate is $d\hat\psi/dt$, the phase change rate in the SRB simulation is increases as $\Delta v$ increases:
\begin{equation}
\frac{d\psi}{dt} \leftarrow
	\frac{d\hat\psi}{dt}\left(
		1+\beta{\Delta v}\bar v
	\right).
\end{equation}
In general, because $\Delta v$ converges to $0$ during the optimization process, the phase change rate is usually close to the reference phase change rate with deviations in the phase change rate only when there are external forces or terrain. 
The motion style can be changed by adjusting the weight $\beta$, which is experimentally set to 0.4 in our experiments.

\textbf{Contact timing adjustment.}
If an unexpectedly strong external force acts on the character, the character cannot touch down at the specified timings, failing to stably control the character.
To overcome this issue, we present a method to control the motion phase $\psi(t)$ by detecting the timing of early or late touch downs. 
When the height $y_t$ of the SRB character’s center of mass is greater than the height $\hat y_{\psi(t)}$ of the reference SRB motion’s center of mass beyond the given threshold (\SI{0.05}{m}) at the timing when touch down should have happened, 
it is considered that the contact between the foot and ground has not been made well yet, and is thus regarded as a late contact.
In such instances, we cut the phase change rate $d\psi/dt$ in half to delay the contact timing.
Conversely, where $y_t$ was smaller than $\hat y_{\psi(t)}$ beyond the threshold (\SI{0.1}{m}) before the contact timing of the SRB character, this is regarded as early touch down. 
In such instances, we double $d\psi/dt$ to allow contact before the given timing.

An earlier study \cite{lee2010data} also presented adjusting the phases of reference motions based on the simulated character's contact state. 
In their study, the phase of the reference motion is discontinuously adjusted upon touchdown and the contact time is strictly governed by the full-body simulator. 
In contrast, 
our simulated character follows the reference SRB motion, 
allowing for a more gradual adjustment of the motion phase, resulting in stable balancing of the character.

\section{Delta computation details}
\label{sec:computing_delta}

\textbf{Calculate COM frame delta.} The average transformation differences between the baseline SRB motion’s COM frame and the full-body reference motion’s COM frame are calculated for each different motion phase in $0 \leq \psi(t)\leq 2\pi$.
The average delta transformation is calculated as per $\psi(t)$ by Equation \ref{eq:com_delta}:
{\color{ysr}
\begin{align}
\label{eq:com_delta}
    \Delta\mb{\bar{T}}_{\psi(t)}
	=
	\frac{
		\sum_{c_n}\log
		\left(
		{\mb {\bar T}_{t}}^{-1} 
		\cdot \check{\mb{T}}_{\psi(t)}
		\right)
	}{
		c_n
	} ,
\end{align}
}
where $c_n$ refers to the number of cycles included in the baseline SRB motion, $\mb {\bar T}_{t}$ the baseline motion COM frame (i.e., SRB Frame) at time $t$, and $\check{\mb{T}}_{\psi(t)}$ the full-body reference motion's COM frame (i.e., reference SRB frame $\hat{\mb T}_{\psi(t)}$) at the corresponding time point. 
The calculated $\Delta\mb{\bar{T}}_{\psi(t)}$ is multiplied to the SRB frame of the SRB character in runtime simulation to calculate the target posture of MMIK, where the values stored in twist format are transformed into a rigid-body transformation by the exponential map.

\textbf{Calculate contact points delta.}
The baseline SRB motion involves smooth changes in the foot's position and orientation, but the feet remain in contact with the ground since it doesn't include foot height.
In order to convert this into a natural human foot movement, the average positions of the contact points of each foot of the full-body reference motion, expressed in the corresponding foot frame of the baseline SRB motion, are calculated for each different motion phase as follows:
\begin{equation}
\label{eq:swingfoot_delta}
    \Delta \bar{\mb{c}}_{\psi(t)}^{jk} =  
\frac{
        \sum_{c_n}
        \textrm{F}\left( \mb{\bar F}_t^j, \mb{\bar f}_t^j \right)^{-1} \cdot \check{\mb{c}}_{\psi(t)}^{jk\{g\}}
        }{
        c_n
    },
\end{equation}
where $\mb{\bar f}_t^j$ and $\mb{\bar F}_t^j$ refer to the global position and orientation of the $j^{th}$ foot in time $t$ for the baseline SRB motion, and $\check{\mb{c}}_{\psi(t)}^{jk\{g\}}$ refers to the global positions of the $k^{th}$ contact points of the $j^{th}$ foot of the corresponding full-body reference motion. 
The calculated $\Delta \bar{\mb{c}}_{\psi(t)}^{jk}$ is applied to the foot state of the SRB character in the runtime simulation to calculate the target contact positions of MMIK.

{\color{ysr}

\textbf{Calculate centroidal velocity delta.} The average differences between the linear and angular velocities of the baseline SRB motion and the centroidal linear and angular velocities of the full-body reference motion are calculated for each different motion phase as follows:
\begin{equation}
    \Delta \bar{\mb v}_{\psi(t)}=	
	\frac{
		\sum_{c_n}
        \textrm{Ad}_{ \left(\mathbf R_t^{y\{g\}}\right)^{\mathsf T}}
	\left(
    \mb{I}^{-1}_f \mb{J}_m \dot{\check{\mb{x}}}_{\psi(t)}-\bar{\mb v}_t
	\right)
	}{
		c_n
	}.
	\label{eq:vel_delta}
\end{equation}
Here, $\mb{I}_f$ refers to the complex rigid body inertia matrix of the full-body character, while momentum Jacobian $\mb{J}_m$ relates the full-body generalized velocity to the generalized centroidal momentum, $\bar{\mb{v}}_t$ refers to the global linear and angular velocities of the baseline SRB motion at time $t$, and $\dot{\check{\mb{x}}}_{\psi(t)}$ the generalized velocity of the full-body reference motion at the corresponding time.
Thus, $\mb{I}^{-1}_f \mb{J}_m \dot{\check{\mb{x}}}_{\psi(t)}$ refers to the full-body reference motion’s centroidal velocity. 
As $\mb{I}_f$ , $\mb{J}_m$, and $\bar{\mb v}_t$ are calculated in the global coordinate system, to convert $\Delta \bar{\mb v}_{\psi(t)}$ in reference to the forward-facing SRB frame, 
the Adjoint map $\textrm{Ad}_{ \left(\mathbf R_t^{y\{g\}}\right)^{\mathsf T}}$ for the inverse of the COM orientation only about the global vertical axis $\mathbf R_t^{y\{g\}}$ and the zero translation vector $\mb 0$ is used.

\section{MMIK Details}
\label{apdx:mmik}

MMIK minimizes the following cost function:
\begin{align} 
\label{eq:mmik}
\nonumber E_{IK}(\mb x) 
          &= \sum_{j}\sum_{k}  \|\mb {C}^{jk}(\mb{x}) - \textrm{F}\left( \mb{F}_t^j, \mb{f}_t^j \right) \cdot \Delta \bar{\mb{c}}_{\psi(t)}^{jk} \|^2 \\
\nonumber & + w_g \|\mb{I}^{-1}_f \mb{J}_m(\mb{x}-\bar{\mb{x}})\|^2 \\
\nonumber & + w_m \|\mb{I}^{-1}_f \mb{J}_m \dot{\mb{x}} - \left( \mb v_t +\textrm{Ad}_{\mathbf R_t^{y\{g\}}} \Delta \bar{\mb v}_{\psi(t)} \right) \|^2\\
          & + w_p \|\mb{\upsilon}_p-\mb{N}_p\mb{J}_p\dot{\mb{x}}\|_{\oplus}^2 + w_v \|\dot{\mb{x}}-\dot{\bar{\mb{x}}}\|^2 + w_r \|\mb{x}-\bar{\mb{x}}\|^2,
\end{align}

where $\mb x$ is the full-body pose, which is the optimization variable, and $\dot{\mb{x}}$ is the time derivative of the full-body pose calculated via backward differentiation. 
The full-body desired pose $\bar{\mb{x}}$ is obtained by applying rigid transformation to the full-body reference pose to align its COM frame $\check{\mb{T}}_{\psi(t)}$ with the COM frame $\bar{\mb{T}}_t \cdot \exp(\Delta \bar{\mb{T}}_{\psi(t)})$ of the simulated SRB character corrected with the COM frame delta calculated from Equation~\ref{eq:com_delta}. 
Its time-derivative $\dot{\bar{\mb{x}}}$  is also calculated by backward differentiation.

    The first term measures the difference between the position of the $k^{th}$ contact point of the $j^{th}$ foot of the full-body pose $\mb x$ ($\mb {C}^{jk}(\mb{x})$) and that of the simulated SRB character calculated by transforming the swing foot delta $\Delta \bar{\mb{c}}_{\psi(t)}^{jk}$ by the rigid transformation constructed form the position $\mb{f}_t^j$ and orientation $\mb{F}_t^j$ of the simulated SRB character's $j^{th}$ foot.

The second term, based on momentum-based geometric mapping \cite{kwon2017momentum}, aims to find $\mb{x}$ for which a change from $\mb{x}$ to $\bar{\mb{x}}$ does not cause any centroidal velocity.
$\mb{I}_f$ is the composite rigid body inertia matrix of the full-body character in the pose $\mb x$, and the momentum Jacobian $\mb{J}_m$ relates the full-body's generalized velocity to the generalized centroidal momentum.

The third term measures the difference between the global linear and angular velocity $\mb v_t +\textrm{Ad}_{\mathbf R_t^{y\{g\}}} \Delta \bar{\mb v}_{\psi(t)}$ of the simulated SRB motion calibrated by the centroidal velocity delta $\Delta \bar{\mb v}_{\psi(t)}$
and the centroidal velocity $\mb{I}^{-1}_f \mb{J}_m \dot{\mb{x}}$ calculated from $\dot{\mb{x}}$.
To convert $\Delta \bar{\mb v}_{\psi(t)}$ calculated in the forward-facing SRB frame to the global frame, Adjoint map $\textrm{Ad}_{\mathbf R_t^{y\{g\}}}$ was used.

The fourth term is only used when external forces are applied on the full-body character, which creates more natural responses to such forces. 
This term constrains the point at which the force is applied to move faster than $\mb{\upsilon}_p$ in the direction $\mb{N}_p$ of the applied force.
$\mb{\upsilon}_p$ is the heuristic speed threshold set in proportion to the magnitude of the external force. 
$\|\cdot\|_{\oplus}$ is assessed as 0 when the internal $\cdot$ has a negative value, and $\mb{J}_p$ is the Jacobian for the point of application of the external force.
This term uses redundant DOFs to create a natural response to the external force. 
Without the term, a non-compliance response will be generated where the whole body will bend uniformly under the external force.

The last two terms are regularization terms that maintain the solution close to the desired full-body pose $\bar{\mb{x}}$ and its time-derivative $\dot{\bar{\mb{x}}}$.

\section{Implementation and training details}
\label{sec:impl_train_details}

\textbf{Implementation.}
The quadratic programming solver was implemented in a Python 3.10.6 environment, and policy and value function network were implemented and trained using Pytorch ver.1.12.1.

\textbf{Policy network.} Our policy network consists of 64 hidden units, configured into two fully connected layers and one output layer. 
All hidden units used tanh for their activation function. The value network is also configured in the same structure. 
While the policy network has as many linear output units as the number of action dimensions, the value network has a single linear unit.
The policy was updated at every m = 1024 samples and mini-batch size n = 256 was used.
A discount factor $\gamma=0.995$ was applied.

\textbf{Episode termination condition.} To ensure an efficient learning journey, an episode should be immediately terminated when the simulation meets certain criteria and move onto the next episode.
In this study, the episodes were set to terminate when the height of the center of mass was too high or low, making it difficult to properly track the reference motion, or when the vertical axis (y-axis) of the SRB frame has an incline of more than 70 degrees against the global y-axis.
The valid height range for the center of mass was set to $\mathrm{min} (0.7 \cdot \hat{y}_{\psi(t)},0.2) < y_t < 2.0$, where $y_t$ refers to the height of the SRB character’s center of mass at time t and $\hat{y}_{\psi(t)}$ refers to that of the reference SRB motion.
Also, episodes were set to terminate when they exceeded 3 seconds.

\textcolor{rev}{
    \textbf{Reward weights.} The values of the reward weights used in our experiments are: $w^s=5$, $w^m=0.1$, $w^p=5$, $w^e=0.4$, $w^{p_1}=1$, $w^{p_2}=5$, $w^{p_3}=5$, and $w^{p_4}=5$. The alive reward $r^s_t$ is set to $1$.
}

\section{Additional experimental results}

\subsection{Sample efficiency}

\begin{figure}
  \centering
  
    \subfigure[\textit{Sprint jumps}]{ %
                \includegraphics[trim=0 0 0 0, clip, width=\columnwidth]{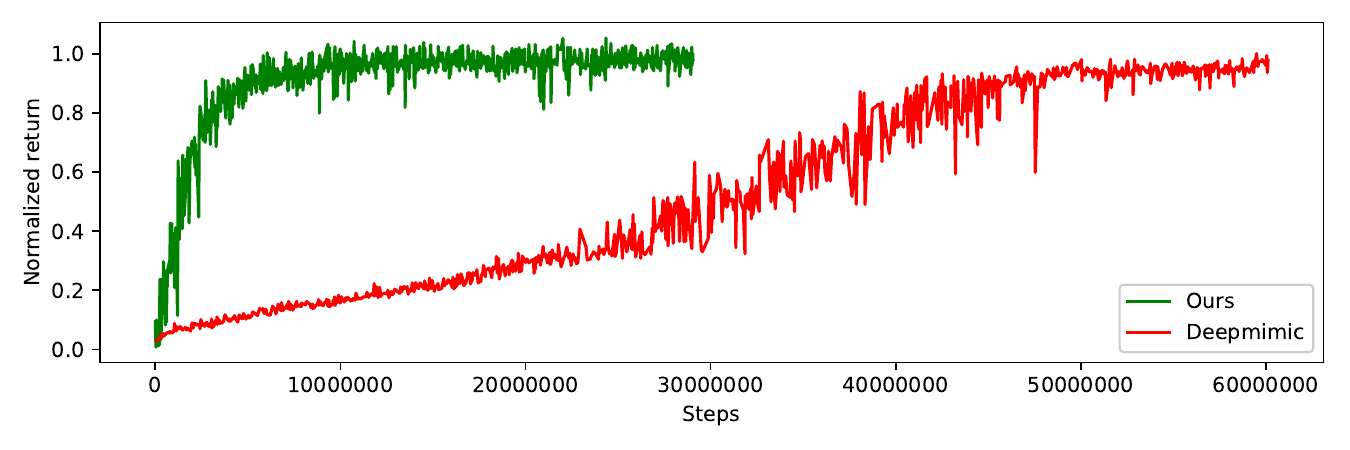}
				}
    \subfigure[\textit{Sprint}]{
                \includegraphics[trim=0 0 0 0, clip, width=\columnwidth]{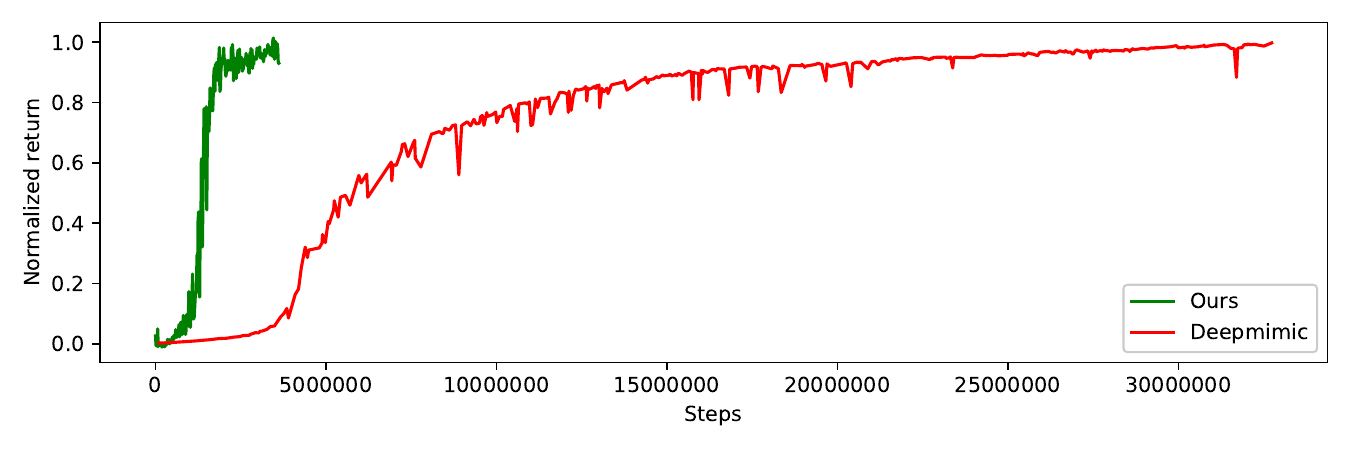}
				}
    \subfigure[\textit{Walk}]{
                \includegraphics[trim=0 0 0 0, clip, width=\columnwidth]{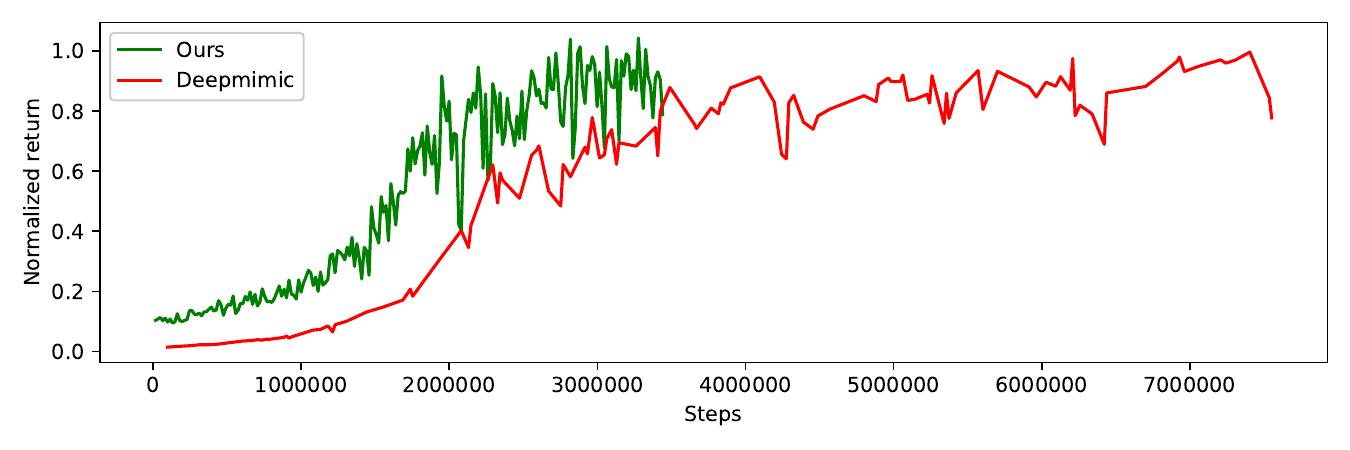}
				}
    \subfigure[\textit{Run}]{
                \includegraphics[trim=0 0 0 0, clip, width=\columnwidth]{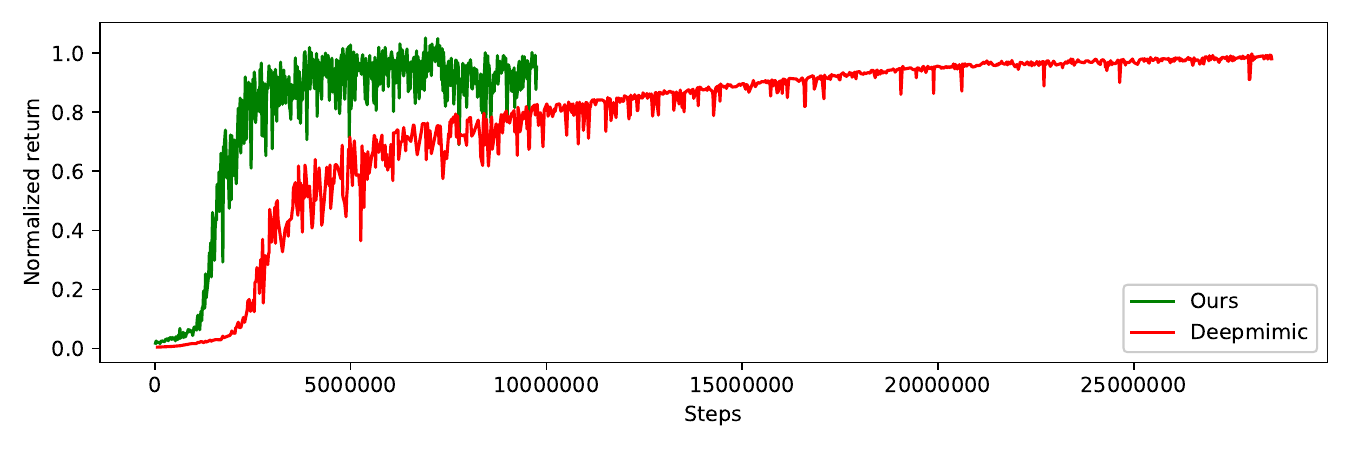}
				}
    \caption{Learning curves comparing our method and our implementation of DeepMimic.
    }
  \label{fig:learning_curves}
\end{figure}

Our approach has advantages in terms of adaptability and sample efficiency because it is based on simplified physics.
However, since it does not rely on full-body dynamics, it cannot fully represent general contact-rich scenarios.
In contrast, DeepMimic~\cite{peng2018deepmimic}, being based on full-body dynamics, can accurately represent various motions in general physically-interacting situations,
but it has lower adaptability due to full-body tracking and requires a large number of samples for learning due to the high dimensionality.
Given these fundamental differences in their characteristics, it is not appropriate to compare the two methods directly.
However, for reference purposes, we would like to discuss their sample efficiency using learning curves.

Figure~\ref{fig:learning_curves} shows the learning curves of our method and DeepMimic for four reference motions.
It can be seen that our method can learn the control policy with a smaller number of samples than DeepMimic.
Specifically, ours requires about 8\% (\textit{Sprint}) to 38\% (\textit{Walk}) samples for the policy to converge.
The number of samples required for convergence with DeepMimic varies greatly depending on the dynamics of the reference motion (\textit{Walk}: about $7.4 \times 10^6$, \textit{Sprint}: about $3 \times 10^7$, \textit{Sprint jumps}: about $5 \times 10^7$), while our method has a relatively small difference in the number of samples required for different reference motions (\textit{Walk}: about $2.8 \times 10^6$, \textit{Sprint}: about $2.5 \times 10^6$, \textit{Sprint jumps}: about $7 \times 10^6$).
We attribute this sample efficiency to the fact that our policy is not trained to output detailed full-body actions based on the detailed state of the full-body, but rather to output the overall desired velocity and contact position of a single rigid body based on the overall state of the single body.
Our method is faster in terms of wall-clock time as well, as it can collect more samples during the same period (ours: about 2000 steps per second, DeepMimic: about 1800 steps per second).

\subsection{Runtime computational statistics}

\begin{table}
\caption
    {Statistics for generating locomotion on a flat ground.
    In the table, average computation time for generating one second of final motion is measured.
Total time includes all the necessary computation time except the rendering time.
}

\label{tab:runtime}
\begin{center}
    \begin{tabular}{|l|l|c|c|c|}
\hline
    \multicolumn{5}{|c|}{ average computation time for 1s full-body motion } \\ 
    \hline
    \hline

    \multicolumn{2}{|c|}{} & Walk     &     Sprint &    Run180    \\ 
    \hline

    \multirow{4}{*}{Ours} & policy output       & 0.007s		& 0.007s &	0.007s	\\ \cline{2-5}
                          & QP simulation       & 0.061s		& 0.034s & 0.042s		 \\ \cline{2-5}
                          & MMIK(with $\Delta$) & 0.23s 		& 0.18s & 0.19s		 \\ \cline{2-5}
                          & total 		        & 0.29s		& 0.22s &	0.24s 		\\ \hline
    \hline

    \multicolumn{2}{|c|}{} & Walk     & Fast run &    Run    \\ 
    \hline

    \multirow{4}{*}{\cite{kwon2020fast}} & motion sketch  & 0.07s		& 0.18s & 0.10s	\\ \cline{2-5}
                            & CDM planning   & 0.16s		& 1.08s & 0.13s 	 \\ \cline{2-5}
                            & MMIK 		     & 0.23s 		& 0.23s & 0.23s	 \\ \cline{2-5}
                            & total 	     & 0.46s		& 1.49s & 0.46s		\\ \hline

\end{tabular}
\end{center}

\end{table}

Table~\ref{tab:runtime} presents the runtime computational statistics for our method and \cite{kwon2020fast}.
Our controllers can generate a full-body motion approximately four times faster than real time, and a large part of the computation time is spent solving MMIK. 
Note that learning is performed much faster because there is no need to solve MMIK during learning. 
For a rough comparison, we included the runtime statistics from \cite{kwon2020fast}, indicating the time required to generate motions similar to ours as closely as possible.
In their work, real-time generation of some motions is possible, but generating faster or shorter stance motions through trajectory optimization requires more time due to increased complexity and longer unactuated phases.

\subsection{MMIK with and without precomputed delta}
\label{apdx:delta_comp}

To validate the impact of the COM frame delta $\Delta\mb{\bar{T}}_{\psi(t)}$ and the centroidal velocity delta $\Delta \bar{\mb{v}}_{\psi\left(t_p\right)}$ (described in Section~\ref{sec:computing_delta}) on the full-body motion quality generated by MMIK, we have compared the simulation results where such terms were turned On/Off. 
The upper body movement increased to the point of looking unsteady when the COM frame delta were switched off, while the upper body and the head shook unnaturally when the centroidal velocity delta was turned off. 
These differences can be best observed in the accompanying video.

\bibliographystyle{ACM-Reference-Format} 
\bibliography{egbibsample2}